\documentclass[a4paper,10pt,onecolumn]{article}

\usepackage[T1]{fontenc}
\usepackage[utf8]{inputenc}
\usepackage[english]{babel}
\usepackage[sort&compress]{natbib}
\usepackage{amsmath}
\usepackage{amssymb}
\usepackage{cancel}

\usepackage{xcolor}
\usepackage{graphicx, url}
\usepackage{grffile}

\graphicspath{{./assets/}}

\usepackage{subcaption}

\usepackage{booktabs}

\title{Bayesian Sparsification Methods for Deep Complex-valued Networks}
\author{Ivan Nazarov, and Evgeny Burnaev}

\newcommand{\real}{\mathbb{R}}
\newcommand{\cplx}{\mathbb{C}}
\newcommand{\iu}{{\jmath}}

\newcommand{\conj}[1]{\overline{#1}}

\newcommand{\diag}[1]{\mathrm{diag}{#1}}

\usepackage{etoolbox}
\makeatletter
\patchcmd{\@citex}{\bfseries ?}{\colorbox{red}{\bfseries ?}}{}{}
\makeatother

\begin{document}
\maketitle

\begin{abstract}
With continual miniaturization ever more applications of deep learning can be found
in embedded systems, where it is common to encounter data with natural representation
in the complex domain. To this end we extend Sparse Variational Dropout to complex-valued
neural networks and verify the proposed Bayesian technique by conducting a large numerical
study of the performance-compression trade-off of $\cplx$-valued networks on two tasks:
image recognition on MNIST-like and CIFAR10 datasets and music transcription on MusicNet.
We replicate the state-of-the-art result by \citet{trabelsi_deep_2018} on MusicNet with
a complex-valued network compressed by $50-100\times$ at a small performance penalty.
\end{abstract}

\section{Introduction} % (fold)
\label{sec:introduction}

% general intro text with motivation
Deep neural networks are an integral part of machine learning and data science toolset
for practical data-driven problem solving. With continual miniaturization ever more
applications can be found in embedded systems. Common embedded applications include
on-device image recognition and signal processing. Despite recent advances in generalization
and optimization theory specific to deep networks, deploying in actual embedded hardware
remains a challenge due to storage, real-time throughput, and arithmetic complexity
restrictions \citep{han_learning_2015}. Therefore, compression methods for achieving high
model sparsity and numerical efficiency without losing much in performance are especially
relevant.
% Solving the storage and artihmetic constraints encourages development of model compression
% and sparsification methods, that focus on favourable trade-off between performance and size.

% applications of cvnn
Complex-valued nature of the data in acoustic and radio signal processing has been
the main driver behind the adoption of $\cplx$-valued neural networks ($\cplx$VNN).
\citet{hirose_complex-valued_2009} argues that the combined phase-magnitude effect of
$\cplx$-valued transformations removes the excess degrees of freedom, that cause
degenerate transformations in $\real$-valued networks with twice the feature dimensions.
Their study demonstrates superiority of $\cplx$VNN in landmine detection using ground
penetrating radar imaging. Other examples, where $\cplx$-valued networks have outperformed
$\real$-valued networks, include magnetic resonance \citep{hui_mri_1995,wang_deepcomplexmri_2020}
and radar imaging \citep{haensch_complex-valued_2010,zhang_complex-valued_2017}, music
transcription and spectral speech modelling \citep{wisdom_full-capacity_2016,trabelsi_deep_2018},
and wireless signal classification \citep{yang_complex_2020}. \citet{tarver_design_2019}
have lowered the out-of-band power leakage with a $\cplx$-valued network for digital
signal predistortion. The networks have also been applied to non-$\cplx$-valued domains,
such as image classification \citep{popa_complex-valued_2017}, sequence modelling
\citep{danihelka_associative_2016}, and motion prediction \citep{wolter_complex_2018},
and for stabilizing back-propagation in RNN \citep{wisdom_full-capacity_2016}.

% motivation and other methods
Despite promising results for embedded signal processing applications, $\cplx$-valued
networks remain a niche in deep learning, and as such little attention has been paid
to compression methods specific to $\cplx$VNN.
Yet there is an abundance of research related to real-valued network compression, and
many results can be applied to $\cplx$VNN. Methods such as knowledge distillation
\citep{hinton_distilling_2015}, which trains a small network to replicate a large
well-trained teacher, low-rank matrix \citep{denton_exploiting_2014} and tensor decomposition
\citep{novikov_tensorizing_2015}, or magnitude-based parameter pruning \citep{zhu_prune_2018}
can be adapted to $\cplx$VNN without modifications. Parameter quantization and conversion
from floating to fixed point arithmetic \citep{courbariaux_training_2015,uhlich_mixed_2020},
appear to be readily applicable as well. For example, \citet{wu_compressing_2019} adapt
$k$-means quantization to $\cplx \simeq \real^2$ parameters and successfully compress
$\cplx$VNN with the ``prune-quantize-code'' procedure of \citet{han_deep_2016}.

Other methods cannot be translated to $\cplx$VNN this straightforwardly. Probabilistic
$\ell_0$ regularization of \citet{louizos_learning_2018} prune networks using multiplicative
$[0, 1]$-valued stochastic masks with distributions having an atom at $0$, yet differentiable
via the reparameterization trick \citep{kingma_auto-encoding_2014}. By sharing a single mask
value within a group of parameters their approach can be adapted to $\cplx$ parameters.
However, methods such as Hessian-based parameter pruning \citep{lecun_optimal_1990}
or Sparse Variational Dropout \citep{molchanov_variational_2017} require additional
considerations.
% the more explicitly zero parameters there is, the less computations is needed.

% segue into why we chose sparse VD rather than anything else
% (sparsity tech large-scale models) gale-elsen-hooker.tex
\citet{gale_state_2019} compare magnitude pruning, $\ell_0$ regularization and Sparse
Variational Dropout (VD) on large-scale models. Their results suggest that VD may achieve
good accuracy-sparsity balance and outperform pruning and $\ell_0$ in deep architectures,
although pruning is preferred for simplicity, stability and speed. They also observe that
VD induces non-uniform sparsity throughout the model, which \citet{he_amc:_2018} have shown
to be essential for superior compression.
% Magnitude pruning induces user-specified sparsity distributions.

% our contribution
Sparse Variational Dropout is a Bayesian Variational Inference method with automatic
parameter relevance determination effect. In this study we extend \emph{Sparse VD} to
$\cplx$VNN, inspired by the results of \citet{gale_state_2019}, and motivated by seldom
application of Bayesian Inference to $\cplx$-valued networks \citep{popa_complex-valued_2017}
and apparent scarcity of compression methods specific to them. We assess the performance-%
compression trade-off of the extension by conducting a large-scale numerical study on
image classification on MNIST-like and CIFAR10 datasets and music transcription on MusicNet.

The paper is structured as follows. Sec.~\ref{sec:variational_dropout} reviews Variational
Dropout, and sec.~\ref{sec:c_valued_networks} provides a brief summary of the inner
workings of complex-valued networks. The main contribution of this study is presented
in sec.~\ref{sec:c_variational_dropout}, where we provide the details of $\cplx$-valued
variational sparsification methods. In sec.~\ref{sec:experiments} we estimate the compression
and performance trade-off on shallow and deep $\cplx$-valued networks, and discuss the
outcomes.

% section introduction (end)

\section{Variational Dropout} % (fold)
\label{sec:variational_dropout}

\subsection{Variational Inference} % (fold)
\label{sub:variational_inference}

% overview
In broad terms Bayesian Inference is a principled framework for reasoning about uncertainty
and updating prior beliefs about model's parameters in accordance with evidence or
empirical data into a posterior distribution. The posterior is useful for inference
regarding unobserved data, predictive statistics, parameter confidence regions, and
model's uncertainty.
% intuitive definition of  of model given data is the same as probability of data under hypothesis (model)
% KL-div min & VI -- \citep{jordan_introduction_1999} approximate Bayesian inference
% SVI -- \citep{hoffman_stochastic_2013} scalable stochastic variant
% SGVB -- \citep{kingma_auto-encoding_2014} differentiable MCMC estimator of the evidence lower bound for SVI

For an observed dataset $
  D = (x_i)_{i=1}^N
$ and statistical model $
  p(D \mid \omega)
    = \prod_i p(x_i \mid \omega)
    % $D$ conditional independence given parameters $\omega$
$ with parameters $\omega$ the Bayes rule transforms prior hypotheses $\pi(\omega)$
about the unknown distribution of model's parameters into the posterior distribution: $
  p(\omega \mid D) = \tfrac{p(D \mid \omega) \pi(\omega)}{p(D)}
$.
%
% In general the marginal likelihood of the dataset is intractable $
%   p(D) = \mathbb{E}_{\pi(\omega)} p(D \mid \omega)
% $.
Save for the relatively simple set-ups, either the posterior distribution itself or the
mathematical expectations it is involved in are analytically intractable or impractical
to compute numerically. Variational Inference (VI), proposed by \citet{jordan_introduction_1999},
can be used in such cases to make approximate inference. The approach finds an approximation
within some distribution family $q_\theta(\omega)$, which is closest to the true posterior
distribution in terms of Kullback-Leibler divergence: $
  KL(q_\theta(\omega) \| p(\omega \mid D))
    % = \int \frac{dQ}{dP} \log{\frac{dQ}{dP}} P(d\omega)
    % = \int \frac{q(\omega)}{p(\omega)} \log{\frac{q(\omega)}{p(\omega)}} p(\omega) d\omega
    % = \mathbb{E}_{\omega \sim q}
    %   \log\frac{q(\omega)}{p(\omega)}
    = \mathbb{E}_{\omega \sim q_\theta}
      \log \tfrac{q_\theta(\omega)}{p(\omega \mid D)}
$. % The approximation family is chosen so that the expectations become tractable.
\citet{jordan_introduction_1999} show that this problem is equivalent to variational
\emph{maximization} of the \emph{Evidence Lower Bound} (ELBO)
\begin{equation}  \label{eq:elbo_general}
  \mathcal{L}(\theta; \lambda)
    = - KL(q_{\theta} \| \pi_{\lambda})
      + \sum_{i=1}^N \mathbb{E}_{\omega \sim q_{\theta}}
        \log p(x_i \mid \omega)
      % \mathbb{E}_{\omega \sim q_{\theta}}
        % \log\frac{q_\theta(\omega)}{\pi_{\lambda}(\omega)}
  % = \mathbb{E}_{\omega \sim q_{\theta}}
  %     \log{p_{\phi}(D \mid \omega) \pi_{\lambda}(\omega)}
  %   + \mathbb{H}(q_{\theta})
  \,,
\end{equation}
where the variational parameters $\theta$ and $\lambda$ parameterize the approximation
and the prior, respectively.
Kullback-Leibler and, by proxy, ELBO are standard objectives in VI, however it is possible
to use other objectives, provided the true posterior $p(\omega \mid D)$ is evaluated only
through $\log p(D \mid \omega)$ and $\log \pi(\omega)$, \citep{ranganath_operator_2016}.

% speed, scale, variance, family of suitable Var approximations
In subsequent years several improvements to Variational Inference approach were introduced.
To make VI able to handle large-scale datasets \citet{hoffman_stochastic_2013} proposed
\emph{Stochastic Variational Inference}, which uses stochastic gradient optimization of
\eqref{eq:elbo_general} based on noisy unbiased gradient estimates of ELBO
computed on random mini-batches from the dataset. \citet{titsias_doubly_2014} translated
the dependence on location–scale parameters of $q_{\theta}$ to the function inside its
expectation and proposed \emph{Doubly Stochastic Variational Inference}. DSVI constructs
an unbiased finite-sample estimator of the gradient of \eqref{eq:elbo_general} by both
subsampling the dataset and sampling from $q_{\theta}$, without forfeiting convergence
of SVI.

% The variance of the stochastic gradient directly affects convergence of SVI.
Independently, \citet{kingma_auto-encoding_2014} proposed \emph{Stochastic Gradient
Variational Bayes}, which is an alternative efficient doubly stochastic estimator
applicable to models with parameters $\omega$ that are continuous random variables
amenable to the \emph{reparameterization trick}, i.e. $
  \omega \sim q_{\theta}(\omega)
$ is equivalent in distribution to $
  \omega = g_{\theta}(\varepsilon)
$ for some non-parametric random variable $
  \varepsilon \sim p(\varepsilon)
$ and $
  g(\varepsilon; \theta)
$ differentiable with respect to $\theta$. The estimator of \eqref{eq:elbo_general}
with $L$ reparameterized draws per element in the mini-batch of size $M$ is given by
\begin{equation}  \label{eq:sgvb_estimator}
  \widetilde{\mathcal{L}}(\theta; \lambda)
    = - KL(q_{\theta} \| \pi_{\lambda})
      % + \frac{N}{\lvert B \rvert} \sum_{i\in B}
        % \log p(x_i \mid \omega_i)b
      + \frac{N}{M L} \sum_{k,l}
        \log p(x_{i_k} \mid g(\varepsilon_{lk}; \theta))
      % + \frac{N}{M L} \sum_{k=1}^M \sum_{l=1}^L
      %   \log p(x_{i_k} \mid g(\varepsilon_{lk}; \theta))
        % \log p(D \mid \omega_{l})
        %   \Big\vert_{\omega_{l} = g(\varepsilon_{l}; \theta)}
      % =
      % - KL(q_{\theta} \| \pi)
      % + N \mathbb{E}_{B \sim \{D\}_M \otimes q_\theta^M}
        % \hat{\mathbb{E}}_{x,\omega \sim B}
          % \log p(x \mid \omega)
    \,,
\end{equation}
where $(x_{i_k})_{k=1}^M$ is a random subsample from $D$ and $
  (\varepsilon_{lk})_{l=1}^L
$ are $k=1..M$ independent iid samples from $p_\varepsilon$. \citet{figurnov_implicit_2018}
extended the scope of the reparameterization gradients to include continuous distributions
such as Gamma and von Mises. To handle the case of non-reparameterizable $\omega$ in
doubly stochastic VI, e.g. discrete random parameters, \citet{titsias_local_2015} proposed
\emph{local expectation gradients}, which is a version of REINFORCE gradient estimator
\citep{williams_simple_1992} with variance reduced by careful use of dependence
structures in the model.
% (reparameterization trick, pathwise gradient) $q_{\theta}(d\omega)$ is a push-forward
% of $p(d\varepsilon)$ by a differentiable map $g(\varepsilon; \theta)$.
% % http://stillbreeze.github.io/REINFORCE-vs-Reparameterization-trick/#fn:1
% rao-blackwellization (http://pyro.ai/examples/svi_part_iii.html)

% The Local Reparameterization Trick reduces the variance of SGVB
% even further by sampling independent weight matrices for
% each data-point inside mini-batch. So we need both original SGVB
% (per sample randomness) and local reparameterization (that is an
% efficient and less noisy equivalent to per-sample weight sampling)
% SGVB is specifically for VAE (local latent variables per data-point)
% what Kingma propose is SGVB for global shared latent variables aka
% WEIGHTS!!
In this study we use the SGVB estimator \eqref{eq:sgvb_estimator} with $L=1$ and the
\emph{local reparameterization trick} proposed by \citet{kingma_variational_2015}. They
argued that this gradient estimator can be made more statistically and computationally
efficient, if the structure of the model permits translating global stochasticity of
$\omega$ down to local intermediate states of computation.
The class of models that allow this include non-recurrent computational graphs,
exemplified by neural networks with parameters $\omega \sim q_\theta$. In their case,
\eqref{eq:sgvb_estimator} would require that the \emph{entire} set of network's parameters
$\omega$ be independently drawn for each element in the mini-batch. Since that the
parameters in a network naturally split into subsets with non-overlapping layer-wise
effects, it is standard to assume that the approximation $
  q_\theta(\omega)
$ is factorized over layers. Furthermore, if $
  % \omega =
  W \in \mathbb{R}^{n\times m}
$ in a linear layer $
  y = b + W^\top x
$ with $
  q_\theta(W)
    = \prod_{ij}
      \mathcal{N}(w_{ij} \vert\, \mu_{ij}, \sigma_{ij}^2)
$, then by virtue of $y$ being a linear transformation of $W$, we get
\begin{equation}  \label{eq:r-gauss-trick}
  q(y) = \prod_i \mathcal{N}\Bigl(
        y_i \big\vert\,
        b_i + \sum_j \mu_{ij} x_j,
        \sum_j \sigma^2_{ij} x_j^2
    \Bigr)
  \,.
\end{equation}
This yields outputs equivalent in distribution to sampling $W$ for each element in
the mini-batch, which produces the SGVB estimator with smaller variance, as demonstrated
by \citet{kingma_variational_2015}.
% Similar idea to \eqref{eq:r-gauss-trick} was proposed by \citet{wang_fast_2013} to speed
% up Gaussian dropout.

% subsection variational_inference (end)

\subsection{Dropout} % (fold)
\label{sub:dropout}

Variational Inference can be used as model regularization and sparsification method
for certain posterior approximation $q_\theta$ and prior $\pi$.

% on dropout
Dropout, proposed by \citet{hinton_improving_2012}, prevents overfitting by injecting
multiplicative binary noise into layer's weights, which breaks up co-adaptations that
could occur during training. \citet{wang_fast_2013} argued that the overall effect of
binary Dropout on the intermediate outputs can be approximated by a Gaussian with weight-input
dependent mean and variance via the Central Limit Theorem. \citet{srivastava_dropout_2014}
proposed using independent $\mathcal{N}(1, 1)$ multiplicative noise, arguing that higher
entropy of a Gaussian has better regularizing effect. \citet{gal_dropout_2016} showed
that Dropout is a Bayesian approximation method with close ties to deep Gaussian Processes
that yields inexpensive model uncertainty estimates. In a study concerning multitask learning
\citet{cheung_superposition_2019} demonstrated the possibility of storing task-specific
parameters in non-destructive superposition within a single network. Regarding Dropout
their argument implies that if the single task setting is viewed as multitask learning
with replicated task, then by sampling uncorrelated binary masks Dropout acts as a
superposition method, utilizing the learning capacity of the network better.
% many identical copies of the same task; \cite[Appendix A.1]{cheung_superposition_2019}
% (fast dropout gauss-approx) wang-manning.tex
% (gauss dropout) srivastava-et-al.tex

% VD is a Bayesian perspective on dropout regularization \citep{hinton_improving_2012} offered
% by \citet{kingma_variational_2015} and repurposed for sparsification by \citet{molchanov_variational_2017}.
%
\citet{kingma_variational_2015} provided a unifying perspective on Dropout, DropConnect
\citep{wan_regularization_2013}, and Gaussian Dropout \citep{wang_fast_2013} through
the lens of Variational Inference and propose \emph{Variational Dropout}. They argued
that the multiplicative noise introduced by Dropout methods induces a distribution
equivalent to a fully factorized variational posterior $
  q_\theta(\omega) = \prod_j q_{\theta}(\omega_j)
$, where $q_{\theta}(\omega_j)$ is $\omega_j = \mu_j \eta_j$ with $
  \eta_j \sim p_\theta(\eta_j)
$ iid from some $p_\theta(\eta)$.
%
% Binary dropout with rate $p \in (0, 1)$ uses $
%   p_\eta
%     = \mathcal{B}\bigl(
%       \{0, \tfrac1{1-p}\}, 1-p
%     \bigr)
% $, whereas in Gaussian dropout $
%   \eta \sim \mathcal{N}(1, \alpha)
% $ for $\alpha = \tfrac{p}{1-p}$.
% This suggests making $\alpha$ a variational parameter
% and optimizing it in \eqref{eq:elbo_general} with an appropriate penalty term. Thus,

Variational Dropout uses \emph{fully factorized Gaussian} approximation $
  q_\theta(\omega)
    = \prod_j \mathcal{N}(\omega_j \vert\, \mu_j, \alpha_j \mu_j^2)
$ and factorized scale invariant log-uniform prior $\pi(\omega)$ with $
  \pi(\omega_j) \propto \lvert \omega_j \rvert^{-1}
$. \citet{molchanov_variational_2017} noticed that $\alpha_j$ reflects the relevance of the
parameter $\omega_j$ it is associated to by being the ratio of its squared mean to its
effective variance. Based on this observation they proposed \emph{Sparse Variational Dropout},
a modification that enables automatic model sparsification by optimizing $\alpha_j$ for each
individual parameter. \citet{louizos_bayesian_2017} extended the idea to structured sparsity
by considering hierarchical prior and variational approximation. They grouped the parameters
$\omega_j$ and coupled them within each one through a shared latent variable, which on the
whole enabled pruning entire input features in each layer.
% {louizos_bayesian_2017} eq. (11) variance -> gaussian entropy -> informative bits
% They also proposed to estiamte bit precision for quantization based on parameter's
% estimated uncertainty (variance or entropy).

Due to factorization assumption, the term $KL(q_\theta \| \pi)$ in \eqref{eq:sgvb_estimator}
for Sparse VD unravels into $
  \sum_j K(\tfrac{\sigma^2_{j}}{\mu_{j}^2})
$ with
\begin{equation}  \label{eq:improper-kl-div-real}
  K(\alpha)
    \propto \frac12 \mathbb{E}_{\varepsilon \sim \mathcal{N}(0, 1)}
        \log{\Bigl\lvert \tfrac1{\sqrt{\alpha}} + \varepsilon \Bigr\rvert^2}
      % \frac12 \mathbb{E}_{\varepsilon \sim \mathcal{N}(0, 1)}
      %   \log{\bigl\lvert 1 + \sqrt{\alpha} \varepsilon \bigr\rvert^2}
      % - \frac12 \log{\alpha}
  \,.
\end{equation}
\citet{kingma_variational_2015} approximated $K(\alpha)$ over $\alpha \in (0, 1)$ by a
polynomial with a logarithmic term, and later \citet{molchanov_variational_2017} refined
the approximation of \eqref{eq:improper-kl-div-real} by weighted sum of a sigmoid and a
soft-plus term.
% see p.6 sec 3.4 of kingma_variational_2015
In appendix~\ref{sec:real-chisq-grad} we verify the derivative of their approximation
against a Monte-Carlo estimate for $\alpha$ varying over a fine $\log$-scale grid and
the exact expression for gradient of \eqref{eq:improper-kl-div-real}.
% (!) good for SGVB: gradient-based methods require unbiased gradient estimates
% (in) it is in poor taste to brag about this beign the first time anyone has derived a
% derivative of an intractable function.

\citet{kharitonov_variational_2018} addressed theoretical issues with improper prior
$\pi$ in Sparse VD, emphasized by \citet{hron_variational_2018}, and proposed
\emph{Automatic Relevance Determination} Variational Dropout, by replacing $\pi(\omega_j)$
with a proper Gaussian prior $
  \pi_\lambda(\omega_j) = \mathcal{N}(\omega_j \vert\, 0, \tau_j^{-1})
$ with learnable precision $\tau_j > 0$ \citep{neal_bayesian_1996}. This recast the VD
as the \emph{Empirical Bayes} approach, which performs Bayesian Inference over $\omega$,
but uses Maximum Likelihood estimates for the hyper-parameters $\lambda$,
\citep{mackay_bayesian_1994}. Maximizing \eqref{eq:sgvb_estimator} over $\tau$, holding
other parameters fixed, yields $
  \tau^*_j = {(\mu_j^2 + \sigma^2_j)}^{-1}
$, whence
\begin{equation}  \label{eq:ard-kl-div-real}
  K(\alpha)
    = \frac12 \log{\bigl(1 + \tfrac1{\alpha} \bigr)}
    \,.
\end{equation}
% \begin{align}
%   \pi(\omega)
%     &
%     = \prod_J \pi(s_J)
%       \prod_{j\in J} \pi(\omega_j \vert s_J)
%     \propto \prod_J \frac1{\lvert s_J \rvert}
%       \prod_{j\in J} \mathcal{N}(\omega_j \vert 0, s_J^2)
%     \,, \\
%   q_\theta(\omega)
%     &
%     % = \prod_J q_\theta(s_J)
%     %   \prod_{j\in J} q_\theta(\omega_j \vert s_J)
%     = \prod_J \mathcal{N}(s_J \vert \mu_J, \alpha_J \mu_J^2)
%       \prod_{j\in J} \mathcal{N}(\omega_j \vert \mu_j s_J, \sigma_j^2 s_J^2)
%     \,.
% \end{align}

% practical difference between them?

Simultaneously the method \citet{molchanov_variational_2017} proposed to use \emph{additive
noise parameterization} in the factorized Gaussian $q_\theta(\omega)$ in conjunction with
the local reparameterization trick. They reverted the $(\mu, \alpha)$ parameterization in
$q_\theta(\omega)$ back to $(\mu, \sigma^2)$, arguing that it reduces the variance of the
SGVB \eqref{eq:sgvb_estimator}, by rendering the gradient with respect to $\mu$ independent
from the local noise, injected by \eqref{eq:r-gauss-trick}. This modification is important
for pruning, since $\mu$ of a relevant parameter serves as the estimate of its value.

% subsection dropout (end)

% section variational_dropout (end)

\section{$\cplx$-valued networks} % (fold)
\label{sec:c_valued_networks}

% implementation
$\cplx$-valued neural networks are networks that rely on the arithmetic in the complex
domain. To achieve this implementations of $\cplx$VNN use the geometric representation
of a complex number as paired \emph{real} and \emph{imaginary} values, $\cplx \simeq \real^2$,
ensuring that the resulting $\real$-valued computational graph respects $\cplx$-arithmetic.
For example, $
  f\colon \cplx^n \to \cplx^m
$ is identified with a real vector-valued function $
  F\colon \real^{2 m} \to \real^{2 m}
$ defined via $
  F(u, v) = (\Re f(u + \iu v), \Im f(u + \iu v))
$, $\Re$ and $\Im$ denoting the real and imaginary parts, respectively. When $f$ is
a $\cplx$-valued linear transformation, the computations are ``wired'' so that
\begin{equation}  \label{eq:cplx-lin-op}
  % F\colon
  %   (u, v)
  %     \mapsto \Bigl(
  %       P u - Q v,
  %       P v + Q u
  %     \Bigr)
  F(u, v)
    =
    \begin{pmatrix}
      P u - Q v \\
      P v + Q u
    \end{pmatrix}
    = \begin{pmatrix}
      P & - Q \\ Q & P
    \end{pmatrix} \begin{pmatrix}
      u \\ v
    \end{pmatrix}
    \,,
\end{equation}
with $
  P, Q \colon \real^{n} \to \real^{m}
$ given by $P = \Re f$ and $Q = \Im f$ restricted to $\real^{n}$.
% The downside is that, for instance, one $\cplx$-valued convolution requires up to
% four $\real$-valued convolution operations.
%
Non-linearities in $\cplx$VNN can be hyperbolic functions or maps that operate on
$\cplx$ numbers in planar form, $
  z \mapsto \sigma(\Re{z}) + \iu \sigma(\Im{z}) % split
$, or polar form $
  % z \mapsto \sigma(\lvert z\rvert, \arg{\!(z)})
  r e^{\iu \phi} \mapsto \sigma(r, \phi)
$.
% \citep{savitha_2011,hirose_generalization_2012,arjovsky_unitary_2016,guberman_complex_2016}

This $\cplx \simeq \real^2$ identification allows straightforward retrofitting of $\cplx$VNN
into existing $\real$-valued auto-differentiation frameworks for deep learning. This
act is backed by Wirtinger ($\cplx\real$) calculus, which enables generalized treatment
of functions of complex argument, by regarding $z$ and its \emph{complex conjugate}
$\conj{z}$ as independent variables and defining derivative operators with respect to
them through partial derivatives with respect to real and imaginary parts. These definitions
simplify manual analysis of $\cplx$ derivatives and satisfy the product and chain rules, respect
complex conjugation and linearity for $\cplx \to \cplx$ maps, and as such were used to define
$\cplx$ version of back-propagation, \citep{benvenuto_complex_1992,guberman_complex_2016}.
However, since auto-differentiation frameworks can algorithmically handle computational
graphs of arbitrary complexity, explicit use of Wirtinger derivatives is not required,
especially considering the fact that the direction of the steepest ascent of a $
  \cplx \to \real
$ function is given by complex conjugate gradient $\nabla_{\conj{z}}$, which coincides
with the classical gradient of the same function viewed as $\real^2 \to \real$,
(see appendix~\ref{sub:wirtinger_calculus}).
%
% in natively $\cplx$-valued backprop it would be necessary to modify the existing sgd
% to explicitly use $\nabla_{\conj{z}} f(z)$, while no such modification is necessary
% in re-im. If some inner activation of a $\cplx$VNN is holomorphic, i.e. satisfies
% Cauchy-Riemann conditions, then that particular step of in the chain rule that is
% backprop can be simplified by assuming one piece of the tracked gradient to be exact
% zero. But other wise it is necessary to track both $\partial_{\conj{z}}$ and $\partial_{z}$.
%
% Training a $\cplx$VNN involves optimizing a real-valued objective with respect to
% complex valued arguments.
%
% The computed gradients are theoretically justified
% - CR dictates the it is necessary to track both $z$ and $\conj{z}$ derivatives
% - the direction of largest increase in f: C-R is given by the gradient with
% respect to the conjugate, which coincides in the re-im representation with 
% the ordinary gradient wrt u and v
% - the Chain Rule in CR holds and when a holomorphic C-C function is backproped
% the autodiff just wastes cycles on computing its conj-derivative (rather than assuming
% it to be zero)

% Although $\cplx$-valued dense layers and back-propagation have been around since early
% 1990-s, one of the earliest accounts
Development of deep $\cplx$-valued networks has been active. \citet{haensch_complex-valued_2010}
put forward $\cplx$-valued convolutional networks, \citet{guberman_complex_2016} and
\citet{popa_complex-valued_2017} developed modifications of pooling, \citet{arjovsky_unitary_2016}
and \citet{wisdom_full-capacity_2016} proposed $\cplx$-valued RNNs with unitary recurrent
transition matrices, and \citet{danihelka_associative_2016} developed $\cplx$-valued
holographic representations for LSTMs. More recently \citet{trabelsi_deep_2018} proposed
$\cplx$-valued batch-normalization and weight initialization, \citet{wolter_complex_2018}
investigated different $\cplx$-valued gating mechanisms for RNNs, and \citet{yang_complex_2020}
proposed $\cplx$-valued self-attention and complex transformer architecture. It merits
noting that \citet{gaudet_deep_2018} generalized $\cplx$VNN further to deep quaternion-valued
networks, and \citet{vecchi_compressing_2020} studied sparsity inducing regularizers for
them.
% complex-valued distributions \citep{pav_moments_2015,taubock_complex-valued_2012},
% and \citep{karseras_caution:_2014}

% This is too abrupt: first intro into what Cplx- valued nets are, then the representation 
% and how it dictates the architecture, and ultimately bioils down to representation
% u(x, y) + j v(x, y) as a R2->R2 map.

% \todo{read on holomorphic nets}

% section c_valued_networks (end)

\section{$\cplx$-Variational Dropout} % (fold)
\label{sec:c_variational_dropout}

In this section we develop Sparse Variational Dropout for $\cplx$VNN by using a fully
factorized complex Gaussian posterior approximation. We outline the $\cplx$ version of
the local reparameterization trick and derive the divergence penalties in \eqref{eq:sgvb_estimator}.
The proposed $\cplx$-valued extension can readily be a part of a hierarchical variational
approximation for structured sparsity \citep{louizos_bayesian_2017}.

\subsection{$\cplx$-Gaussian Distribution} % (fold)
\label{sub:c_gauss_and_local_rep}

A vector $z\in \cplx^m$ has complex Gaussian distribution, $
  q(z) = \mathcal{C}\mathcal{N}_m(\mu, \Gamma, C)
$ with mean $\mu \in \cplx^m$, complex covariance and relation matrices $\Gamma$ and $C$,
respectively, if
\begin{equation}  \label{eq:cn-paired-real-density}
  \begin{pmatrix}
    \Re z \\ \Im z
  \end{pmatrix}
    \sim \mathcal{N}_{2 m}\biggl(
      \Bigl(
        \begin{smallmatrix}
          \Re \mu \\ \Im \mu
        \end{smallmatrix}
      \Bigr),
      \tfrac12 \Bigl(
        \begin{smallmatrix}
          \Re{(\Gamma + C)} & \Im{(C - \Gamma)} \\
          \Im{(\Gamma + C)} & \Re{(\Gamma - C)}
        \end{smallmatrix}
      \Bigr)
    \biggr)
    \,,
\end{equation}
provided $\Gamma$ is positive definite Hermitian matrix, $C^\top = C$, and $
  \conj{\Gamma} \succeq \conj{C} \Gamma^{-1} C
$.
% $\Gamma \succeq 0$, $\conj{\Gamma^\top} = \Gamma$
% notation
% Here, $M^{\top}$ is the matrix transpose, $\conj{M}$ is elementwise complex
% conjugation, and $M^{\hop} = (\conj{M})^\top$ denotes Hermitian conjugate.
%
Matrices $\Gamma$ and $C$ are given by $
  \mathbb{E} (z - \mu)\conj{(z - \mu)^\top}
$ and $
  \mathbb{E} (z - \mu)(z - \mu)^{\top}
$, respectively, and the random vector $z$ is a \emph{circularly symmetric} $\cplx$-Gaussian
vector if $z$ and $\conj{z}$ are uncorrelated, i.e. $C = 0$.
The entropy of $z$ terms of $\Gamma$ and $C$ is
\begin{eqnarray} \label{eq:c-gauss-entropy}
  \mathbb{H}(q)
    &=&
      - \mathbb{E}_{z \sim q} \log{q(z)}
      \\
    &=&
      \tfrac12 \log \det{(\pi e \Gamma)}
      \det{(\pi e (\conj{\Gamma} - \conj{C} \Gamma^{-1} C))}
    \notag \\
    % determinant commutes with complex conjugation, since it is multilinear
    % = \tfrac12 \log \det{(\pi e \Gamma)} \det{(\pi e (\conj{\Gamma} - \conj{C} \Gamma^{-1} C))}
    % = \tfrac12 \log \det{(\pi e \Gamma)} \det{(\pi e \conj{\Gamma})}
    % = \tfrac12 \log \det{(\pi e \Gamma)} \conj{\det{(\pi e \Gamma)}}
    % = \tfrac12 \log \bigl\lvert \det{(\pi e \Gamma)} \bigr\rvert^2
    &=&
      \log \bigl\lvert \det{(\pi e \Gamma)} \bigr\rvert
      \,, \quad \text{ for } C = 0
    \,.
    \notag
\end{eqnarray}
%
% The roles of $\Gamma$ and $C$ are more evident in univariate case: here $
%   \Gamma = \sigma^2 \geq 0
% $, and $C = \sigma^2 \rho \in \cplx$ with $
%   \lvert \rho \rvert \leq 1
% $. $\sigma^2$ determines the ``dispersion'', whereas $\rho$ is determines the
% ``strength'' and ``direction'' of the linear relation between real and imaginary
% parts \citep{lapidoth_capacity_2003}.
% $$
% \frac{\sigma^2}2
%   \begin{pmatrix}
%     1 + \Re{\rho} & \Im{\rho} \\
%     \Im{\rho} & 1 - \Re{\rho}
%   \end{pmatrix}
%   = \frac{\sigma^2}2 
%       \begin{pmatrix}
%         1 + \lvert \rho \rvert \cos\theta
%           & \lvert \rho \rvert \sin\theta \\
%         \lvert \rho \rvert \sin\theta
%           & 1 - \lvert \rho \rvert \cos\theta
%       \end{pmatrix}
%   \,. $$
Parameterization of a univariate $\cplx$-Gaussian distribution is simpler: $
  \mathcal{CN}(\mu, \sigma^2, \sigma^2 \xi)
$ with $\xi \in \cplx$ such that $\lvert \xi \rvert \leq 1$ and $\sigma^2 \geq 0$.
By \eqref{eq:c-gauss-entropy} its entropy is $
  \log \pi e \sigma^2 \sqrt{1 - \lvert \xi \rvert^2}
$.

% $\cplx$-Gaussians are closed under affine transformations
$\cplx$-Gaussianity is preserved under linear transformations, i.e. for $
  A \in \cplx^{n \times m}
$ and $b \in \cplx^{n}$
\begin{equation}  \label{eq:cn-affine}
  b + A z \sim \mathcal{C}\mathcal{N}_n\bigl(
      b + A\mu, A \Gamma \conj{A^\top}, A C A^\top
    \bigr)
  \,.
\end{equation}
Therefore, if we have a $\cplx^{n\times m}$ matrix $W$ with \emph{independent}
$\cplx$-Gaussian entries, i.e.
\begin{equation}  \label{eq:c-gauss-vi-general}
  q(W)
    = \prod_{ij} \mathcal{CN}(
      % W_{ij} \mid
      \mu_{ij}, \Sigma_{ij}, \Sigma_{ij} \xi_{ij}
    )
  \,,
\end{equation}
with $
  \mu, \xi \in \cplx^{n\times m}
$, $\Sigma \in [0, +\infty)^{n\times m}$ and $
  \lvert \xi_{ij} \rvert \leq 1
$, then for $x \in \cplx^m$ and $b \in \cplx^n$ each component $y_i$ of $y = b + W x$
is independent univariate $\cplx$-Gaussian
\begin{equation}  \label{eq:cplx-gauss-trick}
  y_i
    \sim \mathcal{C}\mathcal{N}
      \Bigl(
        b_i + \sum_{j=1}^m \mu_{ij} x_j,
        \, \sum_{j=1}^m \Sigma_{ij} \lvert x_j \rvert^2,
        \, \sum_{j=1}^m \Sigma_{ij} x_j^2 \xi_{ij}
      \Bigr)
    \,.
\end{equation}
This is the $\cplx$-Gaussian version of the local reparameterization trick \eqref{eq:r-gauss-trick}.
It requires three matrix-vector operations: $\cplx$-valued $b + \mu x$ and $C x^2$,
and $\real$-valued $
  \Sigma \lvert x \rvert^2
$, where $C_{ij} = \Sigma_{ij} \xi_{ij}$ and the complex modulus and square are applied
elementwise. \eqref{sub:c_gauss_and_local_rep} can be applied to any layer, the output
of which depends linearly on its parameters, such as convolutional, affine, and bilinear
transformations ($
  (x, z) \mapsto b_j + x^\top W^{(j)} z
$). Similar to the $\real$ case, $\cplx$ convolutions draw independent realizations of $W$
for each spatial patch in the input \citep{molchanov_variational_2017}. This provides
faster computations and better statistical efficiency of the SGVB gradient estimator
by eliminating correlation from overlapping patches \citep{kingma_variational_2015} and
allowing \eqref{eq:cplx-gauss-trick} to efficiently leverage $\cplx$ convolutions of
the relation and variance kernels with elementwise complex squares $x^2$ and amplitudes
$\lvert x \rvert^2$.
% Weight sharing is essentially pushed to shared variational parameters $\theta$ in the approximation $q_\theta$.
% A convolution is a matrix product of $W$, unrolled into a toeplitz matrix, and im-to-col
% flattened intput $x$

For $\cplx$-Sparse Variational Dropout we propose to use fully factorized $\cplx$-Gaussian
approximation \eqref{eq:c-gauss-vi-general} with $\xi_{ij} = 0$ and additive noise
parameterization ($
  \alpha_{ij} = \tfrac{\Sigma_{ij}}{\lvert \mu_{ij} \rvert^2}
$) for weights in dense linear, convolutional and other effectively parameter-affine
layers. Point estimates are used for biases.
% This forces the relation coefficient in \eqref{eq:cplx-gauss-trick} to be zero. 

% subsection c_gauss_and_local_rep (end)

\subsection{The priors} % (fold)
\label{sub:the_priors}

For a fully factorized approximation $q(\omega)$ and factorized prior $
  \pi(\omega) = \prod_{ij} \pi(\omega_{ij})
$, the divergence term \eqref{eq:sgvb_estimator} is
\begin{equation}  \label{eq:elbo-general-kl-div}
  KL(q \| \pi)
    % = \mathbb{E}_{\oemga \sim q(\oemga)} \log \tfrac{q(\oemga)}{\pi(\oemga)}
    % = \sum_{ij} \mathbb{E}_{\omega_{ij} \sim q(\omega_{ij})}
    %   \log \tfrac{q(\omega_{ij})}{\pi(\omega_{ij})}
    % = \sum_{ij} KL\bigl( q(\omega_{ij}) \| \pi(\omega_{ij}) \bigr)
    = - \sum_{ij}
        \mathbb{H}(q(\omega_{ij}))
        + \mathbb{E}_{q(\omega_{ij})} \log{\pi(\omega_{ij})}
    \,.
\end{equation}
We consider two fully factorized priors: an improper prior, resembling VD, and $\cplx$-Gaussian
ARD prior. We omit subscripts ${ij}$ for brevity in this section.

\subsubsection{VD prior} % (fold)
\label{ssub:vd_prior}

From \eqref{eq:c-gauss-entropy} and $\xi = 0$ the KL-divergence for an improper prior $
  \pi(\omega) \propto {\lvert \omega \rvert}^{-\beta}
$ with $\beta \geq 1$ is
\begin{equation}  \label{eq:c-vd-kl-div-raw}
  KL(q\| \pi)
    \propto
    %   \mathbb{E}_{\omega \sim q(\omega)} \log q(\omega)
    %   + \tfrac{\beta}2 \mathbb{E}_{\omega \sim q(\omega)} \log \lvert \omega \rvert^2
    % =
    %   - \log \lvert \det{(\pi e \sigma^2)} \rvert
    %   + \tfrac{\beta}2 \mathbb{E}_{\omega \sim q(\omega)} \log \lvert \omega \rvert^2
    % =
      - \log{\sigma^2}
      + \tfrac{\beta}2 \Bigl(
        \mathbb{E}_{\omega \sim q(\omega)} \log \lvert \omega \rvert^2
      \Bigr)
    \,.
\end{equation}
For $\mu \neq 0$ and $
  \sigma^2 = \alpha \lvert \mu \rvert^2
$ property \eqref{eq:cn-affine} implies $
  \mathcal{CN}(\mu, \sigma^2, 0)
    \sim \mu \cdot \mathcal{CN}(1, \alpha, 0)
$, whence the expectation in brackets is given by
\begin{equation}  \label{eq:expect-improper-term-cplx}
  % \mathbb{E}_{\omega \sim q(\omega)} \log \lvert \omega \rvert^2
  %   % = \mathbb{E}_{\varepsilon \sim \mathcal{CN}(\varepsilon \vert 1, \alpha, 0)}
  %   %   \log \lvert \mu \varepsilon \rvert^2
  %   % = \log \lvert \mu \rvert^2
  %   %   + \mathbb{E}_{\varepsilon \sim \mathcal{CN}(\varepsilon \vert 1, \alpha, 0)}
  %   %       \log \lvert \varepsilon \rvert^2
  %   =
    \log \alpha \lvert \mu \rvert^2
      + \mathbb{E}_{\varepsilon \sim \mathcal{CN}(0, 1, 0)}
          \log{\bigl\lvert \tfrac1{\sqrt{\alpha}} + \varepsilon \bigr\rvert^2}
    \,.
\end{equation}
If $
  (z_i)_{i=1}^m
    \sim \mathcal{CN}(0, 1, 0)
$ iid and $\theta \in \cplx^m$, then $
  \sum_i \lvert \theta_i + z_i \rvert^2
    \sim \chi^2_{2m}(s^2)
$ with $
  s^2 = \sum_i \lvert \theta_i \rvert^2
$, i.e. a non-central $\chi^2_{2m}$ with parameter $s^2$. Its log-moments for general
integer $m \geq1$ have been derived by \citet[p.~2466]{lapidoth_capacity_2003}.
% in fact Appendix X, Lemma 10.1
In particular, for $m=1$ and $\theta\in \cplx$ we have
\begin{equation}  \label{eq:log-moment-for-chi-2}
  \mathbb{E}_{z \sim \mathcal{CN}(0, 1, 0)}
    \log \lvert \theta + z \rvert^2
    = \log \lvert \theta \rvert^2 - \mathop{Ei}( - \lvert \theta \rvert^2)
    \,,
\end{equation}
where $
  \mathop{Ei}(x) = \int^x_{-\infty} t^{-1} e^t dt
$ for $x < 0$ is the Exponential Integral, which satisfies $
  \mathop{Ei}(x) \leq \log{(-x)}
$, $
  % as x to 0 from below
  \mathop{Ei}(x) \approx \log{(-x)} - \gamma
$ as $x\to 0$ ($\gamma$ is Euler's constant) and $
  \mathop{Ei}(x) \geq -e^x
$ for $x \leq -1$.
% 1. by Leibniz rule $Ei(x)$ has -ve derivative on $x < 0$
% 2. $Ei(x) \leq 0$ on $x < 0$, since $t^{-1} e^t \leq t^{-1} < 0$ on $t < 0$
% 3. $Ei(x) \leq \log{(-x)}$ on $x < 0$, since on $x \in [-1, 0)$
% $$
%   \mathop{Ei}(x)
%     = \mathop{Ei}(-1) + \int_{-1}^x t^{-1} e^t dt
%     \leq \mathop{Ei}(-1) + \int_{-1}^x t^{-1} dt
%     = \mathop{Ei}(-1) + \log{(-x)}
%     \leq \log{(-x)}
%   \,. $$
% 3. by l'H{\^o}pital rule $Ei(x)$ is asymptotically $\log{-x}$ as $x \to 0-$
% $$
% % t = -e^{-x} ,  -\log{(-t)} = x
%   \lim_{x\to 0-} \frac{\mathop{Ei}(x)}{\log{(-x)}}
%     = \lim_{x\to 0-} \frac{x^{-1} e^x}{x^{-1}}
%     = 1
%   \,. $$
% 4. \citep{lapidoth_capacity_2003} yields $\lim_{x\to 0-} \log{(-x)} - \mathop{Ei}(x) = \gamma$
% 5. $Ei(x) \geq -e^x$ for any $x < -1$, since $t^{-1} e^t \geq - e^t$ for $t < -1$
% $$
% \mathop{Ei}(x)
%   = \int_{-\infty}^x t^{-1} e^t dt
%   \geq \int_{-\infty}^x -e^t dt = -e^x
%   \,. $$
% 6. $Ei(x) \leq \log{(-x)} - \gamma$ on $x < 0$ and $-e^x \leq Ei(x) \leq 0$ on $x < -1$.
%
Although $\mathop{Ei}$ is an intractable integral, requiring numerical approximations to
compute, its derivative is exact: $
  \tfrac{d}{dx} \mathop{Ei}(x) = \tfrac{e^x}{x}
$ at $x < 0$.

From \eqref{eq:expect-improper-term-cplx} and \eqref{eq:log-moment-for-chi-2}, the terms of
the divergence that depend on the parameters are given by
\begin{equation}  \label{eq:c-vd-kl-div}
  KL(q\| \pi)
    \propto
    %   - \log \lvert \det{(\pi e \sigma^2)} \rvert
    %   + \tfrac{\beta}2 \log \alpha \lvert \mu \rvert^2
    %   + \tfrac{\beta}2 \mathbb{E}_{\varepsilon \sim \mathcal{CN}(\varepsilon \vert 0, 1, 0)}
    %       \log{\bigl\lvert \tfrac1{\sqrt{\alpha}} + \varepsilon \bigr\rvert^2}
    % =
    %   - \log{\sigma^2}  % - \log{\pi e}
    %   + \tfrac{\beta}2 \log \alpha \lvert \mu \rvert^2
    %   + \tfrac{\beta}2 \bigl(
    %     - \log \alpha - \mathop{Ei}( -\tfrac1{\alpha})
    %   \bigr)
    % =
    %   - \log{\sigma^2}
    %   + \tfrac{\beta}2 \log \alpha \lvert \mu \rvert^2
    %   - \tfrac{\beta}2 \log \alpha
    %   - \tfrac{\beta}2 \mathop{Ei}( -\tfrac1{\alpha})
    % =
      % \log \tfrac{\lvert \mu \rvert^\beta}{\sigma^2}
      % \log \tfrac{\lvert \mu \rvert^\beta}{\alpha \lvert \mu \rvert^2}
      \tfrac{\beta-2}2 \log{\lvert \mu \rvert^2}
      + \log{\tfrac1{\alpha}}
      - \tfrac{\beta}2 \mathop{Ei}(- \tfrac1{\alpha})
      \,.
      \tag{\ref{eq:c-vd-kl-div-raw}'}
\end{equation}
We set $\beta = 2$ to make the divergence term depend only on $\alpha$ and add $\gamma$
so that the right-hand side is non-negative \citep[eq.(84)]{lapidoth_capacity_2003}.
Since $\mathop{Ei}(x)$ has simple analytic derivative and \eqref{eq:sgvb_estimator} depends additively
on \eqref{eq:c-vd-kl-div}, it is possible to back-propagate through the divergence without
forward evaluation, which speeds up gradient updates.
% Gradient-based optimiziation of \eqref{eq:sgvb_estimator} usually does not use the value
% of the objective.  In stochastic gradient methods are concerned more with the gradient,
% induced by the objective function, rather than its value.

% subsubsection variational_dropout (end)

\subsubsection{ARD prior} % (fold)
\label{ssub:ard_prior}

We consider the fully factorized circularly symmetric $\cplx$-Gaussian ARD prior $
  \pi_\tau(\omega)
    = \mathcal{CN}\bigl(
      \omega \vert 0, \tau^{-1}, 0
    \bigr)
$ with $\tau > 0$. The per element divergence term in \eqref{eq:elbo-general-kl-div} is
\begin{equation}  \label{eq:emp-bayes-kl-div}
  KL(q \| \pi_\tau)
    = - 1 - \log{(\tau \sigma^2)}
      + \tau \bigl(
        \sigma^2 + \lvert \mu \rvert^2
      \bigr)
    \,.
\end{equation}
%
% This follows from the KL-divergence expression for two multivariate Gaussians
% $$
% % \mathbb{E}_{\omega\sim q_1(\omega)} \log \tfrac{q_1(\omega)}{q_2(\omega)}
% %   =
% KL(q_1\| q_2)
%   =
%   % - \tfrac12 \log\det{(2\pi e \Sigma_1)} + \tfrac12 \log\det{(2\pi \Sigma_2)}
%   - \tfrac12 \log \det{e I}
%   + \tfrac12 \log \frac{\det{\Sigma_2}}{\det{\Sigma_1}}
%   + \tfrac12 \tr{\bigl( \Sigma_2^{-1} \Sigma_1 \bigr)}
%   + \tfrac12 (\mu_1 - \mu_2)^\top \Sigma_2^{-1} (\mu_1 - \mu_2)
%   =
%   - \log\lvert\det{\pi e \Gamma}\rvert
%   + \log\lvert\det{\pi \tau^{-1} I} \rvert
%   + \tau \Re \tr{\Gamma} + \tau \mu^\hop \mu
%   =
%   - n - \log \lvert\det{\tau \Gamma}\rvert
%   + \tau \Re \tr{\Gamma} + \tau \mu^\hop \mu
%   \,. $$
%
In Empirical Bayes the prior adapts to the observed data, i.e. \eqref{eq:sgvb_estimator}
is optimized w.r.t. $\tau$ of each weight's prior. The Maximum Likelihood estimator
of $\tau$ is given by the minimizer \eqref{eq:emp-bayes-kl-div}, i.e. $
  \tau^\ast = (\sigma^2 + \lvert \mu \rvert^2)^{-1}
$, thereby giving
\begin{equation}  \label{eq:emp-bayes-opt-kl}
  KL(q \| \pi_{\tau^\ast})
    % = - 1 - \log{({\tau^\ast} \sigma^2)}
    %   + {\tau^\ast} \sigma^2 + {\tau^\ast} \lvert \mu \rvert^2
    % = \log{((\sigma^2 + \lvert \mu \rvert^2) \tfrac1{\sigma^2})}
    = \log{\bigl(1 + \tfrac{\lvert \mu \rvert^2}{\sigma^2}\bigr)}
    = \log{\bigl(1 + \tfrac1\alpha \bigr)}
    \,.
    \tag{\ref{eq:emp-bayes-kl-div}'}
\end{equation}
Thus in both $\real$ and $\cplx$ cases ARD produces a tractable analytic expression for
the KL-divergence term in \eqref{eq:sgvb_estimator}.

% subsubsection empirical_bayes (end)

\subsubsection{$\cplx$-Variational Dropout via $\real$-scaling} % (fold)
\label{ssub:real_scaling_dropout}

We consider the following parameterization of $W$: $
  W_{ij} = \mu_{ij} \varepsilon_{ij}
$, $\varepsilon_{ij} \in \real$ with $
  \varepsilon_{ij} \sim \mathcal{N}(1, \alpha_{ij})
$, yet $\mu \in \cplx^{n \times m}$. This case corresponds to inference regarding
multiplicative noise $\varepsilon$ rather than the parameters themselves. Under
this parameterization $q(W_{ij})$ is effectively degenerate univariate $\cplx$-Gaussian
\eqref{eq:c-gauss-vi-general} with $
  \Sigma_{ij} = \alpha_{ij} \lvert \mu_{ij} \rvert^2
$ and $
  \xi_{ij} = e^{\iu \phi_{ij}}
  % \xi_{ij} = \bigl(\tfrac{\mu_{ij}}{\lvert \mu_{ij} \rvert}\bigr)^2
$ with $\phi_{ij} = \arg \mu_{ij}$, thereby making the complex relation parameter
in \eqref{eq:cplx-gauss-trick} equal $
  \sum_j \alpha_{ij} (x_{ij} \mu_{ij})^2
$, which is non-zero.
The KL-divergence term coincides with \eqref{eq:improper-kl-div-real}, however the major
drawback of this approximation is that the gradient of the loss with respect to $\mu$ cannot
be disentangled from the local output noise by additive reparameterization.

% subsubsection real_scaling_dropout (end)

% subsection priors_and_kullback_leibler_divergence (end)

% section c_variational_dropout (end)

\section{Experiments} % (fold)
\label{sec:experiments}

To verify the proposed $\cplx$-valued variational sparsification methods presented
above and explore their compression-performance trade-off we carry out a numerical
study of $\cplx$VNN for image classification and music transcription.

% image features
Since image data is not naturally $\cplx$-valued, we preprocess it using the natural
inclusion $\real \hookrightarrow \cplx$ (\texttt{raw}, $\Im z = 0$) or applying the
two-dimensional Fourier Transform (\texttt{fft}), centering the lower frequencies. We
do not train an auxiliary network that synthesizes the imaginary component from $\real$
input data \citep{trabelsi_deep_2018}.
Following \citet{wolter_complex_2018} and \citet{trabelsi_deep_2018}, the class logit
scores are taken as real part of the complex-valued output of a network.

% setup and common experiment settings
The networks are trained in three successive stages in every experiment: the \emph{``pre-train''}
stage for pre-training the network, the \emph{``sparsify''} stage to determine parameter
relevance using Variational Dropout, and the \emph{``fine-tune''} to train the pruned
network (sec.~\ref{sub:staging}). Network's parameters are initialized with values from
the previous stage.
Networks are trained with ADAM optimizer, with the learning rate reset to $10^{-3}$ before
each stage and global $\ell_2$-norm gradient clipping at $0.5$.
% https://pytorch.org/docs/master/_modules/torch/nn/utils/clip_grad.html#clip_grad_norm_

Each experiment is replicated \emph{five} times to account for random effects from
initialization, stochastic gradient optimization, noisy output from intermediate layers,
and non-determinism of computations on GPU.
% experiments took about three weeks on 4 node GTX1080Ti GPUs in fp32

% write a note on the incompressible parameters (bias, batch norm etc.)
The compression rate is calculated based on the number of floating point values needed
to store the network and equals $
  \tfrac{n_\mathtt{par}}{n_\mathtt{par} - n_\mathtt{zer}}
$, where $n_\mathtt{zer}$ is the number of explicit zeros at the ``fine-tune'' stage
and $n_\mathtt{par}$ is the total number of values. In a $\real$-valued network each
parameter counts as \emph{one} value and as \emph{two} values in a $\cplx$VNN. Each
model has a compression limit, determined by biases, shift and scaling in $\real$-
and $\cplx$-valued batch normalization layers.
% variace parameters in q_\theta are completely ignored.

\subsection{Stagewise training} % (fold)
\label{sub:staging}

% The training consists of three successive stages: ``pre-train'' $\to$ ``sparsify'' $\to$ ``fine-tune''.
At the \emph{``pre-train''} stage every network is fit ``as-is'' using deterministic layers
and only the likelihood term from \eqref{eq:sgvb_estimator}.

During the \emph{``sparsify''} stage we make every layer stochastic and apply variational
sparsification (sec.~\ref{ssub:vd_prior}, ~\ref{ssub:ard_prior}, or their $\real$ versions).
We inject a coefficient $
  C \in (0, 1]
$ at the KL divergence term in \eqref{eq:sgvb_estimator}:
\begin{equation}  \label{eq:elbo_with_coef}
  - \frac{C}N KL(q_{\theta} \| \pi_{\lambda})
  + \frac1{M} \sum_{k=1}^M
      \log p_{\phi}(x_{i_k} \mid g(\varepsilon_k; \theta))
  \,.
  \tag{\ref{eq:sgvb_estimator}'}
\end{equation}
In contrast to \cite{molchanov_variational_2017}, who anneal $C$ from zero to one during
training, we use constant $C$ and vary it between runs. This allows us to explore the
compression-performance profile by balancing model's likelihood and posterior's penalty
for diverging form the sparsifying prior in \eqref{eq:elbo_with_coef}. In particular,
higher $C$ implies higher sparsity.
% RFC: citation needed fot this "trick". Cite $\beta$-VAE maybe?

% on compresssion
Between \emph{``sparsify''} and \emph{``fine-tune''} stages we compute masks of non-zero
weights in each layer based on the relevance scores $\alpha$ (sec.~\ref{sub:dropout}).
Since $q_\theta$ factorizes into univariate distributions, a $\cplx$ or $\real$ parameter
is considered non-zero iff $\log \alpha \leq \tau$ for $
  \alpha = \tfrac{\sigma^2}{\lvert \mu \rvert^2}
$.
% on chosing the threshold
The threshold $\tau$ is picked so that the remaining non-zero parameters are within
$\delta$ relative tolerance of their mode with high probability under the approximate
posterior. For a univariate $\real$- or a circularly symmetric $\cplx$-Gaussian random
variable $w$, $
  \tfrac{k \lvert w - \mu \rvert^2}
        {\alpha \lvert \mu \rvert^2}
$ is $\chi^2_k$ distributed with $k=1$ ($\real$) or $2$ ($\cplx$).
For a tolerance $\delta = 50\%$ values $\log \alpha$ below $-2.5$ yield at least $90\%$
chance of a non-zero $\real$/$\cplx$ parameter. We pick $\tau = -\tfrac12$ to retain
parameters sufficiently concentrated around their mode and encourage higher sparsity,
at the same time being aware that $q_\theta$ is merely an approximation. In comparison,
$\tau = 3$ is commonly used as the threshold
\citep{molchanov_variational_2017,kingma_variational_2015}.

At the \emph{``fine-tune''} stage the network reverts back to deterministic architecture
and proceeds the same way as the ``pre-train'' stage, except for training only those
parameters, which are specified by sparsity masks.

% subsection fitting_and_sparsification (end)

\subsection{MNIST-like datasets} % (fold)
\label{sub:mnist_like_datasets}

We conduct a moderately sized experiment on MNIST-like datasets of $28\times 28$ greyscale
images to study the performance-compression trade-off of the proposed $\cplx$-valued Sparse
Variational Dropout: MNIST \citep{lecun_gradient-based_1998}, KMNIST \citep{clanuwat_deep_2018},
EMNIST \citep{cohen_emnist_2017} and Fashion-MNIST \citep{xiao_fashion-mnist_2017}.
%
% dataset size
We deliberately use a \emph{fixed} random subset of \emph{ten thousand} images from the train
split of each dataset to fit the networks and measure the performance with classification
\emph{accuracy} score on the usual test split.
% Limiting the size was done primarily to test the regularization strength of compression
% and to facilitate quicker experiments.

% model architectures
% (fft) C - R - C / 2
% (raw) R - C - 2 * R
We consider two simple architectures in this experiment, which have been chosen for
the purpose of illustrating the compression and understanding the effects of experiment
parameters. \emph{TwoLayerDenseModel} is a wide dense ReLU network $
  784 \to 4096 \to n_\mathtt{out}
$, and \emph{SimpleConvModel} is a ReLU net with two $2d$ $k_5 s_1$ convolutions with
filters $20 \to 50$, two $k_2 s_2$ average pooling steps, and a classifier head
$800 \to 500 \to n_\mathtt{out}$.
For each dataset we experiment with all combinations of model kinds ($\real$ or $\cplx$)
and sparsification methods (VD or ARD). To take into account potential differences in
the capacity of $\cplx$VNN we consider halving or doubling the number of features in
the intermediate layers \citep{monning_evaluation_2018}. Halved $\cplx$VNN are tagged
$\tfrac12\cplx$, and doubled $\real$-valued networks are labelled $2\real$. For \texttt{fft}
we compare $\{\real, \cplx, 2\real\}$ and for \texttt{raw} -- $\{\tfrac12\cplx, \real, \cplx\}$.

% experiment setup
Stages (sec.~\ref{sub:staging}) last for $40$, $75$ and $40$ epochs, respectively, in each
experiment. The sparsification threshold $\tau$ is fixed at $-\tfrac12$, the training batch
size is set to $128$ and the base learning rate ${10}^{-3}$ is reduced after the $10$-th
epoch to ${10}^{-4}$ at every stage.
%
% grid specs
We vary $
  C \in \{
    \tfrac32 2^{-\tfrac{k}2} \colon k=2, \cdots, 38
  \}
$ in \eqref{eq:elbo_with_coef} and repeat each experiment $5$ times to get a sample
of compression-accuracy pairs.

% a couple of plots from MNIST and Fashion-MNIST
\begin{figure}[!t]
  \centering
  \begin{subfigure}[b]{1.\columnwidth}  % imcl2019-style swears at this
    \centering
    \includegraphics[width=\columnwidth]{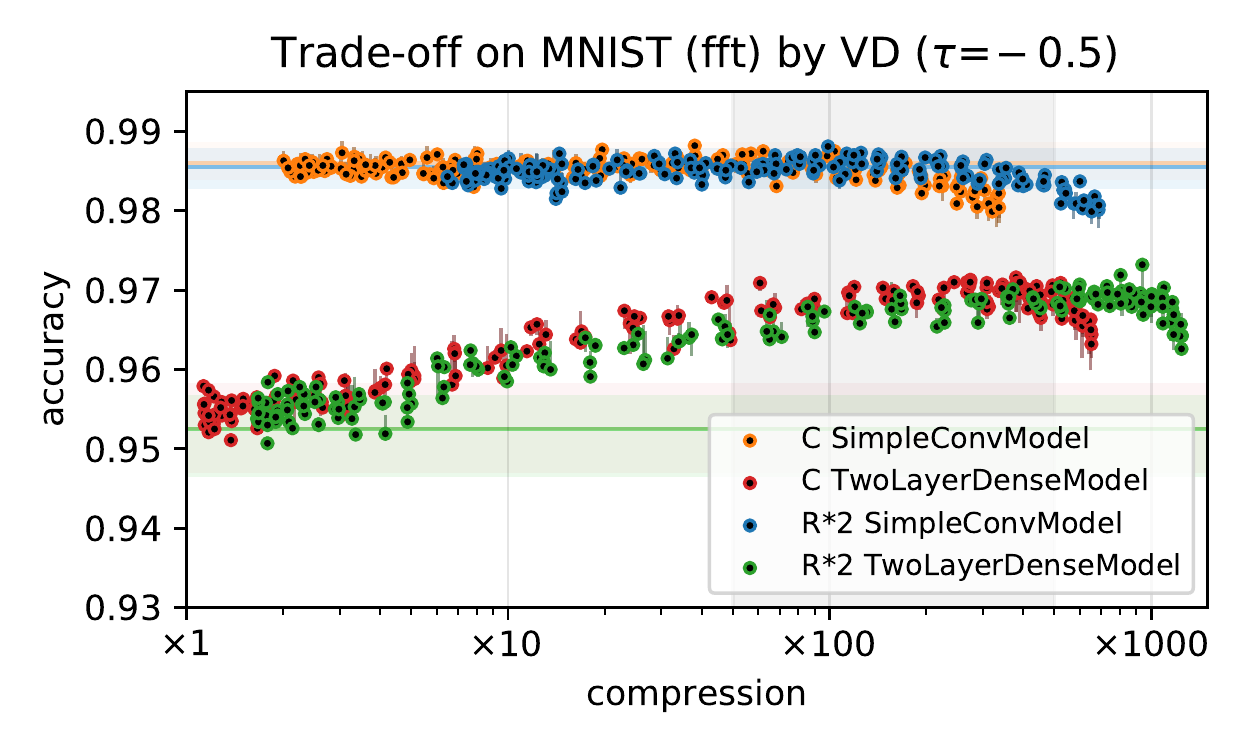}
  \end{subfigure} \\%
  % \hfill
  \begin{subfigure}[b]{1.\columnwidth}  % imcl2019-style swears at this
    \centering
    \includegraphics[width=\columnwidth]{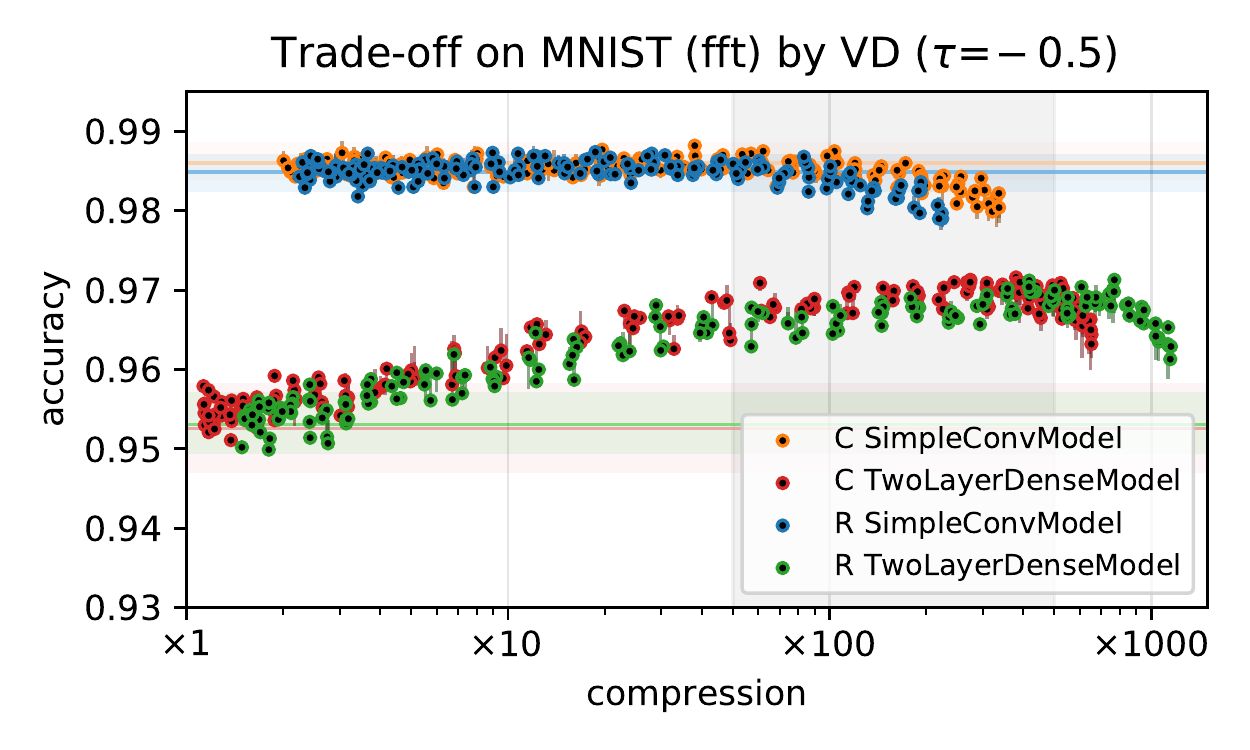}
  \end{subfigure}
  \caption{%
    The compression-accuracy curve (VD, \texttt{fft}, MNIST):
    $2 \real / \cplx$ (\textit{top}) and $\real / \cplx$ (\textit{bottom}).
  }
  \label{fig:mnist-like__trade-off__fft}
\end{figure}

\begin{figure}[!t]
  \centering
  \begin{subfigure}[b]{1.\columnwidth}  % imcl2019-style swears at this
    \centering
    \includegraphics[width=\columnwidth]{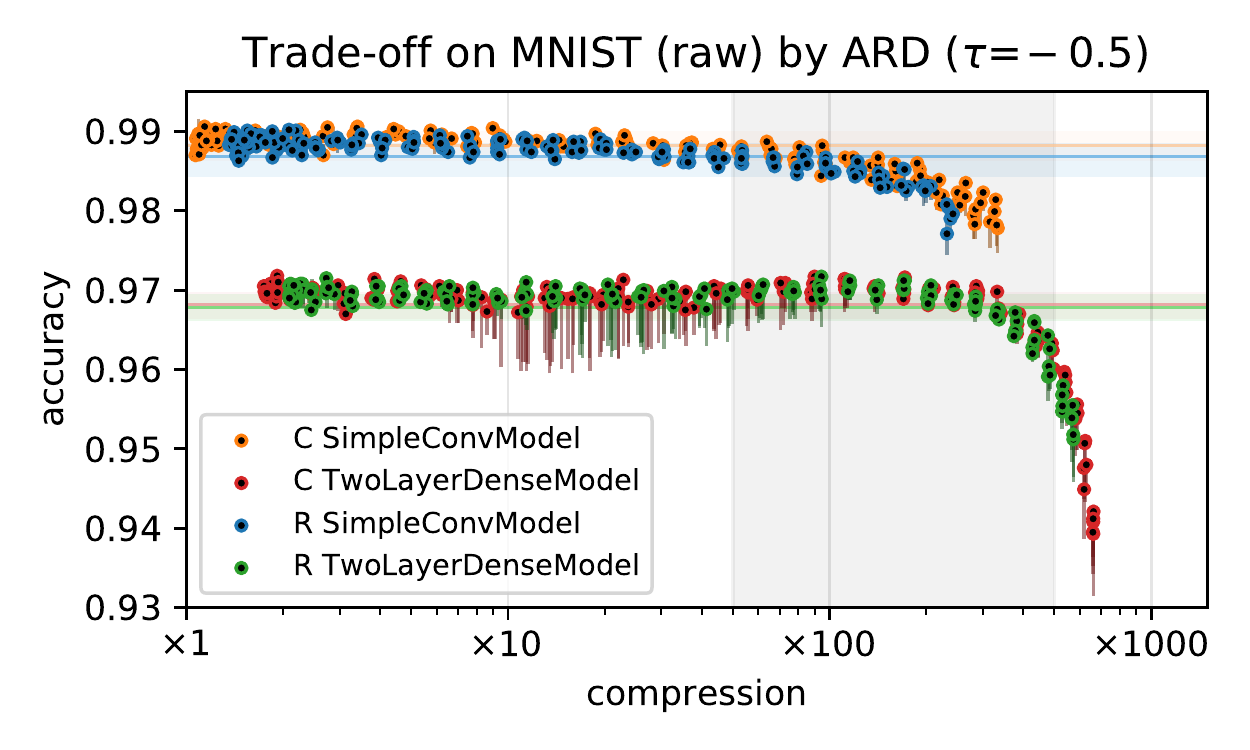}
  \end{subfigure} \\%
  % \hfill
  \begin{subfigure}[b]{1.\columnwidth}  % imcl2019-style swears at this
    \centering
    \includegraphics[width=\columnwidth]{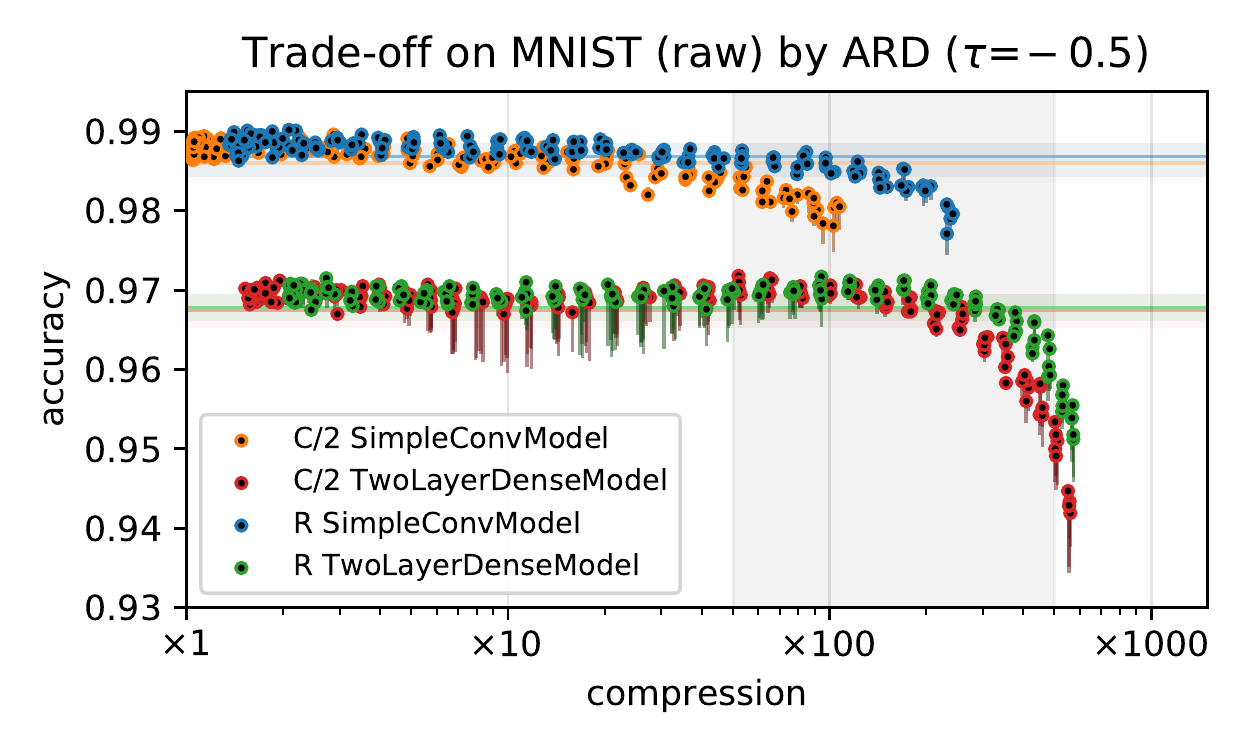}
  \end{subfigure}
  \caption{%
    The compression-accuracy curve (ARD, \texttt{raw}, MNIST):
    $\real / \cplx$ (\textit{top}) and $\real / \tfrac12 \cplx$ (\textit{bottom}).
  }
  \label{fig:mnist-like__trade-off__raw}
\end{figure}

% observations
Figures~\ref{fig:mnist-like__trade-off__fft} and~\ref{fig:mnist-like__trade-off__raw}
depict the resulting compression-accuracy trade-off on MNIST for the models described
above.
Each point represents the trade-off of the compressed network after fine-tuning, while
its tail illustrates the impact of this stage on the performance. Transparent horizontal
bands on each plot represent min-max performance spread of the pre-trained uncompressed
network on the test split.
Results for other MNIST-like dataset are presented appendix~\ref{sec:mnist_like_experiments}.

% observations and conclusions
The overarching conclusion from the conducted experiments is that both $\cplx$-ARD and
$\cplx$-VD methods compress similarly to each other, but for the same value of $C$ in
\eqref{eq:elbo_with_coef} ARD yields marginally lower compression and slightly higher
performance post fine-tuning. For each fixed $C$ the compression rates after ``sparsify''
stage are roughly identical.
% a table for mean comparison against C and compression-performance?
%
% fft: $2\real$ convolutional model trade-off overall better than $\real$, dense -- on par.
% raw: $\cplx$ models trade-off overall better than $\tfrac12\cplx$.
% fft: any-$\real$ dense model compresses better than $\cplx$
% $\cplx$-valued networks do not have as much spare capacity as $2 \real$ due to
%  constraints imposed by multiplication in $\cplx$ numbers
%
At the same time, ``fine-tune'' stage almost always improves performance in high compression
regime ($\times50+$ high $C$ in \eqref{eq:elbo_with_coef}), likely due to regularization from
high sparsity.
% On EMNIST Letters and MNIST datasets in the small compression regime the dense network
% exhibits a trough in accuracy before fine-tuning stage.
Fourier features catch up to the raw data in terms of performance at high compression
rates $\times100+$ only for the \emph{TwoLayerDenseModel}.
% spectral feature useful residual information?
% For it the compression-performance profile exhibits a ramp-like shape, in contrast
% to the raw features, for which the accuracy stays level for moderate rates, but then
% abruptly drops at high rates.
%
Comparison of $\real$ and $\cplx$ networks with matching architecture, i.e. same effective
layer widths in $2\real$ vs. $\cplx$ and $\real$ vs. $\tfrac12\cplx$, shows that doubled
$\real$ networks perform and compress better than $\cplx$, due to higher intrinsic redundancy
unchecked by $\cplx$-arithmetic constraint.
% Convolutional network compress worse because they have less redundant parameters

% subsection mnist_like_datasets (end)

\subsection{CIFAR10} % (fold)
\label{sub:cifar10}

Having verified the $\cplx$ variational sparsification method on MNIST-like datasets
and simple models, we turn to the CIFAR10 dataset comprising $32\times 32$ colour
images of $10$ classes \citep{krizhevsky_learning_2009} and focus on the VGG16 network
\citep{simonyan_very_2015}.
%
% same experimental setup as in MNIST, describe the VGG16 model and its cplx-valued modification
We train the VGG16 network and its $\cplx$ variant, in which we have replaced $\real$-valued
layers with their $\cplx$-valued counterparts. We do not halve or double the features
in any network, since the goal of this experiment is to assess the trade-off for a deep
convolutional network.
%
% differences with mnist
Unlike experiment in sec.~\ref{sub:mnist_like_datasets}, we consider the raw features
only, use full training split, measure accuracy on the usual test split, and allocate
$20$, $40$, and $20$ epochs to each stage.
During training every mini-batch of $128$ samples is augmented by random horizontal flipping
and random cropping, which is done by zero-padding the image with four pixels and extracting
a $32\times 32$ patch from the $40\times 40$ intermediate image.

% compare compressed cplx-vgg against real-vgg
\begin{figure}[!t]
  \centering
  \begin{subfigure}[b]{1.\columnwidth}  % imcl2019-style swears at this
    \includegraphics[width=\columnwidth]{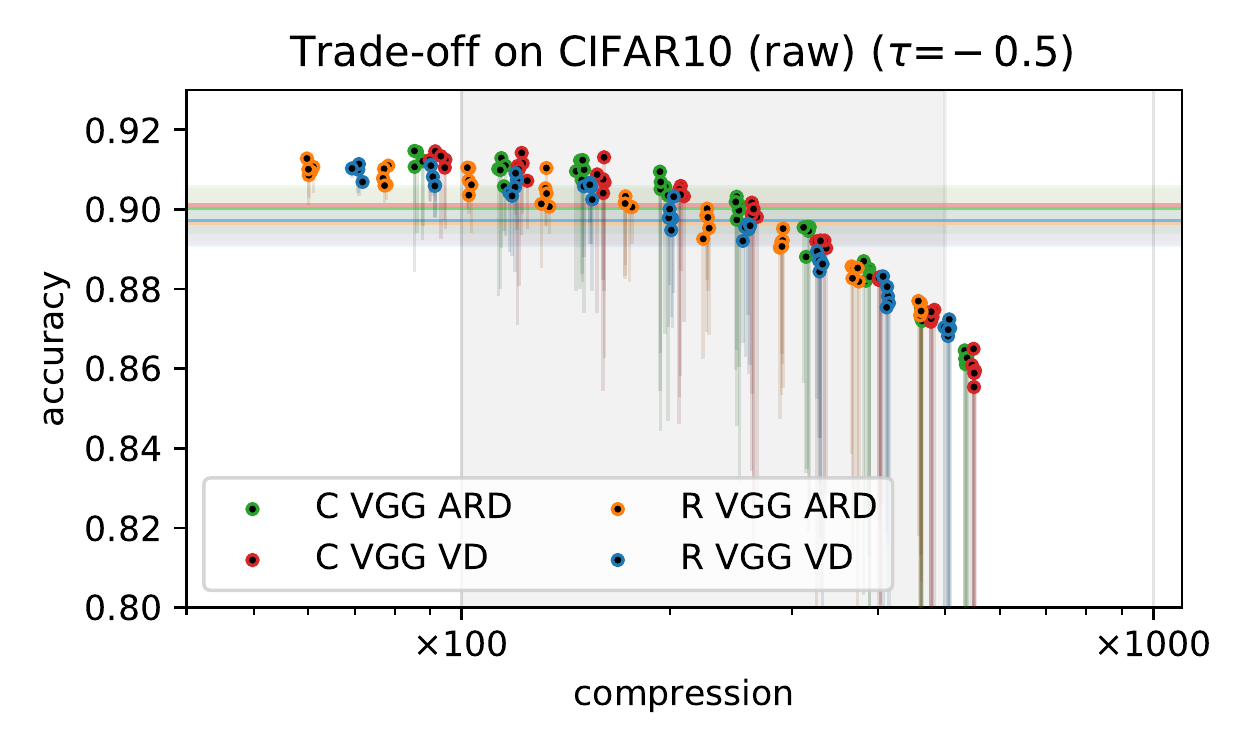}
  \end{subfigure}
  \caption{%
    The compression-accuracy profile for the $\real$ and $\cplx$ VGG16.
  }
  \label{fig:figure__cifar10__trade-off}
\end{figure}

% draw some conclusions
The compression-accuracy curve in figure~\ref{fig:figure__cifar10__trade-off}, constructed
for $C = \tfrac32 2^{-\tfrac{k}2}$ with $k=7, \cdots, 15$, shows that
it is possible to confidently achieve around $\times100$ compression of a deep $\cplx$VNN
without losing accuracy, provided the network is fine-tuned after undergoing Variational
sparsification. Regarding methods themselves, $\cplx$-VD and $\cplx$-ARD follow the same
declining compression-accuracy pattern, but for the same setting $C$ the latter provides
slightly less compression with marginally better accuracy.

% subsection cifar10_100 (end)

\subsection{MusicNet} % (fold)
\label{sub:musicnet}

% describe musicnet as in Trabelsi et al. (2017) also get some hints from thickstun_learning_2017

% The experimental results of the experiments in previous sections indicate that the proposed
% $\cplx$-valued Sparse Variational Dropout achieves the goal of network compression.
% Inspired by this we set out to investigate its effects on in the domain where it has
% been shown that CVNN outperform rvnn.

% Prior experiments demonstrated that the $\cplx$-valued variational sparsification
% what can be expected from prior research.
MusicNet is a corpus of $330$ annotated classical music recordings used for learning
feature representations for music transcription tasks \citep{thickstun_learning_2017}.
\citet{trabelsi_deep_2018} have proposed a $1d$ VGG-like $\cplx$-valued network that
surpassed a similar $\real$-valued network and achieved $72.9\%$ pooled \emph{Average
Precision} on this dataset. Recently \citet{yang_complex_2020} have reported $74.2\%$
AP with a $\cplx$-valued transformer, \citet{thickstun_invariances_2018} have achieved
$77.3\%$ with a four-layer $\real$-valued network on $\log$-spaced spectrogram, and
\citet{draguns_residual_2020} report $78.0\%$ with a residual-shuffle-exchange network.

In this experiment we seek to compress of the $\cplx$VNN proposed by \citet{trabelsi_deep_2018}.
The dataset is split into the same validation and test samples and handled identically
to their study. The input features are $\cplx$-valued Fourier transforms of $4096$-sample
windows from each waveform, and the label vectors are taken from annotations at the middle
of the window. Each epoch lasts for $1000$ random mini-batches of the musical pieces.
% downsample the audio from 44.1kHz to 11kHz, retain only $84$ out of $128$ labels,
% and hold-out the same validation and test compositions
% 
However, we deviate from the set-up used by \citet{trabelsi_deep_2018} by clipping $\ell_2$
norm of the gradients to $0.05$ and shifting the low frequencies of the input to the centre
to maintain spatial locality for convolutions.

Experiments with the uncompressed model aimed at replicating the original result have
shown that early stopping almost always terminates within the first $10-20$ epochs of
the $200$ epochs used in their study, due to the validation performance peaking at
$10-15$ epochs and steadily declining afterwards.
Thus we opt to use shorter stages: $12$, $32$ and $50$ epochs (sec.~\ref{sub:staging}),
with early stopping activated only during the ``fine-tune'' stage. To keep the learning
rate schedule consistent, we scale the learning rate of $10^{-3}$ after $5$, $10$, and
$20$-th epoch by $\tfrac1{10}$, $\tfrac1{20}$ and $\tfrac1{100}$, respectively.

We explore the $\cplx$-VD and $\cplx$-ARD methods by varying $C$ over the grid $
  \{\tfrac14, \tfrac12, \tfrac34, 1\} \cdot 10^{-k}
$ with $k=1, 2, 3$, while keeping $\tau$ at $-\tfrac12$. The performance is measured after
``pre-train'' stage, just before and upon termination of fine-tuning.
Additionally, we test the model of \citet{trabelsi_deep_2018}, in which we purposefully
halve the receptive field of the first convolution from $6$ to $3$ (denoted by suffix $k_3$).
The motivation is to test if the handicap introduced by the forced compression of the most
upstream layer can be alleviated by non-uniform compression, induced by Variational Dropout.
We test only $\cplx$-VD in this sub-experiment, since prior results have not demonstrated
significant superiority of one method over another.  % it should've been ARD

\begin{figure}[!t]
  \centering
  \begin{subfigure}[b]{1.\columnwidth}  % imcl2019-style swears at this
    \centering
    \includegraphics[width=\columnwidth]{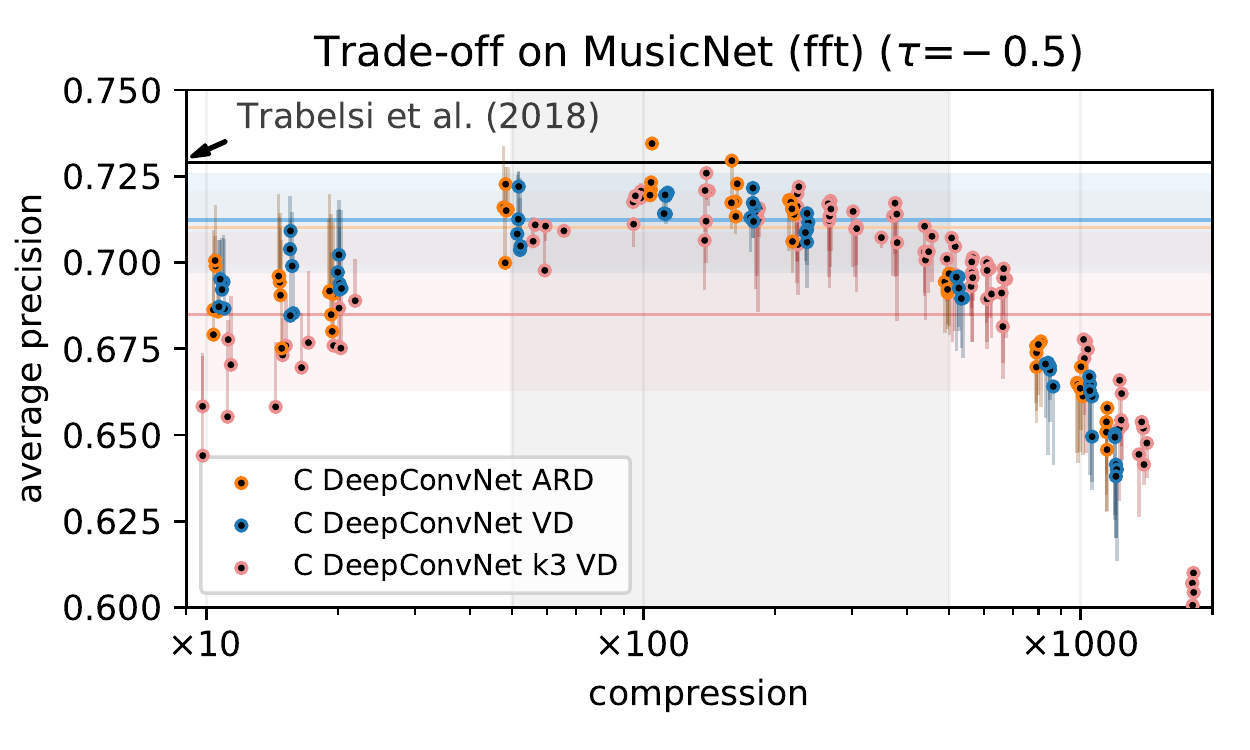}
  \end{subfigure}
  \caption{%
    Performance-compression curve for \textbf{\color{blue} VD}, \textbf{\color{orange} ARD},
    and the ${\color{gray} k_3}$ version compressed with VD.
  }
  \label{fig:musicnet__trade-off}
\end{figure}

The performance-compression frontier in figure~\ref{fig:musicnet__trade-off} shows that
VD and ARD deliver similar compression rates, but ARD slightly outperforms in terms of
the average precision at the cost of marginally lower compression.
At $\times100$ compression level the $k_3$ model outperforms its
uncompressed baseline, but yields lower AP score than the full model. In conjunction with
post-pruning fine-tuning, both $\cplx$-valued variational sparsification methods achieve
average precision level comparable to the result of \citet{trabelsi_deep_2018} with a
network having $50$-$200$ times less parameters.
% any conclusions related to k_3???

% testing the fx of the threshold
\begin{figure}[!t]
  \centering
  \includegraphics[width=1\columnwidth]{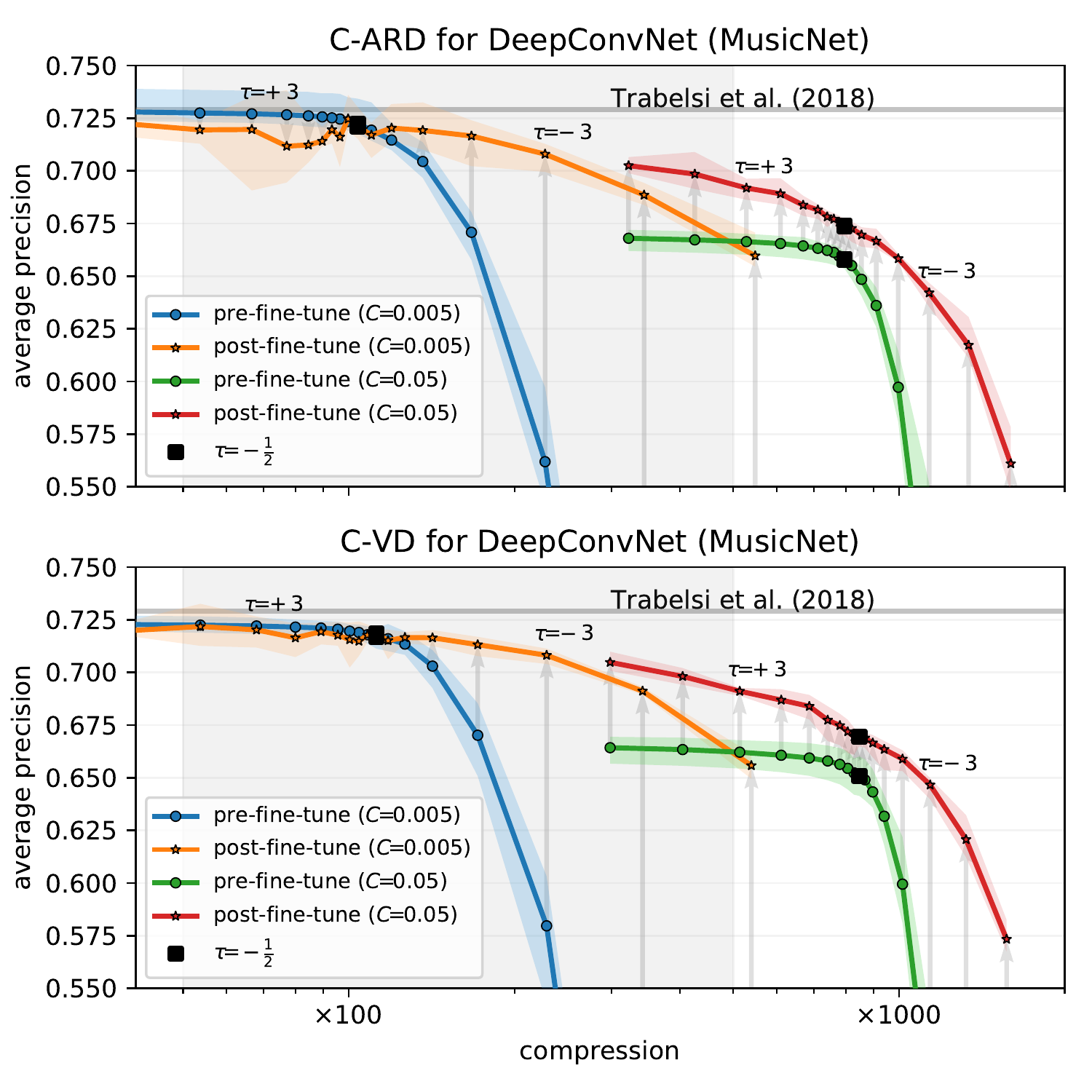}
  \caption{%
    The effect of fine-tuning on performance-compression curves for $
      C\in \{\tfrac1{20}, \frac1{200}\}
    $ in \eqref{eq:elbo_with_coef}.
  }
  \label{fig:hist__and__threshold__tradeoff}
\end{figure}

% effects of extra training epochs and the impact of tau
We take the full models compressed with $C \in \{\tfrac1{20}, \frac1{200}\}$ and re-run
only the fine-tuning stage for various pruning thresholds $
  \tau \in \{\tfrac{k}2\colon k=-8\cdots+8\}
$. The performance-compression curves depicted in figure~\ref{fig:hist__and__threshold__tradeoff}
are parameterized by decreasing $\tau$ from left to right, since models are not re-compressed
which makes $\tau$ monotonically affect the compression rate. From \eqref{eq:elbo_with_coef}
and the relative positions of the curves it can be concluded that $C$ has a much more
substantial impact on the compression profile of each method, than the choice of the
pruning threshold.
% coarser grid than in top plot in order not to wait for 20 compute days, but 9.5 on
% 4 x Geforce 1080Ti, 256 Gb, core-i7 machine, running 4*2 experiments in parallel

% takeaway: fine tune seems to improve performance for highly compressed models, but
% within a limit. Worth a try for moderately compressed models, but willl must likely
% overfit for undercompressed models. Varying $C$ in \eqref{eq:elbo_with_coef} affects
% sparsity more strongly, than varying $\tau$. But this is only observational.

\begin{figure}[!t]
  \centering
  \begin{subfigure}[b]{1.\columnwidth}  % imcl2019-style swears at this
    \centering
    \includegraphics[width=\columnwidth]{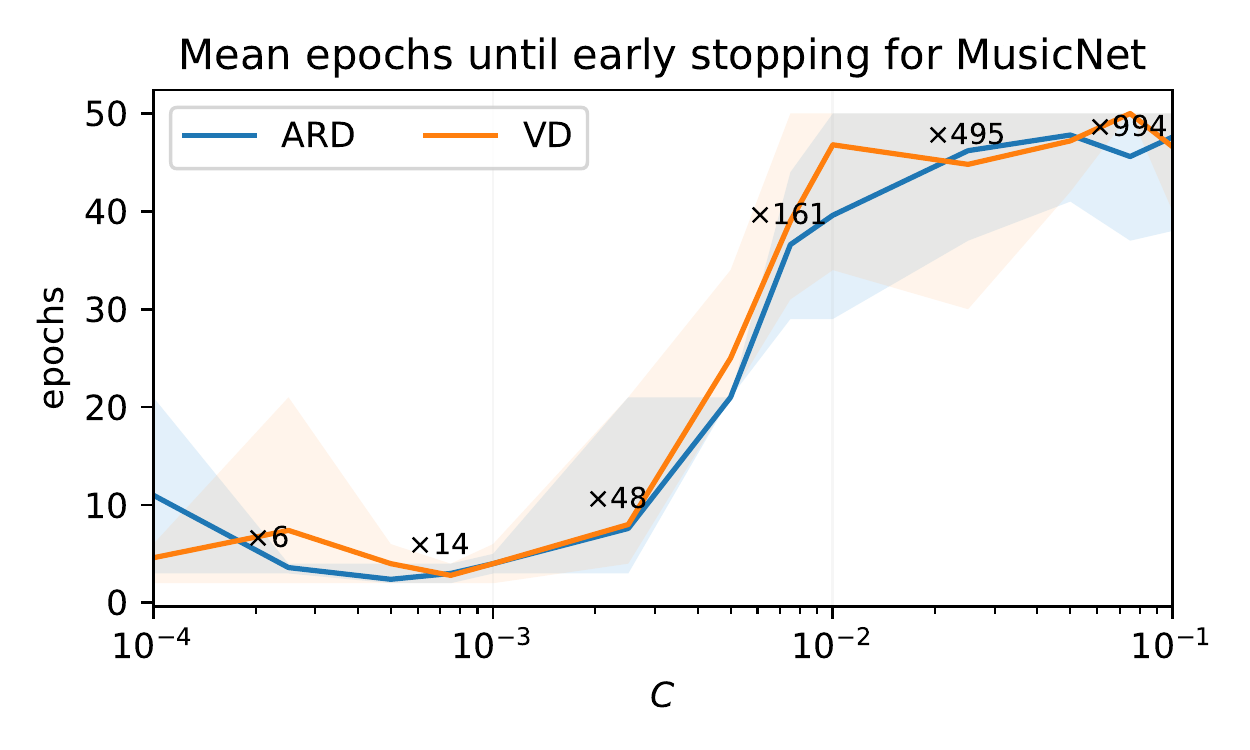}
  \end{subfigure}
  \caption{%
    Early stopping epoch at fine-tuning stage.
  }
  \label{fig:musicnet__early_stopping}
\end{figure}

% why is there such a dramamtic difference in impact of ``fine-tune''?
We provide the following interpretation of the apparent contrast in performance impact
borne by fine-tuning between less than $\times50$ and higher than $\times100$ compression
regimes in figure~\ref{fig:musicnet__trade-off}, also observed in sec.~\ref{sub:mnist_like_datasets}.
%
% * early stopping kick in at around 10-20 epoch
% * performance peaks at 10-15 epoch and then gradually declines
The value of $C$ in \eqref{eq:elbo_with_coef} is a good proxy for the ranking of the
final compression rate since it directly affects the feedback from sparsifying prior.
So, during the $50$ epoch allotted for ``sparsify'' stage, low $C$ prevents the sparsity
inducing prior from pulling the posterior sufficiently away from the likelihood-maximizing
parameters inherited from the ``pre-train'' stage. It is reasonable, therefore, to expect
that for undercompressed models the fine-tuning stage acts essentially as a continuation
of pre-training. And, since we have observed that longer training invariably deteriorates
the validation performance, the ``fine-tune'' stage should lead to overfitting for small
$C$.
Figure~\ref{fig:musicnet__early_stopping} shows that the models, which have been sparsified
with $C$ less than $\tfrac1{400}$, have less than $50\times$ compression and need considerably
less training epochs before early stopping terminates the process.

% subsection musicnet (end)

% section experiments (end)

\section{Conclusion} % (fold)
\label{sec:conclusion}

In this study we have presented $\cplx$-valued variational sparsification methods to
the ever growing set of tools for learning deep $\cplx$-valued neural networks. To
validate these methods we have carried out a large numerical study of $\cplx$VNN with
simple architectures to assess the feasible performance-compression trade-off, and
studied compression of two deep convolutional $\cplx$VNN. At the cost of marginally
lower performance, we have achieved $\times50$-$\times100$ compression of the deep
$\cplx$VNN of \citet{trabelsi_deep_2018} on the MusicNet.
% CVNN capacity grows by \sqrt{2}, compute grows by 2 % RVNN caption by 2 compute by 2. 

Experimental results show that $\cplx$-VD (sec.~\ref{ssub:vd_prior}) and $\cplx$-ARD
(sec.~\ref{ssub:ard_prior}) exhibit trade-off profiles matching their $\real$-valued
counterparts. This makes us confident that the overall conclusion of \citet{gale_state_2019}
is applicable to $\cplx$VNN and the proposed $\cplx$-valued variational sparsification
methods.
% L0 and dropout work in the low-to-mid sparsity range, variational dropout consistently
% ranked behind L0 on Transformer, and was bested by magnitude pruning for sparsity
% levels of 80\% and up. Variational dropout consistently produces models on-par or better
% than magnitude pruning on ResNet-50, and l0 fails to produce sparse models at all.
Furthermore our findings indicate that between each other under similar circumstances
the methods yield comparable compression and performance results, which echoes earlier
results by \citet{kharitonov_variational_2018}.

This study has direct implications for embedded deep learning applications
both in terms of lower storage requirements and higher throughput stemming from fewer
floating point multiplications due to sparsity, despite somewhat higher arithmetic
complexity of $\cplx$-valued networks.

% section conclusion (end)

\section*{Software and Data} % (fold)
\label{sec:software_and_data}

% we strongly encourage the publication of software and data with the camera-ready version
% of the paper whenever appropriate. This can be done by including a URL in the copy.
% do not include URLs that reveal your institution or identity in your submission for review
% * provide an anonymous URL or upload the material as ``Supplementary Material'' into the CMT

The source code for a package based on PyTorch \citep{paszke_pytorch_2019}, which
implements $\cplx$-valued Sparse Variational Dropout and ARD layers and provides other
basic layers for $\cplx$VNN is available at
\url{https://github.com/ivannz/cplxmodule}.
The source code for the experiments and the figures in this study is available at
\url{https://github.com/ivannz/complex_paper/tree/v2020.6}.

% section* software_and_data (end)

\section*{Acknowledgements} % (fold)
\label{sec:acknowledgements}

% Do not include acknowledgements in the initial version of the paper submitted for blind review.

% the final camera-ready version can (and probably should) include acknowledgements
% Typically, this will include thanks to reviewers who gave useful comments, to colleagues
% who contributed to the ideas, and to funding agencies and corporate sponsors that provided
% financial support.
We would like to thank the anonymous reviewers, Evgenii Egorov, Ruslan Kostoev (ADASE)
and Danila Doroshin (Huawei) for their useful comments.
% We thank Skoltech CDISE HPC Zhores cluster staff for providing computational infrastructure.
The authors acknowledge the use of the Skoltech CDISE HPC cluster ``Zhores'' for obtaining
the results presented in this paper.

% section* acknowledgements (end)

% \clearpage

\bibliographystyle{abbrvnat}
\bibliography{references}
% \nocite{*}

% \clearpage

\appendix
\onecolumn

\section{MNIST-like experiments} % (fold)
\label{sec:mnist_like_experiments}

The plots presented in this appendix support the conclusions made in the main text and
provide an overview of the experiments conducted on MNIST-like datasets.

\begin{figure}[b]
  \centering
  \begin{subfigure}[b]{0.5\columnwidth}
    \centering
    % used in main text
    \includegraphics[width=\linewidth]{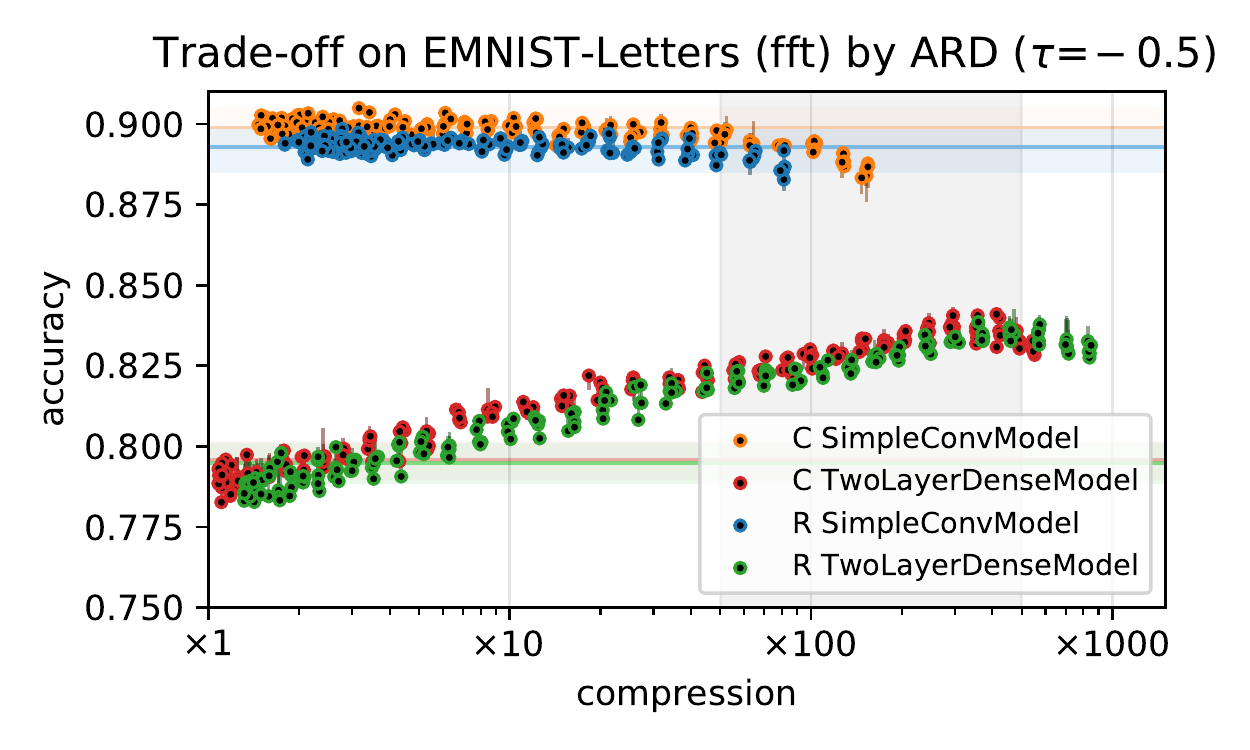}
  \end{subfigure}%
  \begin{subfigure}[b]{0.5\columnwidth}
    \centering
    \includegraphics[width=\linewidth]{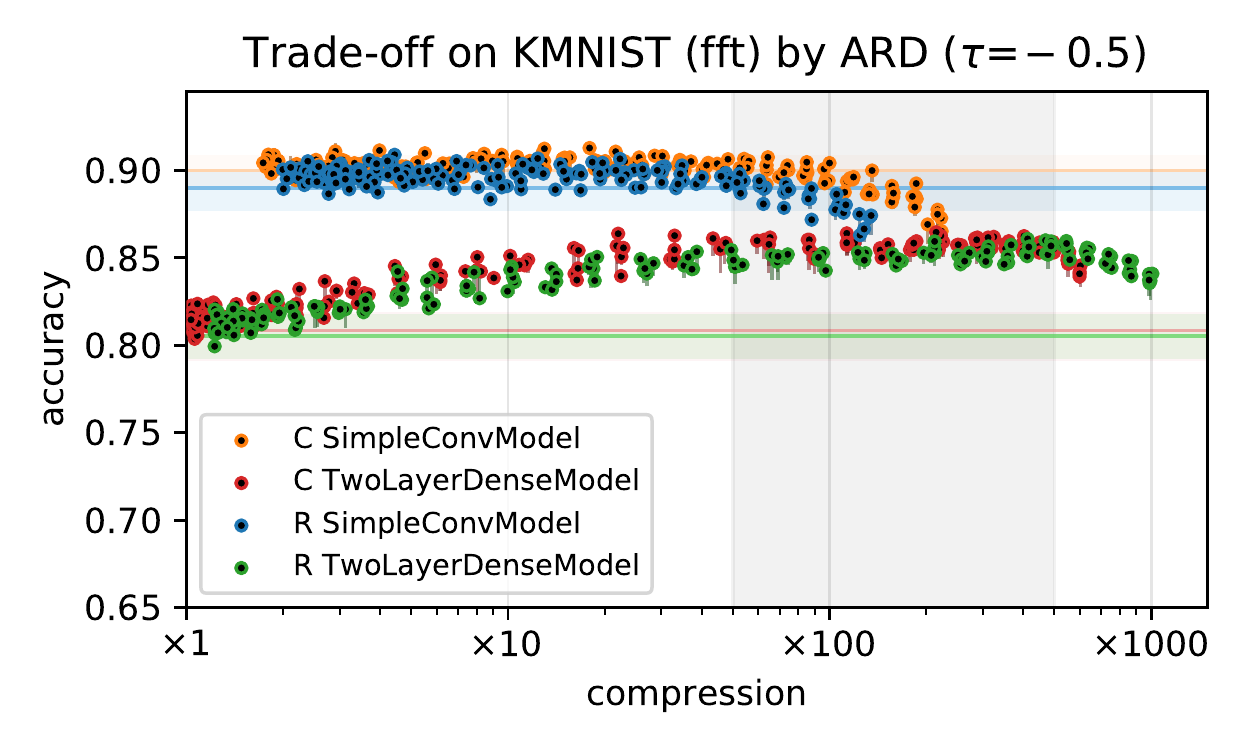}
  \end{subfigure} \\ %
  \begin{subfigure}[b]{0.5\columnwidth}
    \centering
    \includegraphics[width=\linewidth]{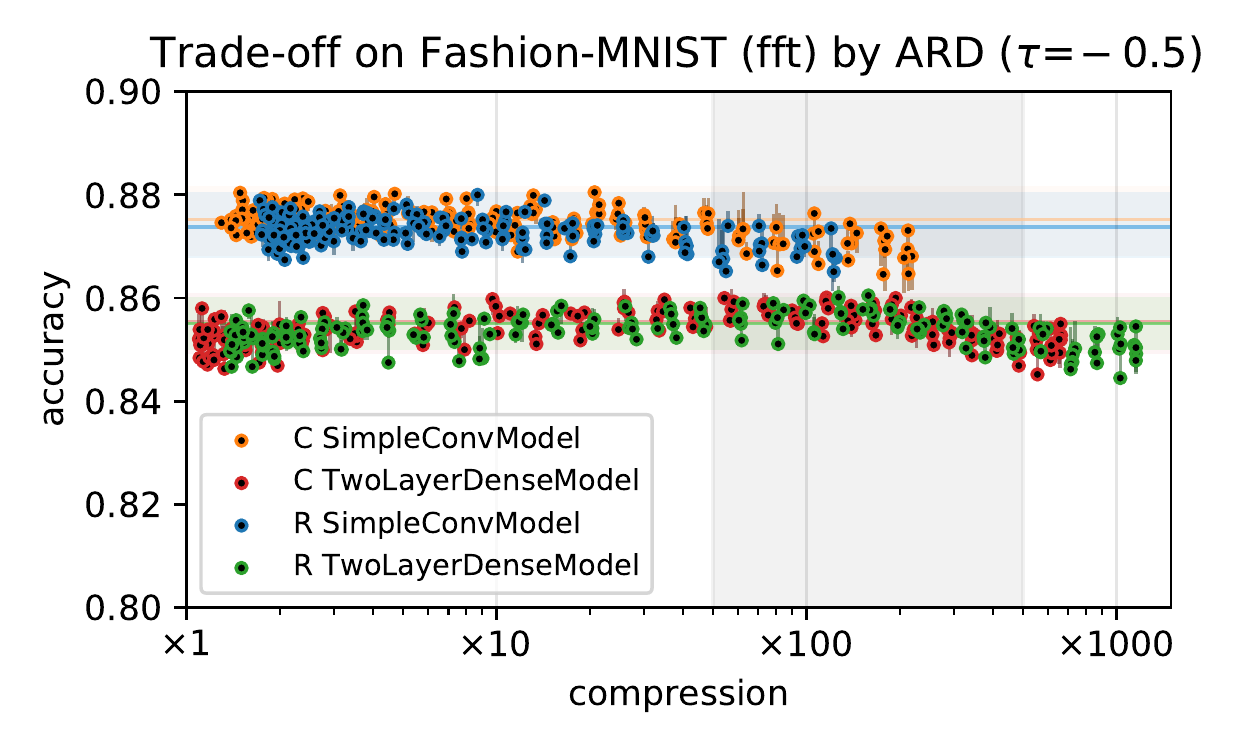}
  \end{subfigure}%
  \begin{subfigure}[b]{0.5\columnwidth}
    \centering
    \includegraphics[width=\linewidth]{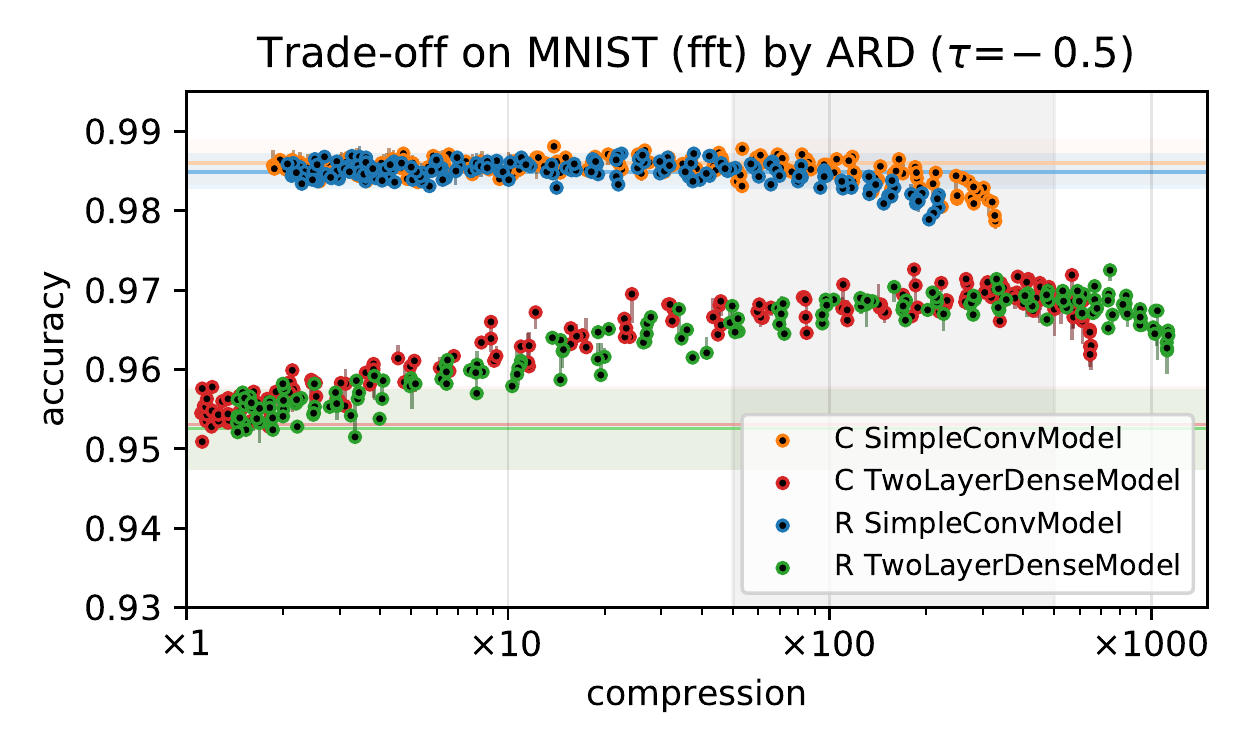}
  \end{subfigure}
  \caption{%
    The trade-off of ARD method for $\real$ and $\cplx$ models using Fourier features.
  }
  \label{fig:appendix__mnist-like__trade-off__ARD__fft}
\end{figure}

\begin{figure}[b]
  \centering
  \begin{subfigure}[b]{0.5\columnwidth}
    \centering
    % used in main text
    \includegraphics[width=\linewidth]{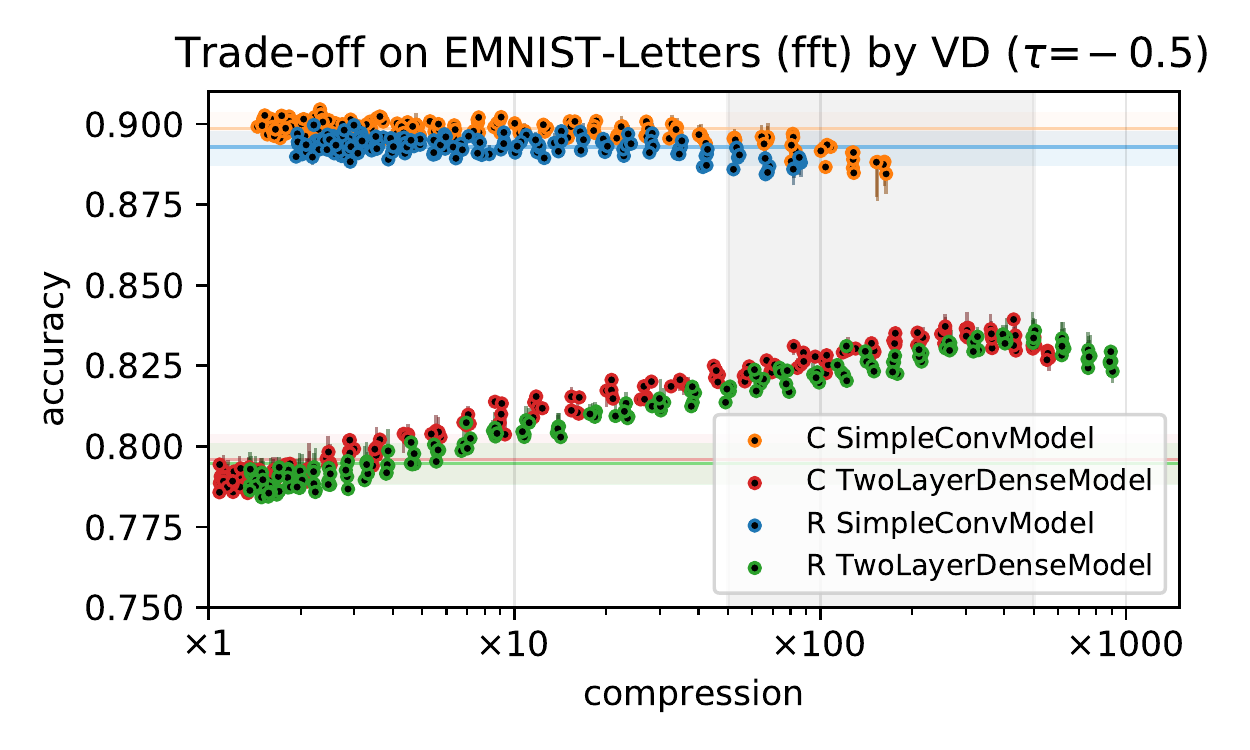}
  \end{subfigure}%
  \begin{subfigure}[b]{0.5\columnwidth}
    \centering
    \includegraphics[width=\linewidth]{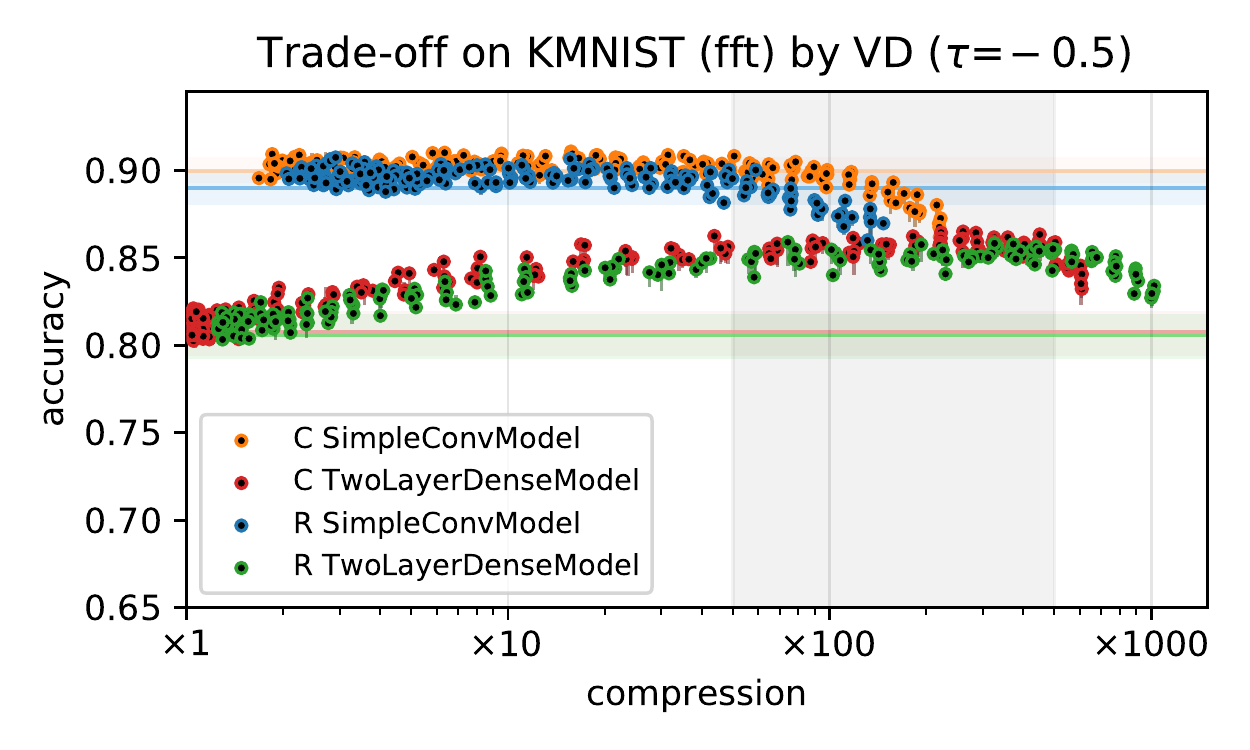}
  \end{subfigure} \\ %
  \begin{subfigure}[b]{0.5\columnwidth}
    \centering
    \includegraphics[width=\linewidth]{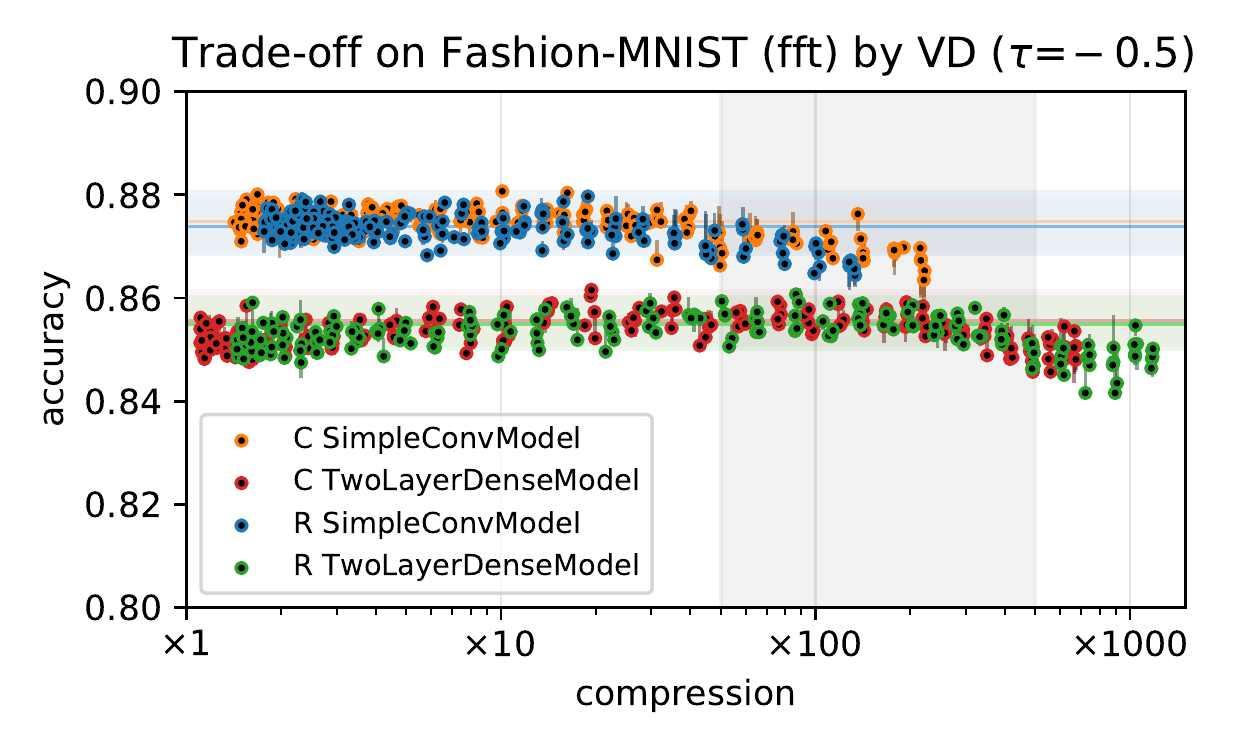}
  \end{subfigure}%
  \begin{subfigure}[b]{0.5\columnwidth}
    \centering
    \includegraphics[width=\linewidth]{figure__mnist-like__trade-off/appendix__VD__mnist__fft__-0.5.pdf}
  \end{subfigure}
  \caption{%
    The trade-off of VD method for $\real$ and $\cplx$ models using Fourier features.
  }
  \label{fig:appendix__mnist-like__trade-off__VD__fft}
\end{figure}

\begin{figure}[b]
  \centering
  \begin{subfigure}[b]{0.5\columnwidth}
    \centering
    % used in main text
    \includegraphics[width=\linewidth]{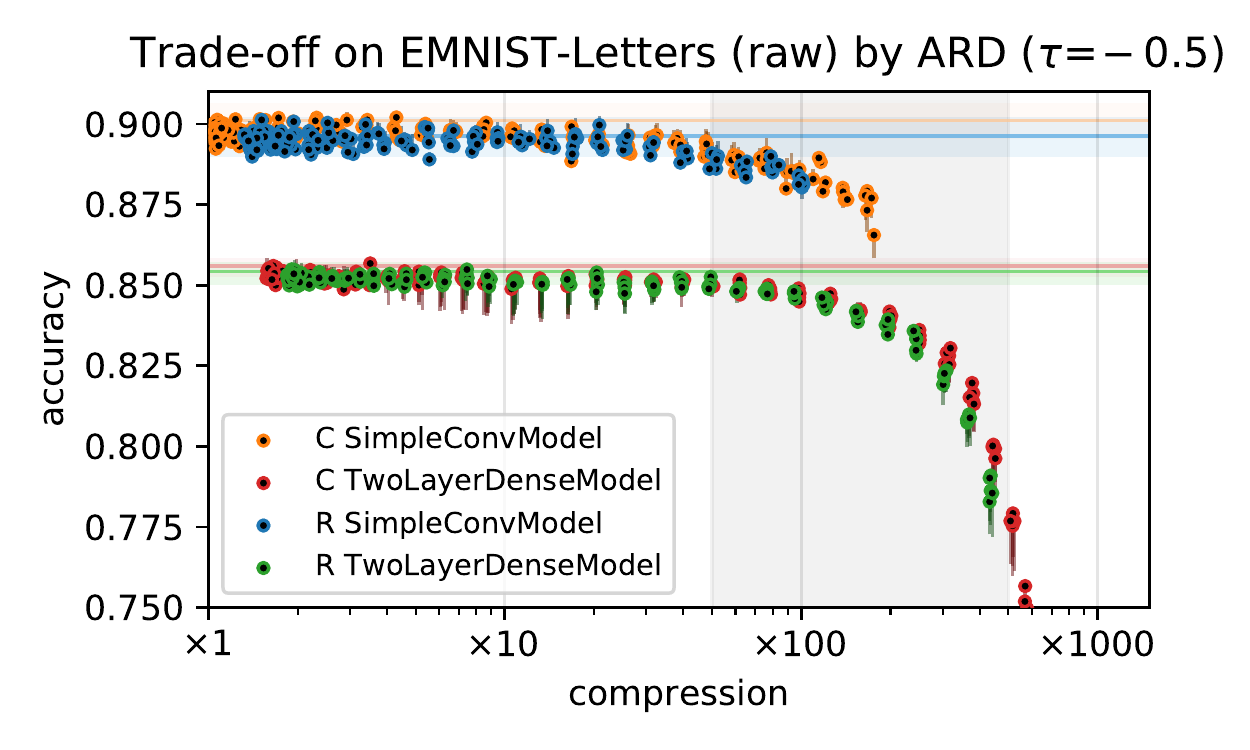}
  \end{subfigure}%
  \begin{subfigure}[b]{0.5\columnwidth}
    \centering
    \includegraphics[width=\linewidth]{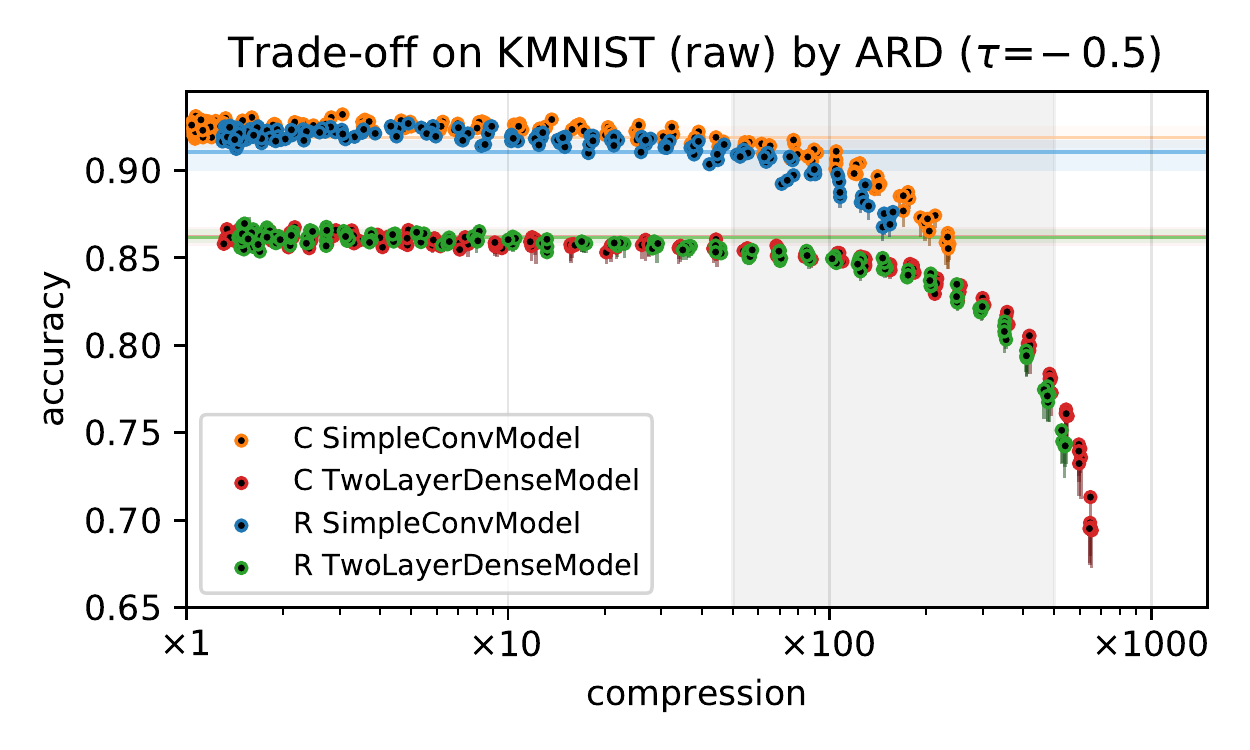}
  \end{subfigure} \\%
  \begin{subfigure}[b]{0.5\columnwidth}
    \centering
    \includegraphics[width=\linewidth]{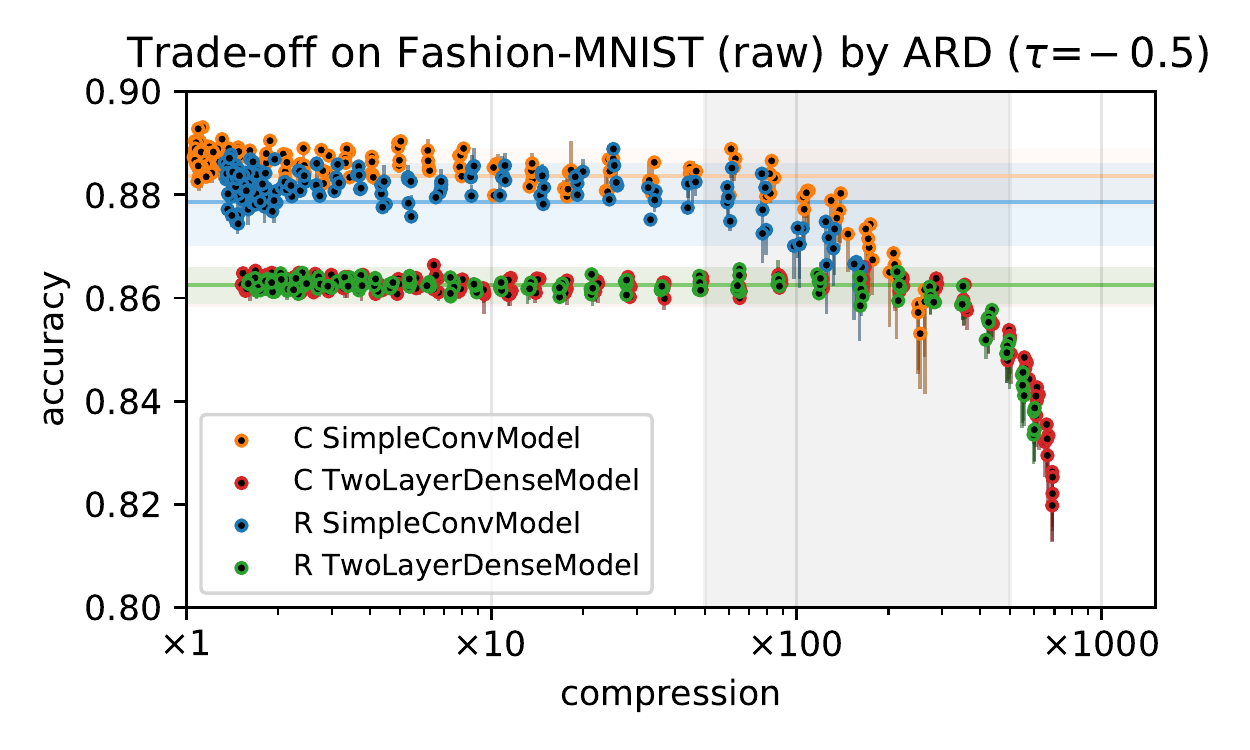}
  \end{subfigure}%
  \begin{subfigure}[b]{0.5\columnwidth}
    \centering
    \includegraphics[width=\linewidth]{figure__mnist-like__trade-off/appendix__ARD__mnist__raw__-0.5.pdf}
  \end{subfigure}
  \caption{%
    The trade-off of ARD method for $\real$ and $\cplx$ models using raw features.
  }
  \label{fig:appendix__mnist-like__trade-off__ARD__raw}
\end{figure}

\begin{figure}[b]
  \centering
  \begin{subfigure}[b]{0.5\columnwidth}
    \centering
    % used in main text
    \includegraphics[width=\linewidth]{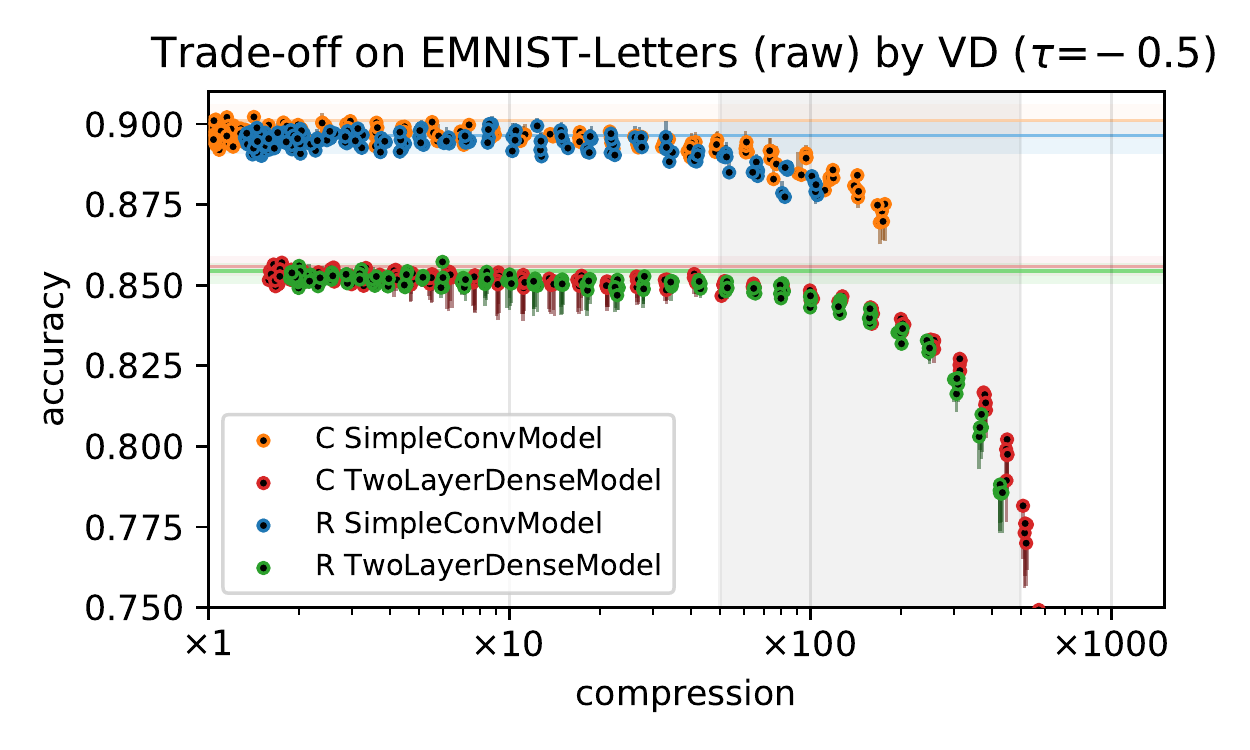}
  \end{subfigure}%
  \begin{subfigure}[b]{0.5\columnwidth}
    \centering
    \includegraphics[width=\linewidth]{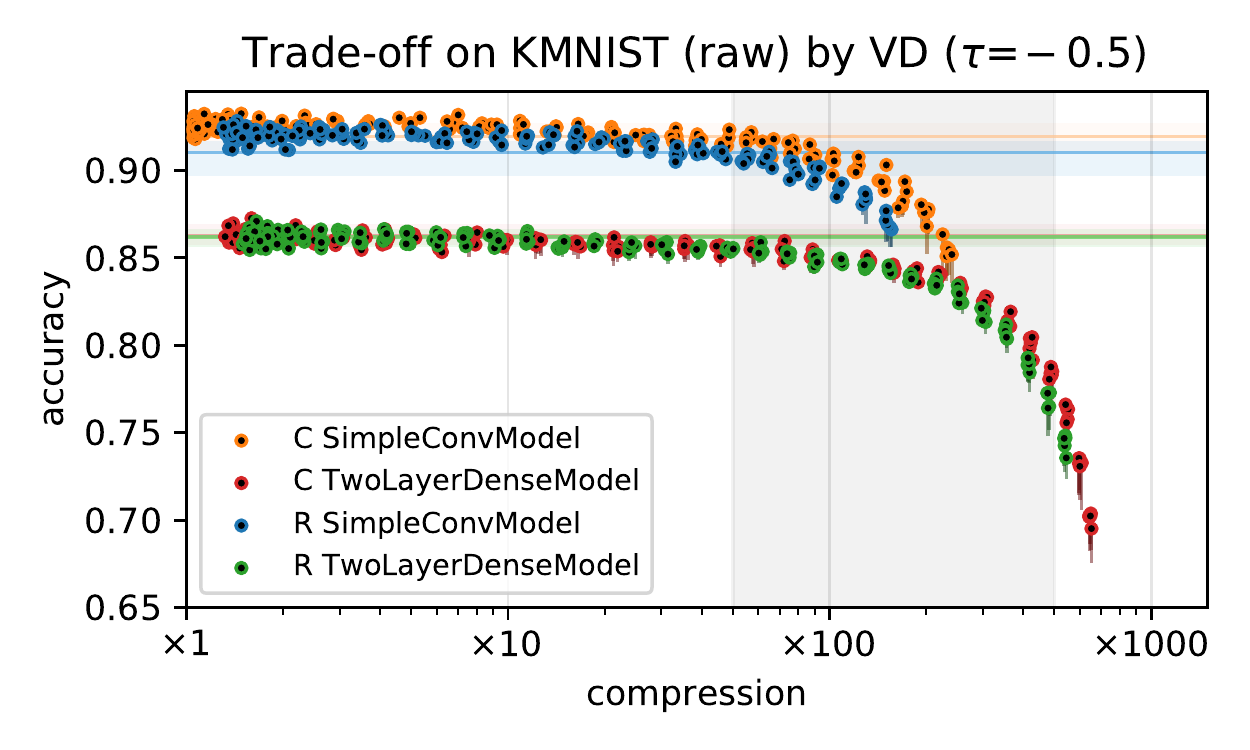}
  \end{subfigure} \\%
  \begin{subfigure}[b]{0.5\columnwidth}
    \centering
    \includegraphics[width=\linewidth]{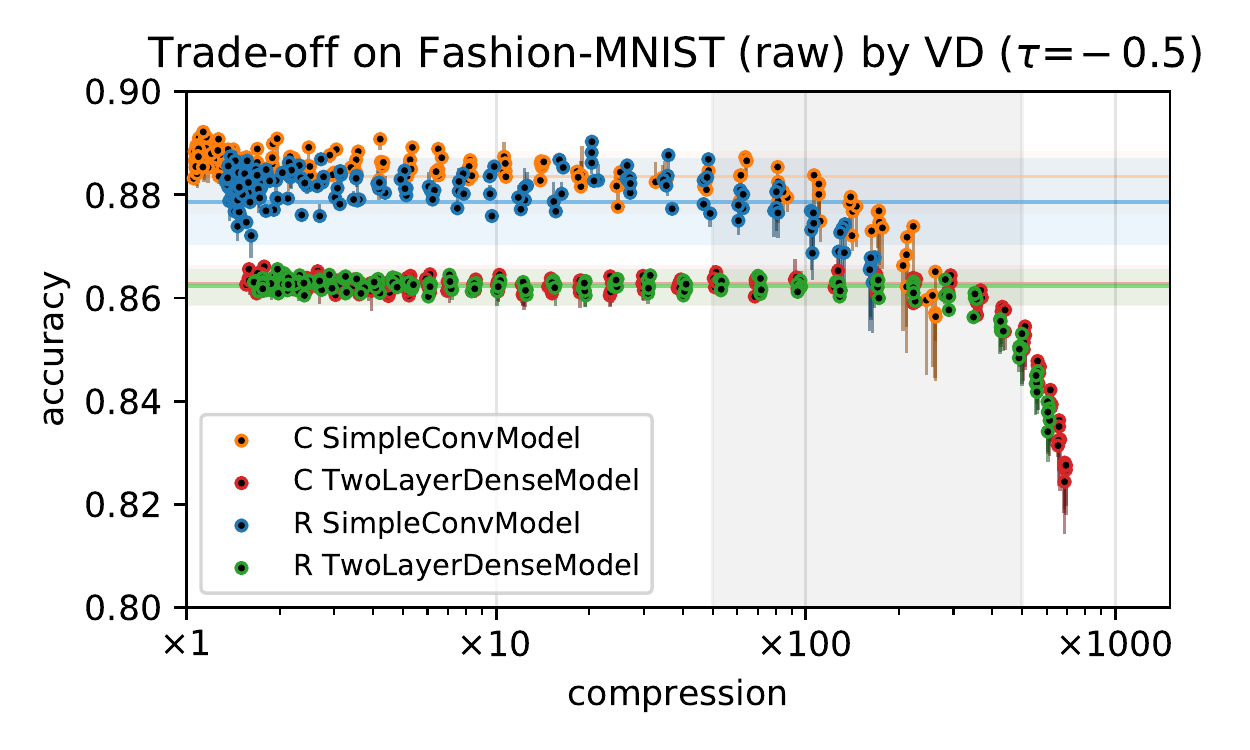}
  \end{subfigure}%
  \begin{subfigure}[b]{0.5\columnwidth}
    \centering
    \includegraphics[width=\linewidth]{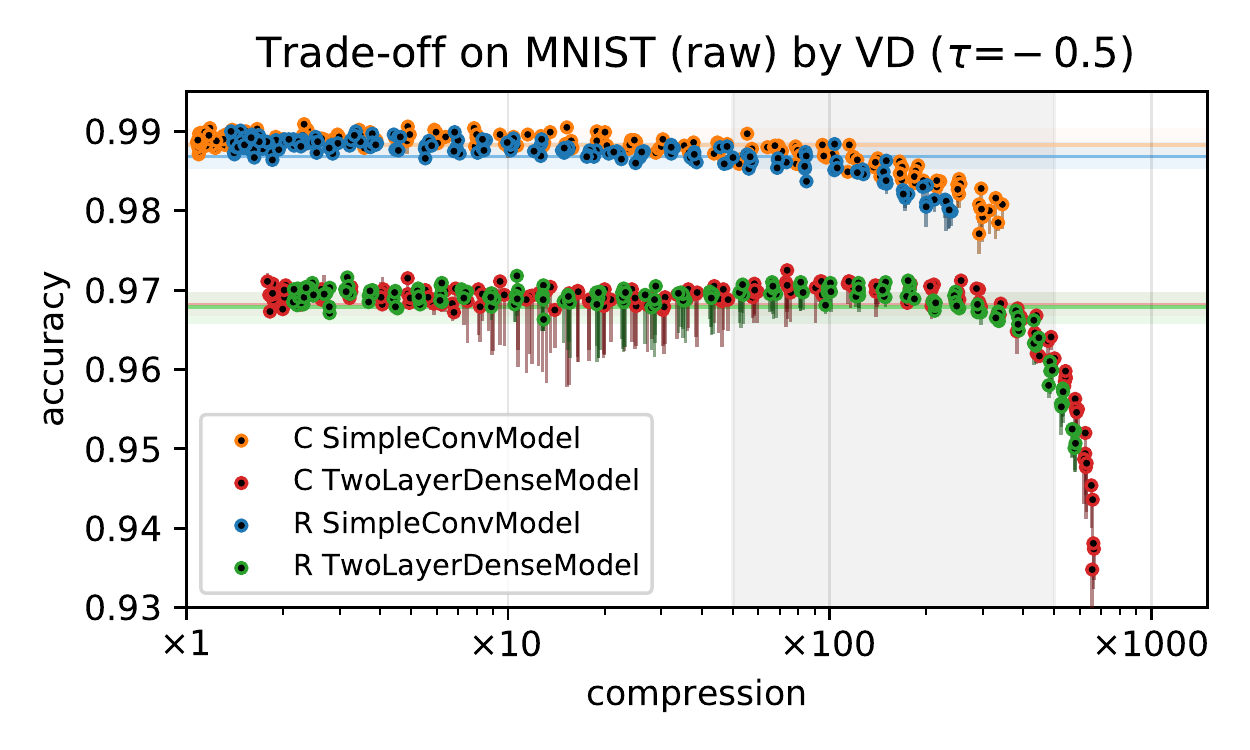}
  \end{subfigure}
  \caption{%
    The trade-off of VD method for $\real$ and $\cplx$ models using raw features.
  }
  \label{fig:appendix__mnist-like__trade-off__VD__raw}
\end{figure}

Each figure shows the compression-accuracy trade-off of a particular method and input
features for \emph{SimpleConvModel} and \emph{TwoLayerDenseModel} models for all four
of the studied datasets (described in the main text): EMNIST-Letters on the \emph{top-left},
KMNIST -- \emph{top-right}, Fashion MNIST -- \emph{bottom-left}, and MNIST on the
\emph{bottom-right}.
Figures \ref{fig:appendix__mnist-like__trade-off__ARD__fft}, \ref{fig:appendix__mnist-like__trade-off__VD__fft},
\ref{fig:appendix__mnist-like__trade-off__ARD__raw}, and \ref{fig:appendix__mnist-like__trade-off__VD__raw}
present $\real$ and $\cplx$ models with \emph{the same intermediate feature sizes}.

\begin{figure}[b]
  \centering
  \begin{subfigure}[b]{0.5\columnwidth}
    \centering
    \includegraphics[width=\linewidth]{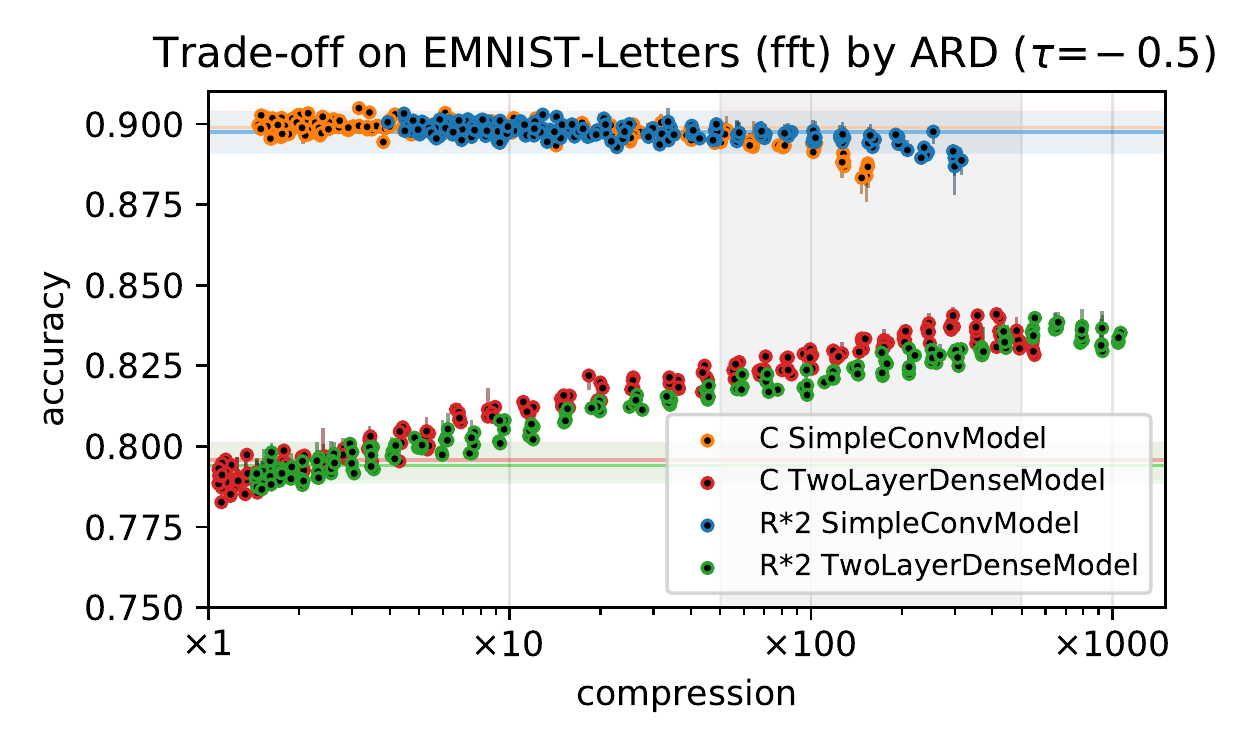}
  \end{subfigure}%
  \begin{subfigure}[b]{0.5\columnwidth}
    \centering
    \includegraphics[width=\linewidth]{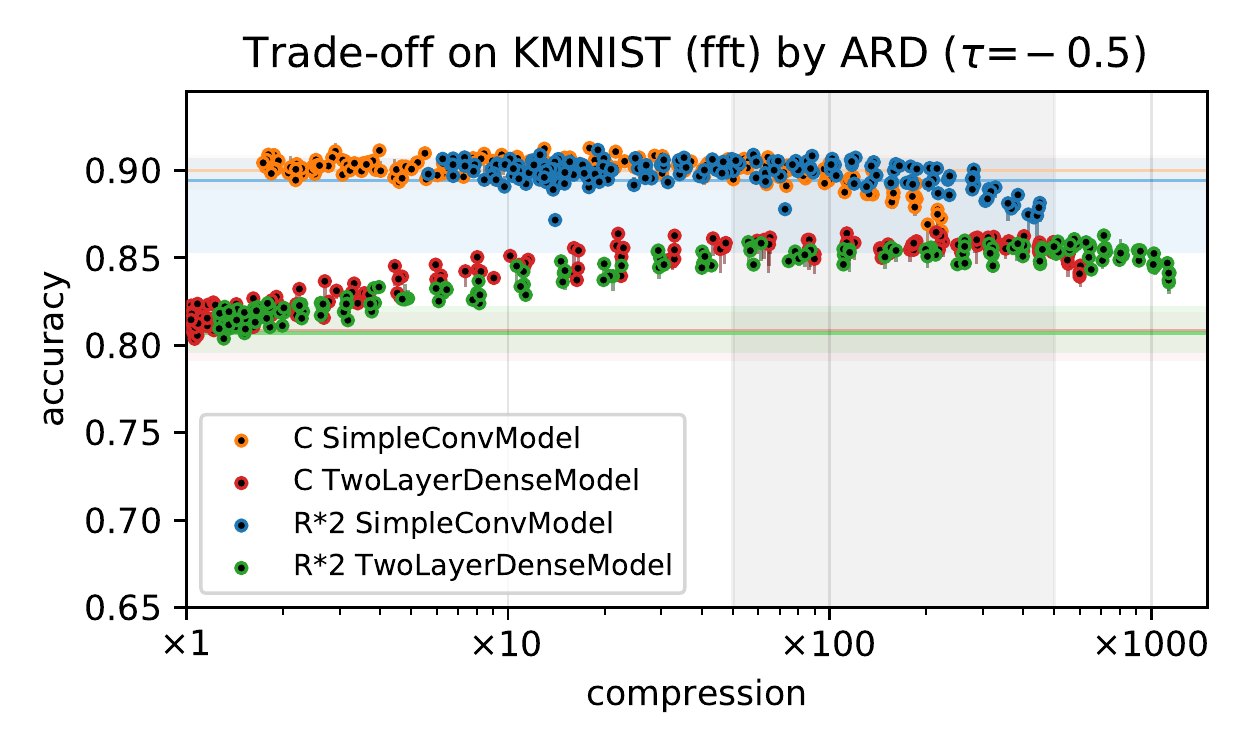}
  \end{subfigure} \\ %
  \begin{subfigure}[b]{0.5\columnwidth}
    \centering
    \includegraphics[width=\linewidth]{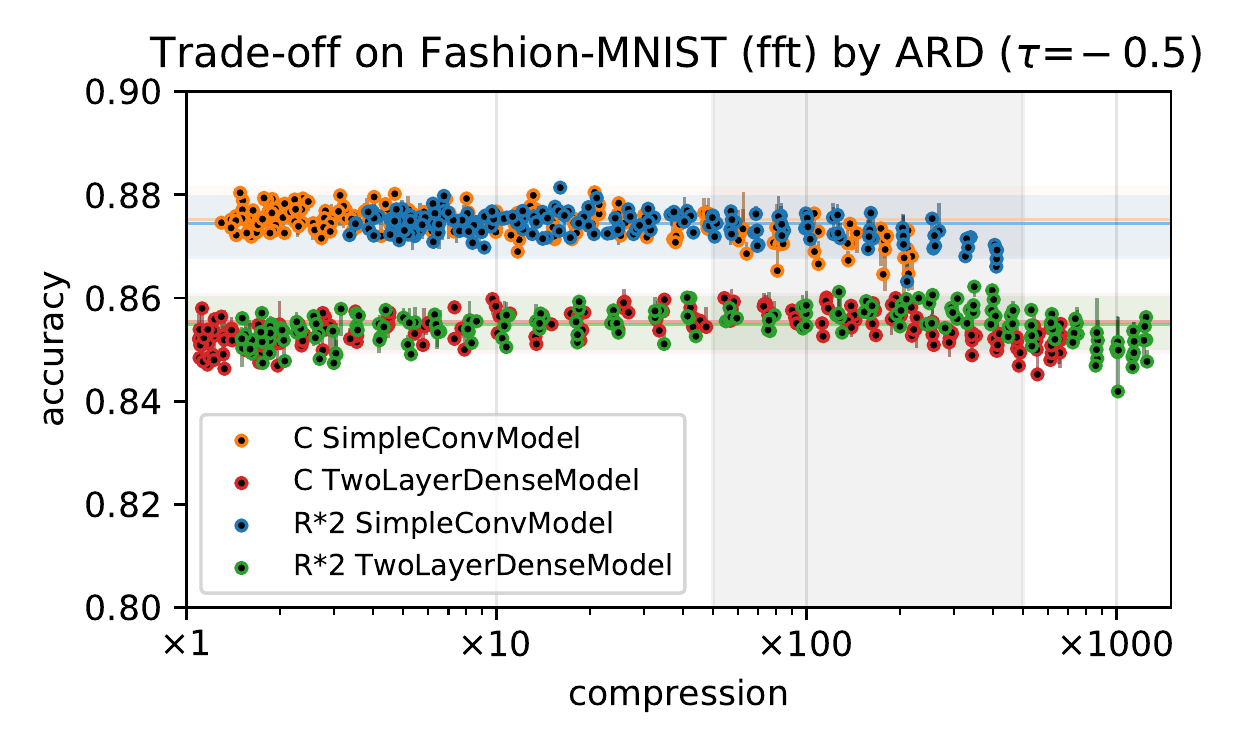}
  \end{subfigure}%
  \begin{subfigure}[b]{0.5\columnwidth}
    \centering
    \includegraphics[width=\linewidth]{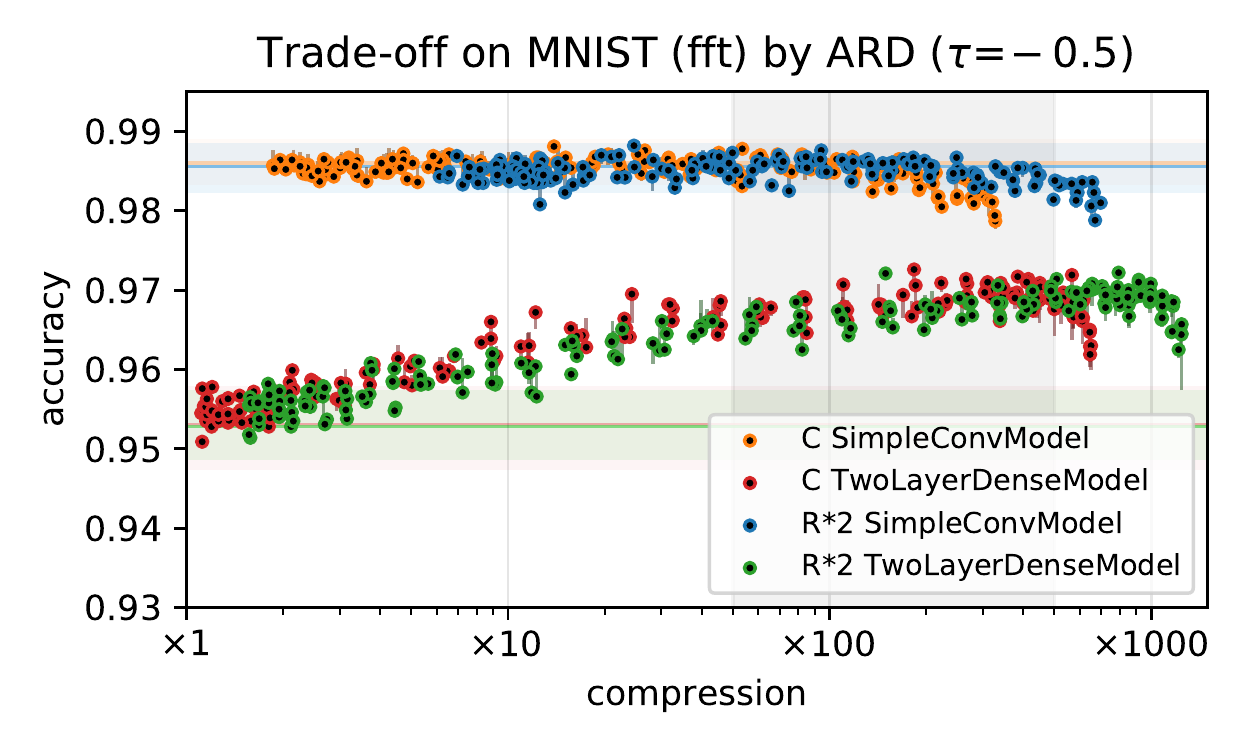}
  \end{subfigure}
  \caption{%
    The trade-off of ARD method for $2\real$ and $\cplx$ models using Fourier features.
  }
  \label{fig:appendix__cmp__mnist-like__trade-off__ARD__fft}
\end{figure}

\begin{figure}[b]
  \centering
  \begin{subfigure}[b]{0.5\columnwidth}
    \centering
    \includegraphics[width=\linewidth]{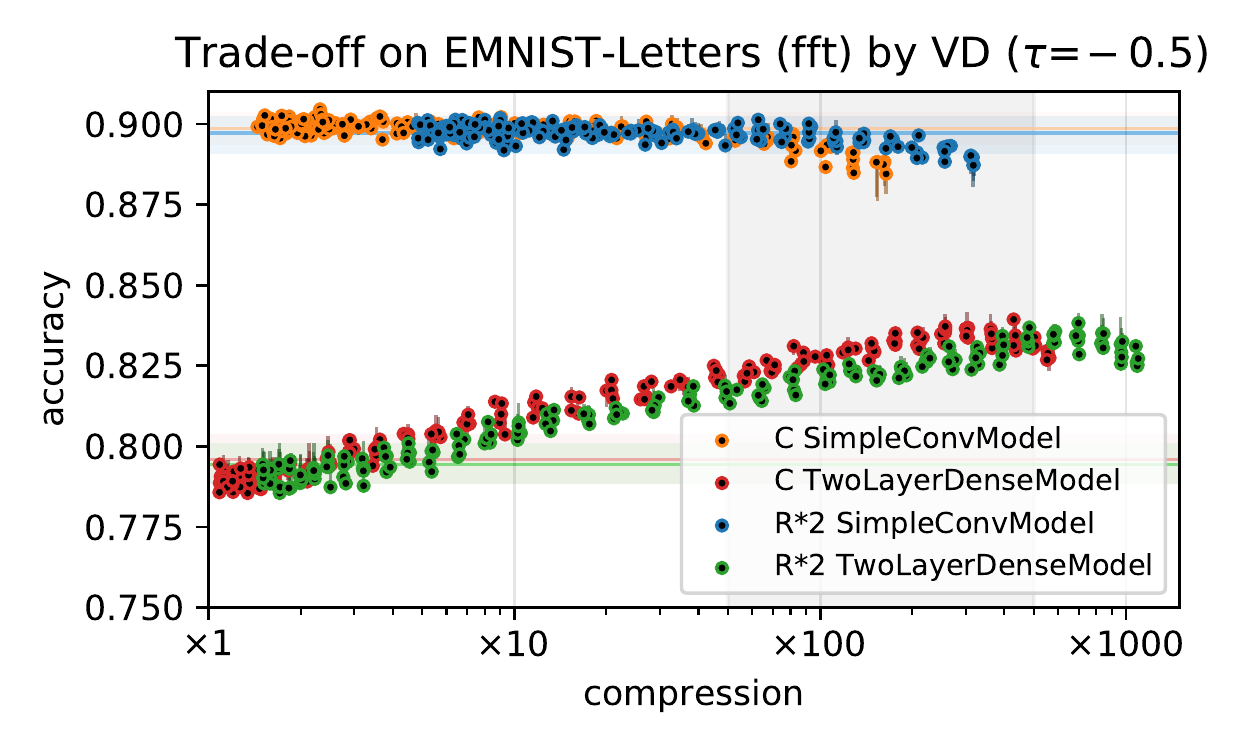}
  \end{subfigure}%
  \begin{subfigure}[b]{0.5\columnwidth}
    \centering
    \includegraphics[width=\linewidth]{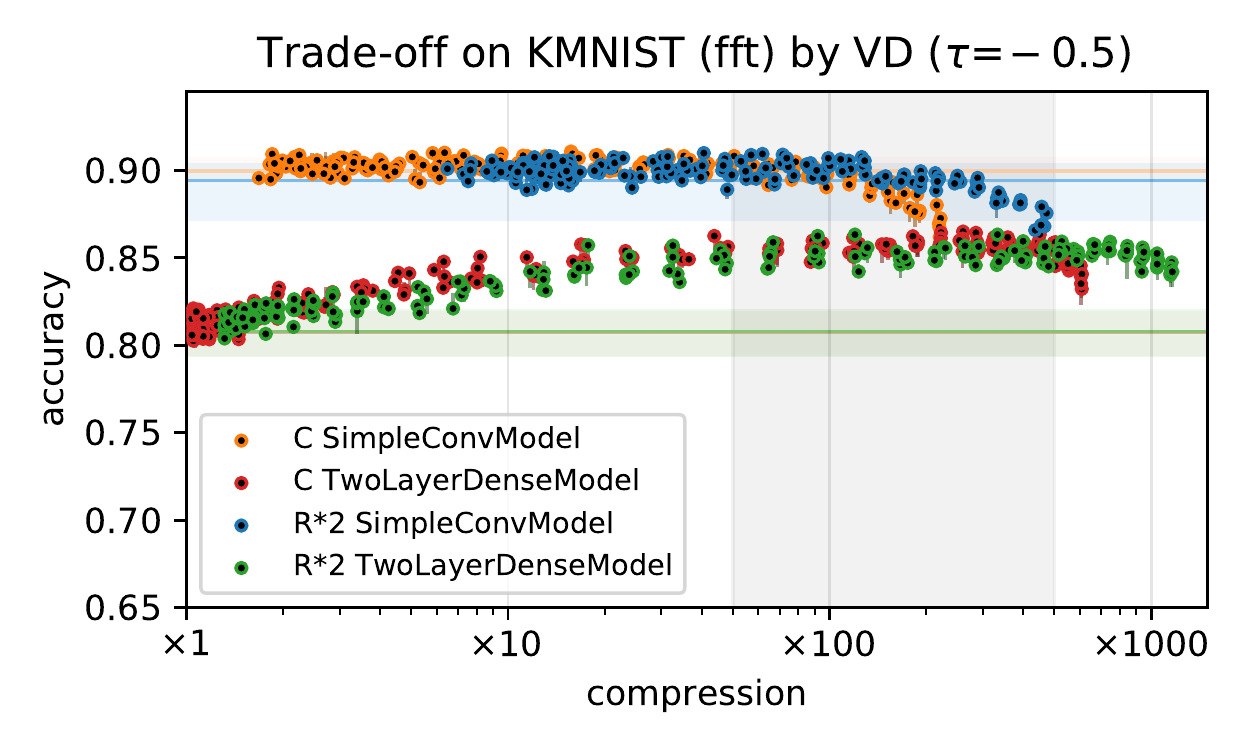}
  \end{subfigure} \\ %
  \begin{subfigure}[b]{0.5\columnwidth}
    \centering
    \includegraphics[width=\linewidth]{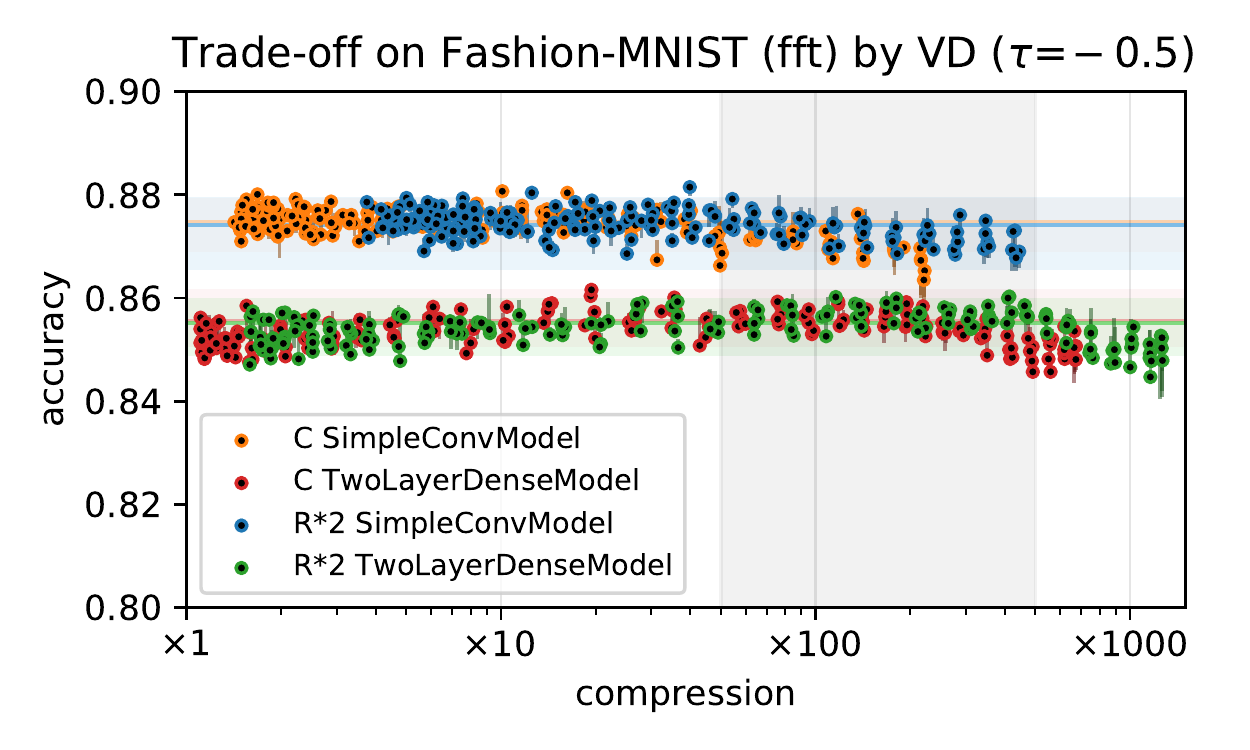}
  \end{subfigure}%
  \begin{subfigure}[b]{0.5\columnwidth}
    \centering
    \includegraphics[width=\linewidth]{figure__mnist-like__trade-off/appendix__cmp__VD__mnist__fft__-0.5.pdf}
  \end{subfigure}
  \caption{%
    The trade-off of VD method for $2\real$ and $\cplx$ models using Fourier features.
  }
  \label{fig:appendix__cmp__mnist-like__trade-off__VD__fft}
\end{figure}

\begin{figure}[b]
  \centering
  \begin{subfigure}[b]{0.5\columnwidth}
    \centering
    \includegraphics[width=\linewidth]{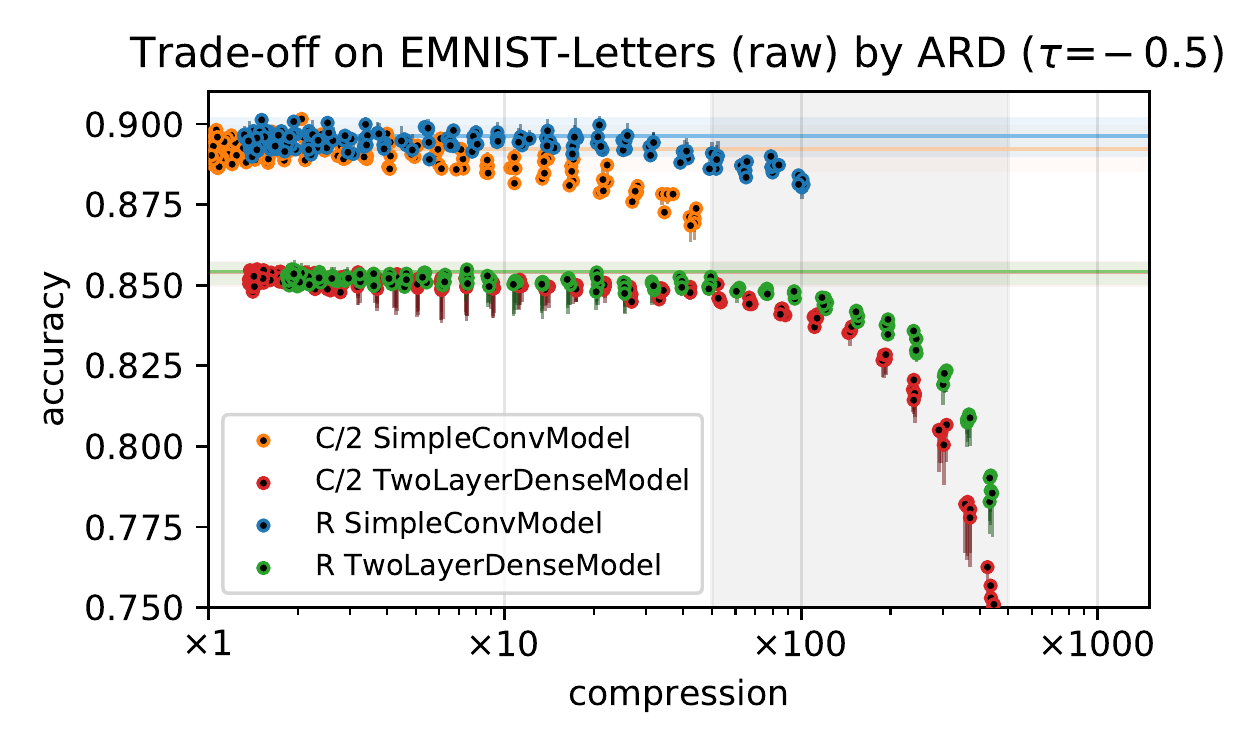}
  \end{subfigure}%
  \begin{subfigure}[b]{0.5\columnwidth}
    \centering
    \includegraphics[width=\linewidth]{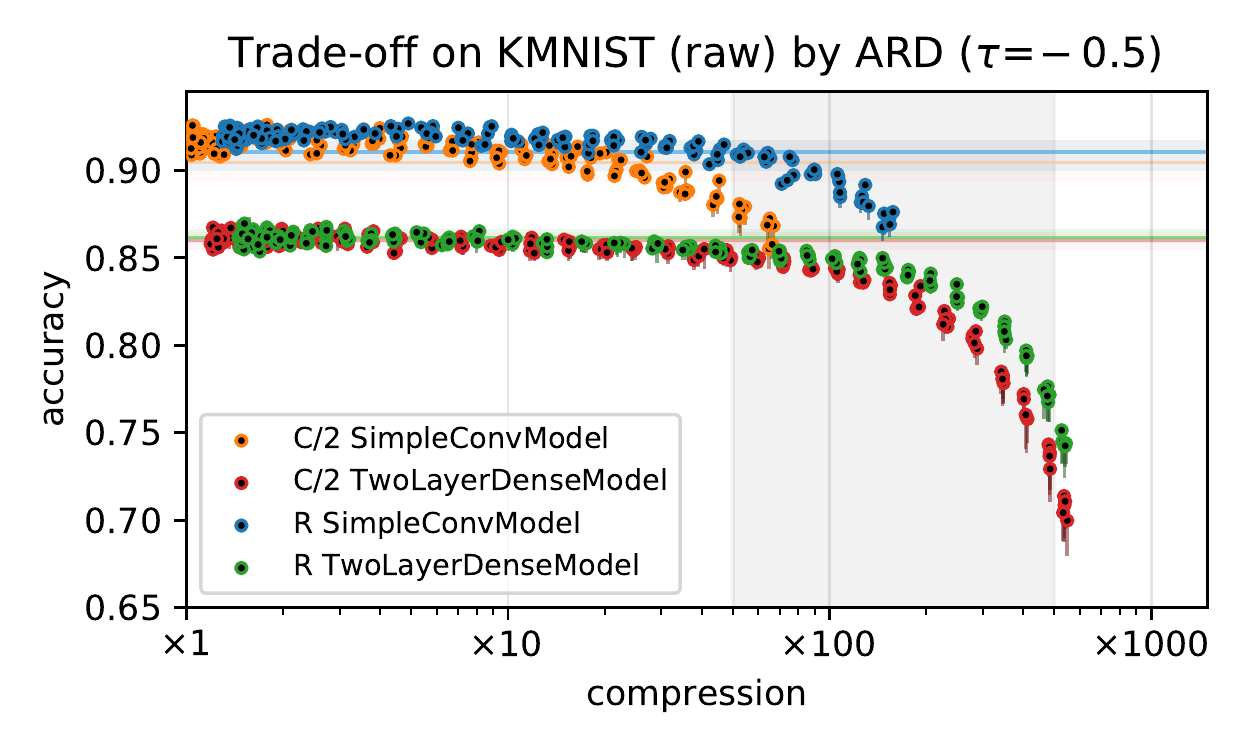}
  \end{subfigure} \\%
  \begin{subfigure}[b]{0.5\columnwidth}
    \centering
    \includegraphics[width=\linewidth]{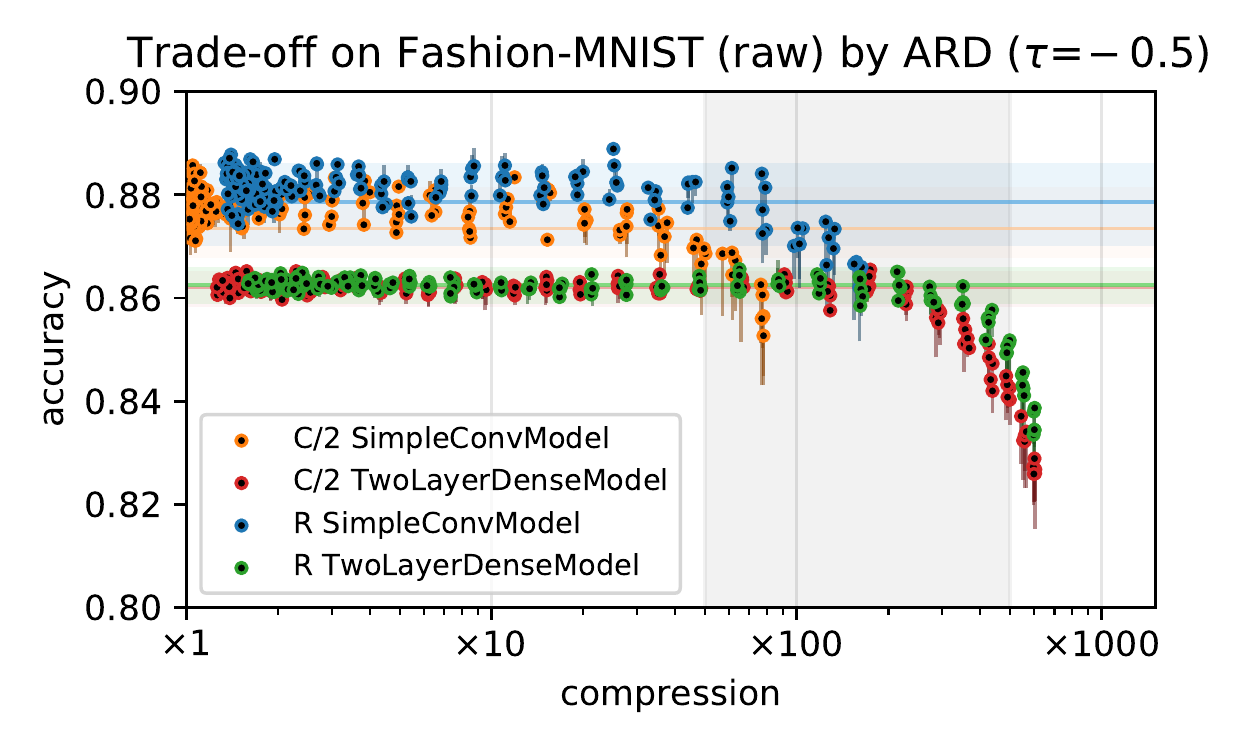}
  \end{subfigure}%
  \begin{subfigure}[b]{0.5\columnwidth}
    \centering
    \includegraphics[width=\linewidth]{figure__mnist-like__trade-off/appendix__cmp__ARD__mnist__raw__-0.5.pdf}
  \end{subfigure}
  \caption{%
    The trade-off of ARD method for $\real$ and $\tfrac12\cplx$ models using raw features.
  }
  \label{fig:appendix__cmp__mnist-like__trade-off__ARD__raw}
\end{figure}

\begin{figure}[b]
  \centering
  \begin{subfigure}[b]{0.5\columnwidth}
    \centering
    \includegraphics[width=\linewidth]{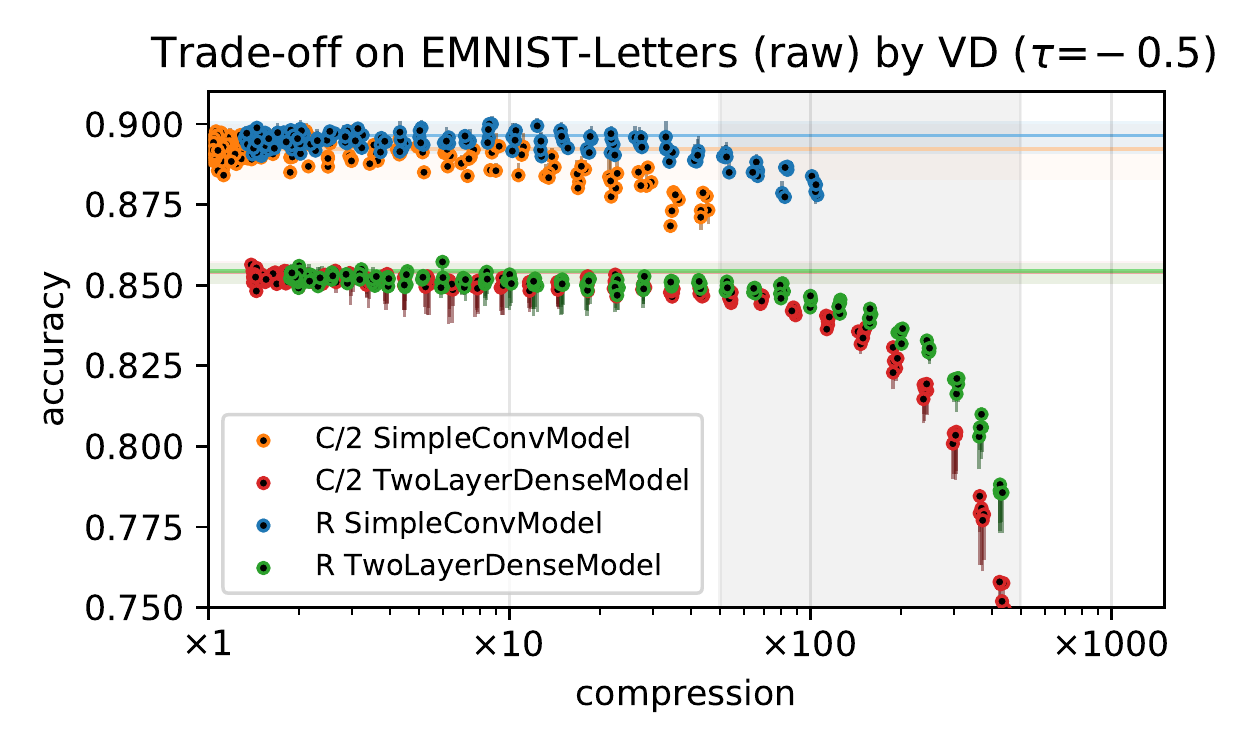}
  \end{subfigure}%
  \begin{subfigure}[b]{0.5\columnwidth}
    \centering
    \includegraphics[width=\linewidth]{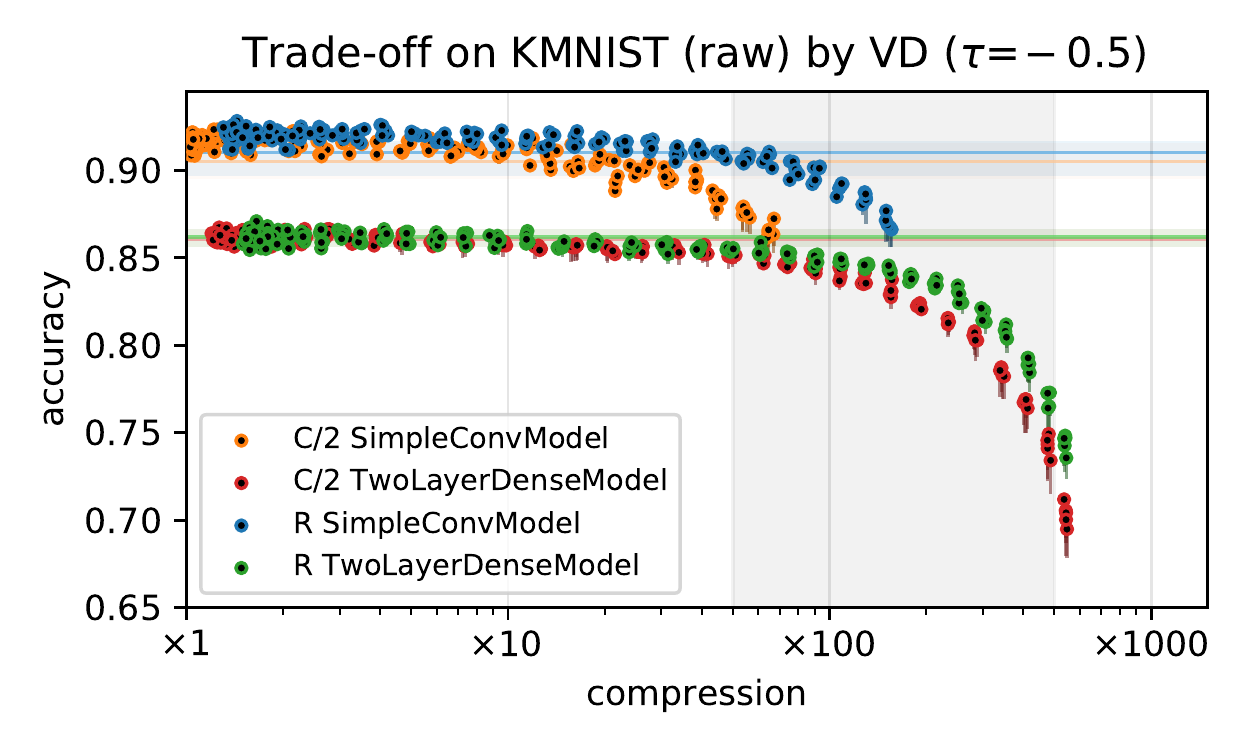}
  \end{subfigure} \\%
  \begin{subfigure}[b]{0.5\columnwidth}
    \centering
    \includegraphics[width=\linewidth]{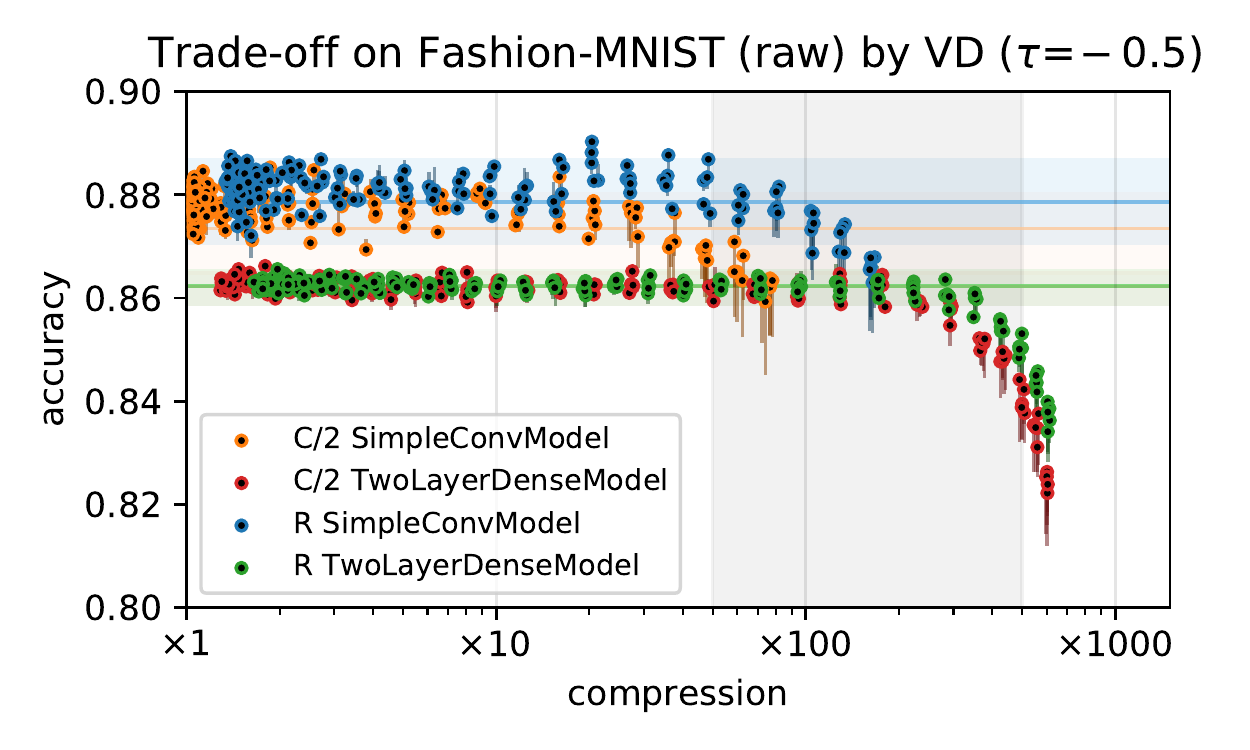}
  \end{subfigure}%
  \begin{subfigure}[b]{0.5\columnwidth}
    \centering
    \includegraphics[width=\linewidth]{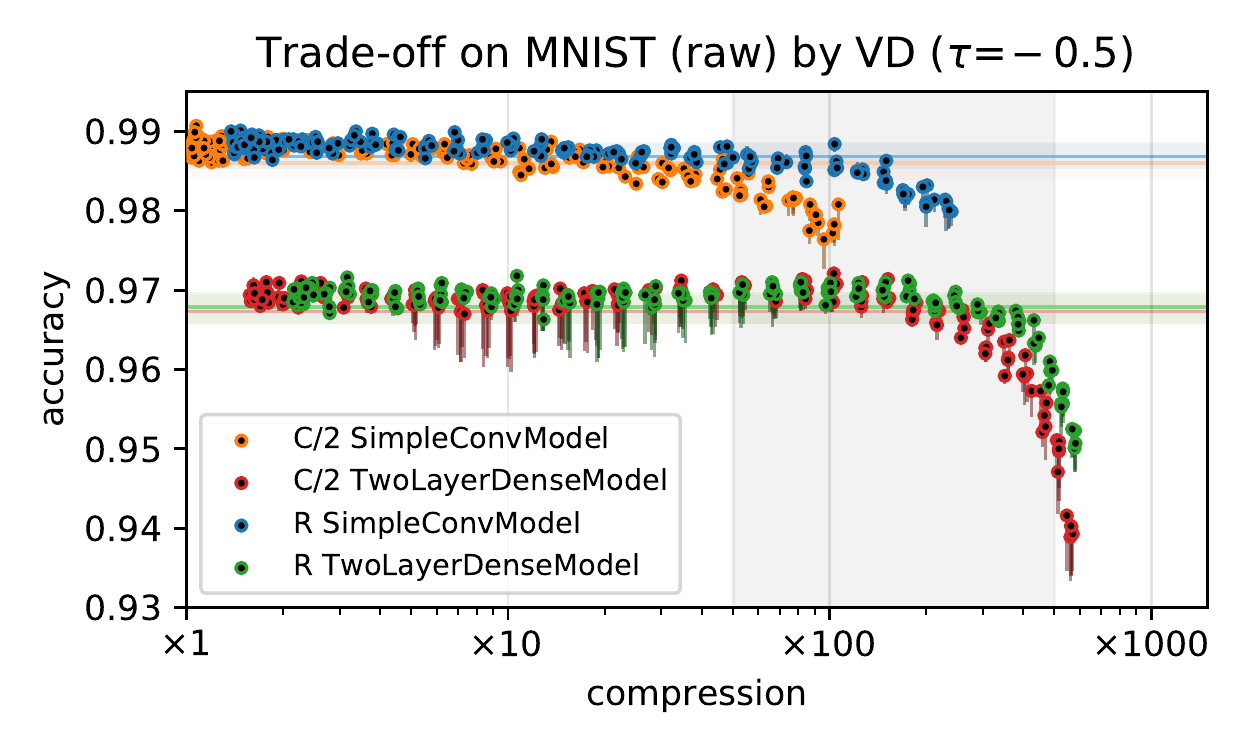}
  \end{subfigure}
  \caption{%
    The trade-off of VD method for $\real$ and $\tfrac12\cplx$ models using raw features.
  }
  \label{fig:appendix__cmp__mnist-like__trade-off__VD__raw}
\end{figure}

We compare $\real$ networks against $\tfrac12 \cplx$ with half the number of parameters
for raw input features on figures \ref{fig:appendix__cmp__mnist-like__trade-off__ARD__raw},
and \ref{fig:appendix__cmp__mnist-like__trade-off__VD__raw}, and $2 \real$ with
double the number of parameters against $\cplx$ for Fourier input features on figures
\ref{fig:appendix__cmp__mnist-like__trade-off__ARD__fft} and
\ref{fig:appendix__cmp__mnist-like__trade-off__VD__fft}.

% \clearpage

% section mnist_like_experiments (end)

\section{Complex-valued local reparameterization} % (fold)
\label{sec:complex_valued_local_reparameterization}

% Why does this warrant an extra section in the appendix?
In this section we show \eqref{eq:cplx-gauss-trick}.

%notation
By $e_i\in \real \hookrightarrow \cplx$ we denote the $i$-th unit vector of dimensionality
\emph{conforming} to the matrix-vector expression it is used in, $[M]$ denotes \emph{row-major}
flattening of a matrix $M$ into a vector, i.e. in lexicographic order of its indices.
Furthermore $\diag{(\cdot)}$ embeds vectors into matrices with zeros everywhere except
the diagonal, and $\otimes$ is the Kronecker product, for which we note the following
identities $
  [P Q R] = (P \otimes R^\top) [Q]
$, $
  (P \otimes Q)^\top = (P^\top \otimes Q^\top)
$, and $
  (P \otimes R) (C \otimes S) = P Q \otimes R S
$ \citep{petersen_matrix_2012}.
% It follows from the definition of the Kronecker product
% of $P$ and $R^\top$ and the row-major order vectorization.

If we assume a factorized $\cplx$-Gaussian approximation \eqref{eq:c-gauss-vi-general}
for $
  W \in \cplx^{n\times m}
$, then $[W]$ is $\cplx$-Gaussian vector with
\begin{equation}  \label{eq:c-gauss-vi-general-vec}
  [W]
    \sim \mathcal{C}\mathcal{N}_{nm} \bigl(
      [\mu], \diag{[\Sigma]}, \diag{[C]}
    \bigr)
  \,,
\end{equation}
where with $C_{ij} = \Sigma_{ij} \xi_{ij}$, $\Sigma_{ij} \geq 0$, and $
  \lvert C_{ij} \rvert^2 \leq \Sigma_{ij}
$. Then for any $x \in \cplx^m$ and $b\in \cplx^n$ we have $
  y = W x + b
    = (I_n \otimes x^\top) [W] + b
$, whence the covariance and relation matrices of $y$ are
\begin{eqnarray}
  \bigl(I_n \otimes x^\top\bigr)
    \diag{[\Sigma]}
  \conj{\bigl(I_n \otimes x^\top\bigr)^\top}
    &=& \sum_{ij}
      \bigl(I_n \otimes x^\top\bigr)
      \biggl(
        (e_i \otimes e_j)
          \Sigma_{ij}
        (e_i \otimes e_j)^\top
      \biggr)
      \conj{\bigl(I_n \otimes x^\top\bigr)^\top}
    \notag \\
    &=& \sum_{ij}
      \bigl(e_i \otimes x^\top e_j \bigr)
        \Sigma_{ij}
      \conj{\bigl(e_i \otimes x^\top e_j \bigr)^\top}
    \notag \\
    % &=& \sum_{ij}
    %     \Sigma_{ij}
    %   \bigl(e_i e_i^\top \otimes (x^\top e_j) \conj{(x^\top e_j)^\top} \bigr)
    % \notag \\
    % &=& \sum_{i} (e_i e_i^\top) \sum_{j}
    %     \Sigma_{ij} (x^\top e_j)(e_j^\top \conj{x})
    % \notag \\
    &=&
    %   \sum_{i} (e_i e_i^\top) \sum_{j}
    %     \Sigma_{ij} (x_j \conj{x_j})
    % =
      \sum_{i=1}^n (e_i e_i^\top)
        \biggl\{
          \sum_{j=1}^m \Sigma_{ij} \lvert x_j \rvert^2
        \biggr\}
    \,,  \label{eq:cn-gauss-lrt-cov} \\
  \bigl(I_n \otimes x^\top\bigr)
    \diag{[C]}
  \bigl(I_n \otimes x^\top\bigr)^\top
    &=& \sum_{ij}
      \bigl(I_n \otimes x^\top\bigr)
      \biggl(
        (e_i \otimes e_j)
          C_{ij}
        (e_i \otimes e_j)^\top
      \biggr)
      \bigl(I_n \otimes x^\top\bigr)^\top
    \notag \\
    % &=& \sum_{ij}
    %   \bigl(e_i \otimes x^\top e_j \bigr)
    %     C_{ij}
    %   \bigl(e_i \otimes x^\top e_j \bigr)^\top
    % \notag \\
    % &=& \sum_{ij}
    %     C_{ij}
    %   \bigl(e_i e_i^\top \otimes (x^\top e_j) (x^\top e_j)^\top \bigr)
    % \notag \\
    % &=& \sum_{i} (e_i e_i^\top) \sum_{j}
    %     C_{ij} (x^\top e_j)(e_j^\top x)
    % \notag \\
    &=&
    %   \sum_{i} (e_i e_i^\top) \sum_{j}
    %     C_{ij} (x_j x_j)
    % =
      \sum_{i=1}^n (e_i e_i^\top)
        \biggl\{
          \sum_{j=1}^m C_{ij} x_j^2
        \biggr\}
    \,.  \label{eq:cn-gauss-lrt-rel}
\end{eqnarray}
Since \eqref{eq:cn-gauss-lrt-cov} and \eqref{eq:cn-gauss-lrt-rel} are diagonal, the
vector $y$ has independent univariate $\cplx$-Gaussian components, whence
\eqref{eq:cplx-gauss-trick} follows.

% section complex_valued_local_reparameterization (end)

\section{Backpropagation through $\cplx$-networks} % (fold)
\label{sub:wirtinger_calculus}

% essential intro into Wirtinger calculus (CR)
% https://math.stackexchange.com/a/444493
Wirtinger ($\cplx\real$) calculus relies on the natural identification of $\cplx$ with $
  \real^2
$, and regards $
  f\colon \cplx \to \cplx
$ as an algebraically equivalent function $F\colon \real^2 \to \cplx$ defined $
  f(z) = f(u + \iu v) = F(u, v)
$. It enables general treatment of functions of vector $\cplx$-argument that possess partial
derivatives with respect to real and imaginary parts, yet are not required to satisfy
Cauchy-Riemann conditions. In $\cplx\real$ calculus the complex argument $z$ and its
conjugate $\conj{z}$ act as independent variables and $f(z)$ is treated as $f(z, \conj{z})$
by way of geometric transformations $z = u + \iu v$ and $\conj{z} = u - \iu v$.

Wirtinger partial derivative operators are formally defined as $
  \tfrac{\partial}{\partial z}
    = \tfrac12 \bigl(
      \tfrac{\partial}{\partial u}
      - \iu \tfrac{\partial}{\partial v}
    \bigr)
$ and $
  \tfrac{\partial}{\partial \conj{z}}
    = \tfrac12 \bigl(
      \tfrac{\partial}{\partial u}
      + \iu \tfrac{\partial}{\partial v}
    \bigr)
$ and differentials are $dz = du + \iu dv$ and $d\conj{z} = du - \iu dv$. In this paradigm
The usual rules of calculus, like chain and product rules, follow directly
from the definition of the operators, e.g.
\begin{equation}
    \frac{\partial (f\circ g)}{\partial z}
    = \frac{\partial f(g(z))}{\partial g} \frac{\partial g(z)}{\partial z}
    + \frac{\partial f(g(z))}{\partial \conj{g}} \frac{\partial \conj{g(z)}}{\partial z}
    % = \nabla G \nabla F
  \,.  \notag
\end{equation}
The total differential of $f$ at $z = u + \iu v \in \cplx$ is
\begin{eqnarray}  
df(z)
  &=& \frac{\partial f}{\partial z} dz
    + \frac{\partial f}{\partial \conj{z}} d\conj{z}
  \notag \\
  % &=& \frac12\biggl(
  %    \frac{\partial F}{\partial u}
  %    - \iu \frac{\partial F}{\partial v}
  % \biggr) (du + \iu dv)
  % + \frac12\biggl(
  %    \frac{\partial F}{\partial u}
  %    + \iu \frac{\partial F}{\partial v}
  % \biggr) (du - \iu dv)
  % \notag \\
  &=& \frac12 \biggl(
     \frac{\partial F}{\partial u} du
     - \cancel{\iu \frac{\partial F}{\partial v} du}
     + \cancel{\iu \frac{\partial F}{\partial u} dv}
     + \frac{\partial F}{\partial v} dv
  \biggr)
  + \frac12 \biggl(
     \frac{\partial F}{\partial u} du
     + \cancel{\iu \frac{\partial F}{\partial v} du}
     - \cancel{\iu \frac{\partial F}{\partial u} dv}
     + \frac{\partial F}{\partial v} dv
  \biggr)
  \notag \\
  % &=& \frac{\partial F}{\partial u} du
  %    + \frac{\partial F}{\partial v} dv
  % \notag \\
  &=& dF(u, v)
  \,,  \notag
\end{eqnarray}
At the same time the Cauchy-Riemann conditions $
  -\iu \tfrac{\partial F}{\partial v} = \tfrac{\partial F}{\partial u}
$ can be expressed as $
  \tfrac{\partial f}{\partial \conj{z}} = 0
$. Thus $\cplx\real$ calculus subsumes the usual $\cplx$-calculus of holomorphic functions,
since $f(z) = f(z, \conj{z})$ is constant with respect to $\conj{z}$ in the latter. 

In optimization-related tasks the objective is $
  f\colon \cplx\to \real
$, meaning that if it were to satisfy the Cauchy-Riemann conditions, then it necessarily
should have been constant. Nevertheless, the expression of the $\cplx\real$ gradient is
compatible with what is expected, when $f$ is treated like a $\real^2$ function. For such
$f$ we have $\conj{f} = f$, which implies $
  \tfrac{\partial f}{\partial \conj{z}}
    = \tfrac{\partial \conj{f}}{\partial \conj{z}}
    = \conj{\tfrac{\partial f}{\partial z}}
$, whence
\begin{equation*}
  df
    = \tfrac{\partial f}{\partial z} dz
      + \tfrac{\partial f}{\partial \conj{z}} d\conj{z}
    % = \tfrac{\partial f}{\partial z} dz
    %   + \conj{\tfrac{\partial f}{\partial z}} d\conj{z}
    = \tfrac{\partial f}{\partial z} dz
      + \conj{\tfrac{\partial f}{\partial z} dz}
    % = 2 \Re
    %   \Bigl(
    %     \conj{\tfrac{\partial f}{\partial \conj{z}}} dz
    %   \Bigr)
    = 2 \Re
      \bigl(
        \tfrac{\partial f}{\partial z} dz
      \bigr)
    \,.
\end{equation*}
Therefore the gradient of $f$ at $z$ is given by $
  \nabla_{\conj{z}} f(z)
    = \conj{\tfrac{\partial f}{\partial z}}
    = \tfrac{\partial F}{\partial u}
    + \iu \tfrac{\partial F}{\partial v}
$. The identification $\cplx \simeq \real^2$, backed by Wirtinger calculus, and emulation
of $\cplx$-arithmetic in computational graphs with $\real$-valued operations makes it
possible to reuse $\real$ back-propagation and existing auto-differentiation frameworks.

% section wirtinger_calculus (end)

\section{Gradient of the KL-divergence in $\real$ case} % (fold)
\label{sec:real-chisq-grad}  % gradient_of_the_kl_divergence_in_R_case

In this appendix we study the approximation proposed by \citet{molchanov_variational_2017}
for the KL divergence term \eqref{eq:improper-kl-div-real} for $\real$ Sparse Variational
Dropout. Following the logic of \citet{lapidoth_capacity_2003} we derive the expression
for $
  \tfrac{d}{d\log \alpha} K(\alpha)
$. Acknowledging that the same result was obtained by \citet[eq.~(5)]{hron_variational_2018},
we provide this appendix for the sake of completeness.
% The derivation in this section different from \citet{lapidoth_capacity_2003}.
% The results are identical: they use DCT to establish interchangeability of limits
% (sum-integral-derivative), whilst we allude to calculus theorems on differentiation
% and integration of power series within the radius of their convergence (which are
% essentially corollaries to the DCT).

% \citet{pav_moments_2015} correctly notice that \cite[p. 2466]{lapidoth_capacity_2003} has a missing $\log{2}$ term.

For $(z_i)_{i=1}^m \sim \mathcal{N}(0, 1)$ iid and $
  (\mu_i)_{i=1}^m \in \mathbb{R}
$, the random variable $W = \sum_i (\mu_i + z_i)^2$ has non-central $\chi^2$ distribution
with shape $m$ and non-centrality parameter $\lambda = \sum_i \mu_i^2$, i.e. $
  W\sim \chi^2_m(\lambda)
$. Therefore, the divergence \eqref{eq:improper-kl-div-real} has the form
\begin{equation}  \label{eq:kl-div-appendix}
  K(\alpha)
    \propto \frac12 \mathbb{E}_{
        W\sim \chi^2_1\bigl(\tfrac1{\alpha}\bigr)
      } \log W
    \tag{\ref{eq:improper-kl-div-real}'}
  \,.
\end{equation}
$W$ can alternatively be represented as a Poisson mixture of ordinary $\chi^2$
distributions: if $Z_{\mid J} \sim \chi^2_{m + 2J}$ for $
  J \sim \mathcal{P}ois(\tfrac{\lambda}2)
$ then $W \sim Z$. Therefore, expanding the conditional expectation gives
\begin{equation}  \label{eq:cond_exp_expansion}
  \mathbb{E}_{W\sim \chi^2_m(\lambda)} \log W
    = \mathbb{E}\Bigl(
        \mathbb{E} \bigl( \log W \mid J)
      \Bigr)
    = \mathbb{E}_{J\sim {\mathcal{P}ois}(\tfrac\lambda2)}
      \Bigl(
        \mathbb{E}_{W\sim \chi^2_{m + 2 J}} \log W
      \Bigr)
    \,.
\end{equation}
Since $\chi^2_\nu$ is Gamma distribution $\Gamma(\tfrac\nu2, \tfrac12)$, it can be shown
that the logarithmic moment $
  \mathbb{E}_{W\sim \chi^2_\nu} \log W
$ is $
  \psi\bigl(\tfrac\nu{2}\bigr) - \log \tfrac12
$, where $\psi$ is the digamma function ($\psi(x) = \tfrac{d}{dx} \log \Gamma(x)$).
% \begin{equation}  \label{eq:digamma}
%   \psi(x)
%     % = \frac{d}{dx} \log \Gamma(x)
%     = \frac1{\Gamma(x)}
%       \int_0^\infty
%         u^{x-1} e^{-u} \log{u}
%       \, du
%     \,.
% \end{equation}
By expanding expectation of a Poisson random variable we get $
  \mathbb{E}_{W\sim \chi^2_m(\lambda)} \log W
    = \log2 + g_m\bigl(\tfrac\lambda2\bigr)
$, where
\begin{equation}  \label{eq:exp_digamma}
  g_m(x)
    % = \mathbb{E}_{J\sim {\mathcal{P}ois}(x)}
    %   \psi\bigl(\tfrac{m + 2 J}{2}\bigr)
    = e^{- x} \sum_{j\geq 0}
        \frac{x^j}{j!}
          \psi\bigl(\tfrac{m + 2 J}{2}\bigr)
    \,.
\end{equation}
Making use of the property $
  \psi(z + 1) = \psi(z) + \tfrac1z
  % \psi(1 + J + \tfrac{m}2) = \tfrac2{m + 2J} + \psi(J + \tfrac{m}2)
$ of the digamma funciton for $z > 0$, we conclude that the power series in \eqref{eq:exp_digamma}
converges for any $x \geq 0$. Therefore the derivative of \eqref{eq:exp_digamma} is
given by
\begin{equation}  \label{eq:g_m_series}
  \frac{d}{dx} g_m(x)
    % = \mathbb{E}_{J\sim {\mathcal{P}ois}(x)}
    %     \bigl( \tfrac{J}x - 1 \bigr)
    %   \psi\bigl(\tfrac{m + 2 J}{2}\bigr)
    = - g_m(x)
      + e^{- x} \sum_{j \geq 0}
          \frac{x^j}{j!}
        \biggl(
          \psi\bigl(\tfrac{m + 2 j}{2}\bigr)
          + \frac{2}{m + 2 j}
        \biggr)
    % = e^{- x} \sum_{j \geq 0}
    %     \frac{x^j}{j!} \frac{2}{m + 2 j}
    \,.
\end{equation}
By manipulating the partial sums within \eqref{eq:g_m_series} we get
\begin{equation}  \label{eq:g_m_series_int}
  \frac{d}{dx} g_m(x)
    = e^{- x} \sum_{j \geq 0}
      \frac{x^j}{j!} \frac1{j + \tfrac{m}2}
    % = e^{- x} x^{- \tfrac{m}2} \sum_{j \geq 0}
    %   \frac1{j!}
    %   \frac{x^{j + \tfrac{m}2}}{j + \tfrac{m}2}
    = e^{- x} x^{- \tfrac{m}2} \sum_{j \geq 0}
      \frac1{j!} \int_0^x t^{j + \tfrac{m}2 - 1} dt
    \,.
\end{equation}
Furthermore, the functions $
  t \mapsto \sum_{j=0}^J \frac1{j!} t^{j + \tfrac{m}2 - 1}
$ are non-decreasing on $(0, x)$ with growing $J$ and converge to $
  t^{\tfrac{m}2 - 1} e^t
$, which implies by the Monotone Convergence Theorem that
\begin{equation}  \label{eq:g_m_integral}
  \frac{d}{dx} g_m(x)
    = e^{- x} x^{- \tfrac{m}2}
      \int_0^x \sum_{j \geq 0}
        \frac1{j!} t^{j + \tfrac{m}2 - 1} dt
    = e^{- x} x^{- \tfrac{m}2}
      \int_0^x t^{\bigl(\tfrac{m}2 - 1\bigr)} e^t dt
    \,.
\end{equation}
Substituting $u^2 = t$ on $[0, \infty]$ with $2u du = dt$ and letting $
  I_m
  % \colon \mathbb{R} \to \mathbb{R}
  \colon x \mapsto e^{-x^2} \int_0^x u^{m - 1} e^{u^2} du
$ yields
\begin{equation}  \label{eq:g_m_integral-sub}
  \frac{d \eqref{eq:cond_exp_expansion}}{d \lambda}
    = \frac12 \frac{d}{dx} g_m(x)
        \bigg\vert_{x=\tfrac\lambda2}
    % = x^{-\tfrac{m}2} e^{-x}
    %     \int_0^x t^{\tfrac{m}2 - 1} e^t dt
    = e^{-x}
      x^{-\tfrac{m}2}
      \int_0^{\sqrt{x}} u^{m - 1} e^{u^2} du
        \bigg\vert_{x=\tfrac\lambda2}
    % = e^{-x}
    %   \frac1{\sqrt{x}}
    %   \int_0^{\sqrt{x}} e^{u^2} du
    %     \bigg\vert_{x=\tfrac\lambda2}
    = \Bigl(\sqrt{\tfrac2\lambda}\Bigr)^m I_m\Bigl(\sqrt{\tfrac\lambda2}\Bigr)
    \,.
\end{equation}
Since $\alpha$ is non-negative, it is typically parameterized via its logarithm, whence
the derivative of \eqref{eq:kl-div-appendix} with respect to $\log\alpha$ follows from
\eqref{eq:g_m_integral-sub} for $m=1$ and $\lambda = \tfrac1\alpha$:
\begin{equation}  \label{eq:real-kl-div-deriv-log}
  \frac{d K(\alpha)}{d\log \alpha}
    % = \tfrac12 \sqrt{\tfrac2\lambda} F\Bigl(\sqrt{\tfrac\lambda2}\Bigr)
    % for \lambda = \tfrac1\alpha
    %   \tfrac{d\lambda}{d\alpha} = - \alpha^{-2}
    %   \tfrac{d\alpha}{d\log \alpha} = \alpha
    % = - \alpha \frac1{\alpha\sqrt{2\alpha}} F\Bigl(\tfrac1{\sqrt{2\alpha}}\Bigr)
    = - \frac1{\sqrt{2\alpha}} I_1\bigl(\tfrac1{\sqrt{2\alpha}}\bigr)
    \,.
\end{equation}

\begin{figure}[!t]
  \centering
  \includegraphics[width=\columnwidth]{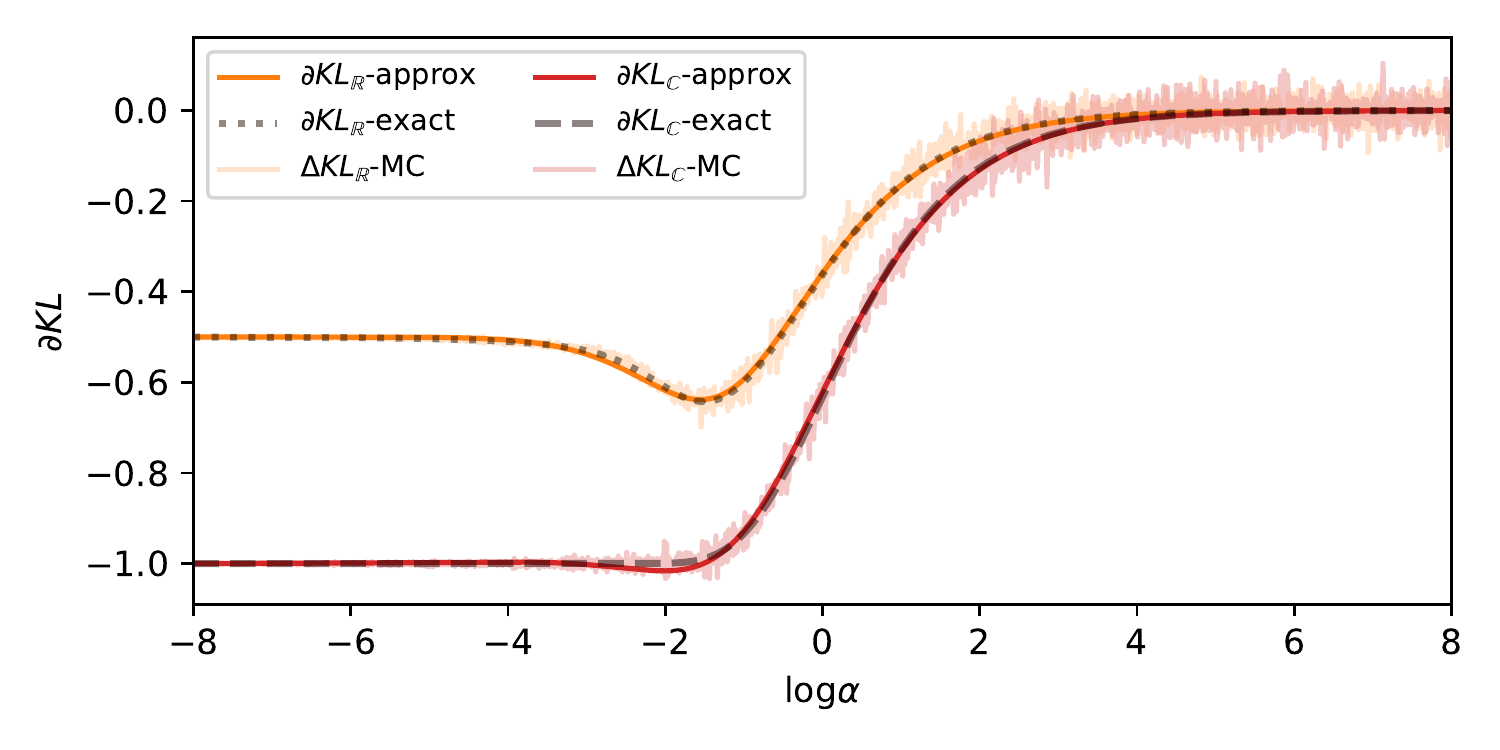}
  \caption{$\tfrac{d K(\alpha)}{d \log{\alpha}}$ of the approximation
  \eqref{eq:improper-kl-div-real-approx}, MC estimate of \eqref{eq:improper-kl-div-real},
  and the exact derivative using \eqref{eq:real-kl-div-deriv-log}.}
  \label{fig:molchanov-derivative-replica}
\end{figure}

We compute the Monte-Carlo estimate of \eqref{eq:improper-kl-div-real} on a sample
of $10^7$ draws over an equally spaced grid of $\log \alpha$ in $[-12, +12]$ of size
$4096$. The approximation proposed by \citet{molchanov_variational_2017} is given in
\eqref{eq:improper-kl-div-real-approx}, with coefficients $
  k_1 = 0.63576
$, $
  k_2 = 1.8732
$, and $
  k_3 = 1.48695
$. The derivative of the approximation with respect to $\log \alpha$ follows
\eqref{eq:real-kl-div-deriv-log} within $4\%$ of relative tolerance, see
fig.~\ref{fig:molchanov-derivative-replica}. 
% see eq.~(14): flit overall sign, use $C = -k_1$, $1-\sigma(x) = \sigma(-x)$
\begin{equation}  \label{eq:improper-kl-div-real-approx}
  \eqref{eq:improper-kl-div-real}
% - KL(\mathcal{N}(w \mid \mu, \alpha \mu^2) \|
%     \tfrac1{\lvert w \rvert})
%   \approx
%     k_1 \sigma(k_2 + k_3 \log \alpha)
%     + C \big\vert_{C = -k1}
%     - \tfrac12 \log (1 + e^{-\log \alpha})
  % = \tfrac12 \log \alpha
  %   - \mathbb{E}_{\varepsilon \sim \mathcal{N}(1, \alpha)}
  %   \log{\lvert \varepsilon \rvert} + C
  \approx
    \frac12 \log{\bigl(1 + e^{-\log \alpha}\bigr)}
    + k_1 \sigma\bigl(- (k_2 + k_3 \log \alpha)\bigr)
    % - k_1 \bigl(1 - \sigma(k_2 + k_3 \log \alpha)\bigr)
  \,,
\end{equation}
% in parctial implementations we suggest using softplus to numerically comptue the last
% term, since it is typically implemented in a floating-point stable fashion.
Similarly, the forward difference estimate of the derivative \eqref{eq:real-kl-div-deriv-log}
very closely (up to sampling error). For sake of completeness, we compute a similar
Monte-Carlo estimate for the KL divergence term in \eqref{eq:c-vd-kl-div} for $\cplx$-valued
Variational Dropout with $\beta = 2$, fit the best approximation \eqref{eq:improper-kl-div-real-approx},
and compare it against the exact $\log \alpha$ derivative $
  \frac{d \eqref{eq:c-vd-kl-div}}{d\log \alpha}
    % = \frac{d \alpha}{d\log \alpha} \frac{d}{d\alpha} \bigl(
    %   \log \tfrac1\alpha - \mathop{Ei}\bigl(-\tfrac1\alpha\bigr)
    % = \alpha \bigl(
    %   - \tfrac1\alpha
    %   - \tfrac{e^x}{x} \bigg\vert_{x=-\tfrac1\alpha} \alpha^{-2}
    % \bigr)
    = e^{-\tfrac1\alpha} - 1
$.

% section real-chisq-grad (end)

% section appendix (end)

\end{document}